\documentclass{article}
\usepackage{paper_style,times}

\usepackage[utf8]{inputenc} %
\usepackage[T1]{fontenc}    %
\usepackage{hyperref}       %
\usepackage{url}            %
\usepackage{booktabs}       %
\usepackage{amsfonts}       %
\usepackage{nicefrac}       %
\usepackage{silence}
\usepackage{microtype}      %
\usepackage{xcolor}         %
\usepackage{amsmath}
\usepackage{algpseudocode}
\usepackage{graphicx}
\usepackage{svg}
\usepackage{float}
\usepackage{afterpage}
\usepackage{wrapfig}
\usepackage{tabularx}
\usepackage{longtable}
\usepackage{caption}
\usepackage[belowfloats]{footmisc}
\usepackage{listings}
\usepackage[most]{tcolorbox}
\usepackage{algorithm}
\usepackage{placeins}
\usepackage{multirow}

\definecolor{idebg}{HTML}{1E1E1E}
\definecolor{idefg}{HTML}{D4D4D4}
\definecolor{idekw}{HTML}{569CD6}
\definecolor{idestr}{HTML}{CE9178}
\definecolor{idecom}{HTML}{6A9955}

\lstdefinestyle{idepython}{
  language=Python,
  basicstyle=\ttfamily\small\color{idefg},
  keywordstyle=\color{idekw},
  stringstyle=\color{idestr},
  commentstyle=\color{idecom},
  backgroundcolor=\color{idebg},
  showstringspaces=false,
  breaklines=true,
  columns=fullflexible,
  keepspaces=true,
  frame=none,
  numbers=none,
  xleftmargin=0pt,
  aboveskip=0pt,
  belowskip=0pt
}

\newtcblisting{idecode}{
  listing only,
  listing style=idepython,
  colback=idebg,
  colframe=black!35,
  arc=2mm,
  boxrule=0.4pt,
  left=3mm,
  right=3mm,
  top=2mm,
  bottom=2mm
}

\usepackage{array}

\definecolor{boxbg}{HTML}{FFFFFF}
\definecolor{systemframe}{HTML}{CBD5E1}
\definecolor{agentframe}{HTML}{93C5FD}
\definecolor{systemtitle}{HTML}{475569}
\definecolor{agenttitle}{HTML}{1D4ED8}
\definecolor{textgray}{HTML}{1F2937}
\definecolor{codebg}{HTML}{FAFAFA}
\definecolor{codeborder}{HTML}{E5E7EB}

\definecolor{pykw}{HTML}{0A66C2}
\definecolor{pystr}{HTML}{A31515}
\definecolor{pycom}{HTML}{6A737D}

\lstdefinestyle{macPython}{
  language=Python,
  basicstyle=\ttfamily\scriptsize\color{textgray},
  keywordstyle=\bfseries\color{pykw},
  stringstyle=\color{pystr},
  commentstyle=\itshape\color{pycom},
  backgroundcolor=\color{codebg},
  frame=single,
  rulecolor=\color{codeborder},
  framerule=0.35pt,
  framesep=5pt,
  breaklines=true,
  breakatwhitespace=false,
  columns=fullflexible,
  keepspaces=true,
  showstringspaces=false
}

\newtcolorbox{systembubble}{
  enhanced,
  breakable,
  width=\linewidth,
  colback=boxbg,
  colframe=systemframe,
  boxrule=0.55pt,
  arc=5pt,
  left=10pt,
  right=10pt,
  top=6pt,
  bottom=8pt,
  before skip=0.6em,
  after skip=0.6em,
  borderline west={2.4pt}{0pt}{systemtitle},
  fontupper=\small\color{textgray},
  title={System},
  fonttitle=\bfseries\small\color{systemtitle},
  coltitle=systemtitle,
  colbacktitle=boxbg,
  colframe=systemframe,
  titlerule=0.35pt,
  toptitle=4pt,
  bottomtitle=4pt,
  lefttitle=8pt,
  righttitle=8pt
}

\newtcolorbox{agentbubble}{
  enhanced,
  breakable,
  width=\linewidth,
  colback=boxbg,
  colframe=agentframe,
  boxrule=0.55pt,
  arc=5pt,
  left=10pt,
  right=10pt,
  top=6pt,
  bottom=8pt,
  before skip=0.6em,
  after skip=0.6em,
  borderline west={2.4pt}{0pt}{agenttitle},
  fontupper=\small\color{textgray},
  title={Agent},
  fonttitle=\bfseries\small\color{agenttitle},
  coltitle=agenttitle,
  colbacktitle=boxbg,
  colframe=agentframe,
  titlerule=0.35pt,
  toptitle=4pt,
  bottomtitle=4pt,
  lefttitle=8pt,
  righttitle=8pt
}

\definecolor{codebg}{HTML}{FAFAFA}
\definecolor{codeborder}{HTML}{E5E7EB}
\definecolor{textgray}{HTML}{1F2937}
\definecolor{pykw}{HTML}{0A66C2}
\definecolor{pyfun}{HTML}{7C3AED}
\definecolor{pystr}{HTML}{A31515}
\definecolor{pycom}{HTML}{6A737D}
\definecolor{pyop}{HTML}{374151}

\lstdefinestyle{paperPython}{
  language=Python,
  basicstyle=\ttfamily\small\color{textgray},
  keywordstyle=\bfseries\color{pykw},
  stringstyle=\color{pystr},
  commentstyle=\itshape\color{pycom},
  identifierstyle=\color{textgray},
  emph={
    LM,valid,overlap_feedback,feedback,trace_feedback,max_delta,add,contained_motifs,P
  },
  emphstyle=\color{pyfun},
  emph={[2]
    motifs,history,tests,delta,overlap,winning,drawish,max_overlap
  },
  emphstyle={[2]\color{pyop}},
  backgroundcolor=\color{codebg},
  frame=single,
  rulecolor=\color{codeborder},
  framerule=0.35pt,
  framesep=7pt,
  xleftmargin=0pt,
  xrightmargin=0pt,
  aboveskip=0.25em,
  belowskip=0.25em,
  breaklines=false,
  columns=fullflexible,
  keepspaces=true,
  showstringspaces=false,
  tabsize=4
}

\usepackage{binprop_frontpage}
\hypersetup{
  pdftitle={Explaining Chess and Go in Natural Language},
  pdfauthor={Ori Press and Alexander Golubev}
}
\title{Explaining Chess and Go in Natural Language}

\begin{document}

\thispagestyle{empty}
\begin{binpropfrontmatter}
  \vspace{0.3em}
  {\fontsize{18pt}{22pt}\selectfont\bfseries
    Explaining Chess and Go in Natural Language
  }\par
  \vspace{0.3em}

  \begin{spacing}{1.15}
    \textbf{Ori Press}\qquad \textbf{Alexander Golubev}
  \end{spacing}
  \vspace{0.5em}

  {\small Nebius AI\par}
  \vspace{3pt}
  {\small Correspondence: \href{mailto:oripress@nebius.com}{oripress@nebius.com}\par}
  \vspace{3pt}
  {\small \href{https://www.explainthatmove.com/}{%
    \tikz[baseline=-0.7ex,line width=0.35pt,draw=black]{%
      \draw (0,0) circle (3.4pt);
      \draw (0,0) ellipse [x radius=1.4pt,y radius=3.4pt];
      \draw (-3.4pt,0) -- (3.4pt,0);
    }\,ExplainThatMove.com}\par}
  \vspace{0em}

\begin{abstract}
Modern chess engines play far beyond human ability, often choosing moves whose underlying ideas are difficult for humans to understand. We propose an automatic metric for evaluating chess explanations, grounded in chess-engine evaluations. Evaluating state-of-the-art language models using this metric, we find that they perform surprisingly poorly, often relying on surface-level board observations rather than robust, transferable chess concepts. We then introduce BinProp, a training-free method that enables these same models to discover reusable chess concepts and use them to explain moves in natural language. BinProp improves explainability for every model tested, nearly tripling the average score compared with direct prompting. Beyond chess, the same method discovers useful concepts in Go, draughts, and shogi without additional tuning.
\end{abstract}
\end{binpropfrontmatter}

\begin{figure}[!b]
    \centering
    \includegraphics[width=0.996\linewidth,trim=0 0 0 14bp,clip]{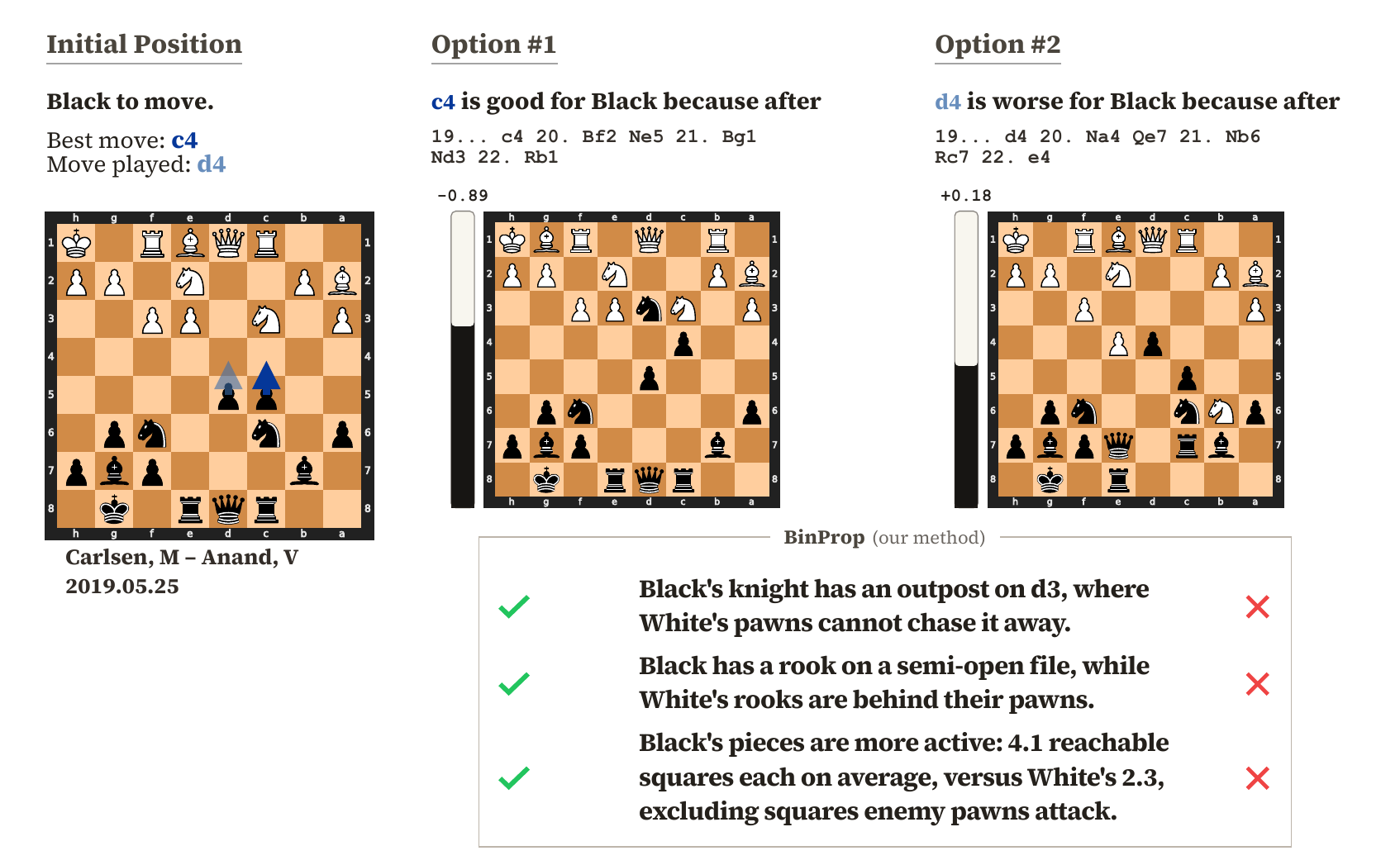}
    \caption{Black to move in Carlsen--Anand (2019). Anand played \texttt{d4} (light blue); Stockfish's top move is \texttt{c4} (dark blue). Our method, BinProp, explains the difference between the moves by finding properties that appear in one continuation but not the other. Further examples appear in \S~\ref{appdx:chess_game_examples}.}
    \label{fig:fig1}
\end{figure}

\section{Introduction}

Modern chess-playing algorithms, called chess engines, are vastly superior to human players, but their decisions are often difficult to understand. Since Shannon's formulation of computer chess~\citep{shannon1950xxii}, engines have followed the same basic paradigm: search through possible continuations, evaluate the resulting positions, and choose the move whose continuation looks best using MiniMax~\citep{v1928theorie}. Modern engines apply this framework at enormous scale: Stockfish 18 \citep{stockfish18}, the strongest chess engine at the time of writing,\footnote{Stockfish 18 is the highest-rated chess engine on the CCRL 40/15 rating list; see Computer Chess Rating Lists (CCRL), \url{https://computerchess.org.uk/4040/index.html}}, evaluates millions of positions per second using a neural-network evaluation function.

This gap between playing strength and understanding has recently attracted increasing attention~\citep{bjornsson2024chess, spinnato2025towards, barthelemy2025chess}. An engine's evaluation tells us which move it prefers, but not the chess ideas behind that choice. Our goal is to explain in natural language why a chess engine prefers one move over another.

Chess and Go engines are especially interesting to explain because they can reveal ideas that human players overlook. Chess masters highlighted AlphaZero's ``march of the rook's pawn'': against a kingside-castled king, it often advances the h-pawn to h6 to restrict the king's movement~\citep{sadler2019gamechanger}. In Game 2 against Lee Sedol, AlphaGo's Move 37 was so unusual that its team estimated only a one-in-10,000 chance of a human playing it~\citep{hassabis2016alphago}. Understanding the ideas behind engine moves could help improve human play.

Explaining why an engine prefers one move over another is challenging because the quality of a move cannot be judged solely by looking at the current board. A position with a material advantage, for example, may still be losing if the opponent has a reply that leads to forced checkmate. Instead, the value of a move depends on the continuations it enables and those it prevents. We therefore frame explanations as \emph{contrastive}: given two candidate moves, we ask what becomes true after the first move but not after the alternative. As shown in Fig.~\ref{fig:fig1}, these explanations take the form of concrete chess motifs, such as passed pawns, restricted pieces, or superior piece activity.

Our contributions are as follows. First, we propose an engine-grounded metric for measuring the quality of chess explanations. Second, we evaluate four open-weight and four frontier large language models and show that their explanations often rely on surface-level observations rather than robust, transferable chess concepts. Third, we introduce BinProp, a training-free method for discovering reusable concepts and using them to explain moves in natural language. BinProp improves explainability for every model tested on chess, raising \texttt{Fable 5}'s score from \(3.1\%\) with direct prompting to \(29.6\%\), and \texttt{GPT-5.6 Sol}'s from \(6.2\%\) to \(20.5\%\). It also extends to Go, draughts, and shogi without additional tuning.

\section{Evaluating Explanations}\label{sec:evaluation_explanations}

\afterpage{%
\begin{figure}[!t]
    \centering
    \includegraphics[width=0.98\linewidth]{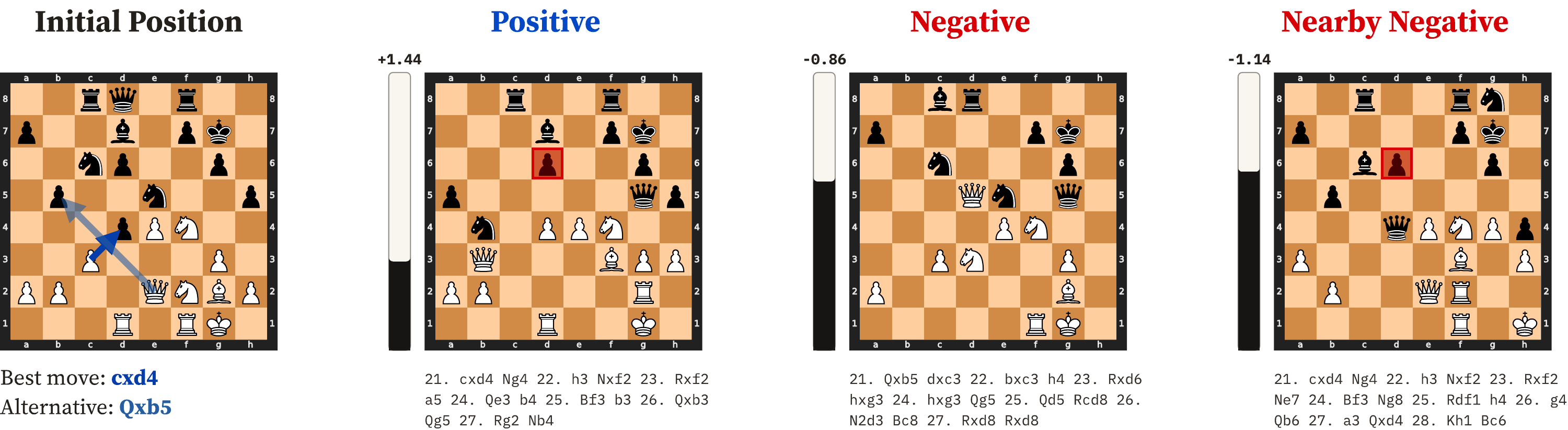}
    \caption{\textbf{Brittle explanations explain the positive and negative boards, but fail to generalize on nearby negatives.} Positive and nearby negative positions follow the top move in the starting position, \texttt{cxd4} (blue), while negatives start from the second-best move, \texttt{Qxb5} (light blue). The brittle motif ``Black has a pawn on d6'' (highlighted in red) returns \texttt{True} on the positive and \texttt{False} on the negative positions, but also returns \texttt{True} on the nearby negative position, yielding a low explainability score.}
    \label{fig:pos_neg_nearby}
\end{figure}
}

Our goal is to generate explanations such as: ``If you play this move, your opponent's king will come under pressure; after a weaker alternative, it will not.'' In this work, we focus on binary concepts, or motifs, that are either present or absent in a position, making explanations easier to interpret. In this section, we describe how we build a test set of positions to evaluate the quality of a given explanation.

The main challenge is that a motif may not appear immediately after a move is played: a move can be good because of what becomes possible only several moves later. We therefore use Stockfish 18 to generate multiple plausible continuations after the best move and a weaker alternative. In evaluation, we give models these Stockfish-generated continuations, so the evaluation tests motif quality rather than move calculation.

Concretely, we sample positions where the engine's top move gives a modest advantage, while an alternative move loses that advantage. We sample \textbf{positive} positions from continuations that begin with the top move and preserve the advantage, while \textbf{negative} positions come from continuations that begin with an alternative move. The task is to find a motif, written as a Python function, that returns \texttt{True} on positive positions and \texttt{False} on negative positions.

Separating positives from negatives is not enough: a motif may simply capture an irrelevant difference between the boards. Figure~\ref{fig:pos_neg_nearby} shows one example: \emph{Black has a pawn on d6} separates the two sets, but says little about why one side has an advantage. We call such motifs brittle. To catch them, we use \textbf{nearby negatives}: we keep the early moves of a positive continuation but replace a later move with a weaker one that loses the advantage. Because nearby negatives share a prefix of moves with positives, the resulting boards tend to be similar, helping catch brittle motifs.

\subsection{Evaluation Setup} 
\paragraph{Starting positions.} We randomly sample 100 positions from master-level games in PGNMentor~\citep{pgnmentor}. We exclude positions with an overwhelming advantage, where the explanation is often trivial, and forced mates, where the engine's mating line itself provides the explanation. The best move must evaluate between \(+75\) and \(+125\) centipawns\footnote{A centipawn is one hundredth of a pawn in engine evaluation. A \(75\)--\(125\) centipawn advantage corresponds to roughly a \(57\)--\(61\%\) win chance under Lichess's centipawn-to-win\% formula~\citep{lichessAccuracy}.} for the side to move. The alternative must evaluate between \(-125\) and \(+50\) centipawns, at least \(75\) centipawns below the best move and within \(50\) centipawns of the second-best move.

\begingroup
\setlength{\columnsep}{13pt}
\setlength{\intextsep}{4pt}
\begin{wrapfigure}{r}{0.58\linewidth}
    \centering
    \includegraphics[width=\linewidth]{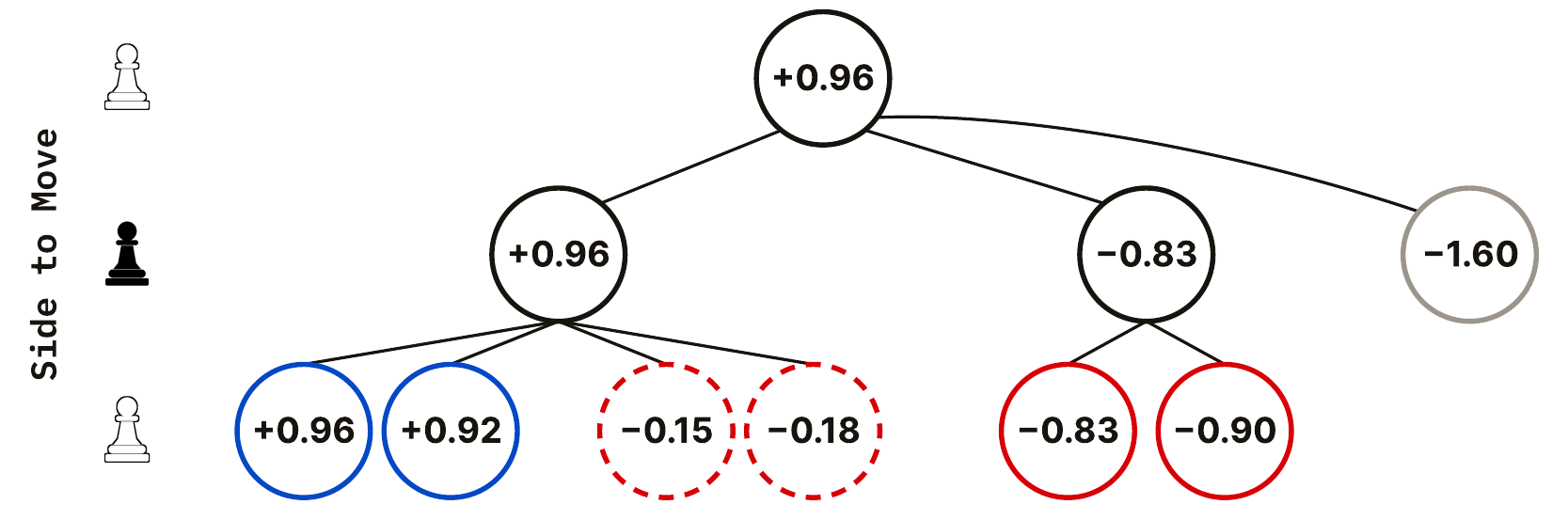}
    \caption{\textbf{Sampling positions, an illustrative example.} Nodes show engine evaluation. Blue/red marks positive/negative; dashed red marks nearby negatives. Gray moves score more than 50 centipawns below second-best and are discarded.}
    \label{fig:sampling-tree}
\end{wrapfigure}

\paragraph{Continuation sampling.} For each starting position, we construct a set of 12 positives, 12 negatives, and 8 nearby negatives. We retain only \emph{quiet} positions: neither king is in check and no immediate material tactic is visible. This requirement excludes boards midway through tactical exchanges (\S~\ref{appdx:chess_quiet_positions}). We sample these positions across several plausible continuations, rather than a single engine line (Figure~\ref{fig:sampling-tree}).

We start two continuations from the same position: one with the best move and one with the weaker alternative. After each, Stockfish plays both sides for at least six plies\footnote{A ply is a single player's move, so one move by each side is two plies.}, stopping at the first quiet position. If none is reached by ply 16, we discard the line.

\endgroup

To generate the set of boards, we branch into several continuations from each quiet position. We explore up to 12 additional plies, considering up to six highly ranked Stockfish moves at each turn. We retain 12 distinct quiet positions after each of the two moves: positives preserve the best move's advantage, while negatives start with the worse move and keep roughly the same evaluation. Every retained position has the same player to move as in the starting position.

\paragraph{Nearby negatives.} We start from a quiet position reached along a positive continuation, then have the advantaged player play a weaker move instead of Stockfish's best move. After Stockfish's best reply, we retain the position as a nearby negative only if it is quiet and the advantage is lost. We collect eight per starting position.

\paragraph{Evaluation.} We show the model four positives and four negatives, called the \emph{support set}. Nearby negatives are never shown to the model. We then evaluate its motifs on 24 positions it has not seen: eight positives, eight negatives, and eight nearby negatives. For each group, we compute the fraction of correct outputs: \texttt{True} for positives and \texttt{False} for either kind of negative. A motif's explainability score is the lowest of these three fractions. Additional dataset construction details are in \S~\ref{appdx:chess_data_collection}.

\section{Method}

In this section, we introduce BinProp (short for \textbf{Bin}ary \textbf{Prop}erties), our method for explaining chess engine decisions. BinProp first finds general chess motifs that occur more often in winning than losing positions then uses them to explain why the engine prefers one move over another.

\paragraph{Finding Chess Motifs.}

To find these motifs, we construct a set of 200k positions from the Lichess Elite Database~\citep{lichessEliteDatabase}. All evaluations are expressed from the perspective of the player to move: 100k positions evaluate between \(+125\) and \(+250\) centipawns, and 100k evaluate between \(-250\) and \(-125\) centipawns. We call these the winning and losing sets, respectively. 

BinProp runs as an iterative loop. At each step, a language model proposes a motif: a Python function \(f\) that maps a board state to \texttt{True} or \texttt{False}. We evaluate \(f\) on both sets; a useful motif fires more often on winning positions. For example, consider the motif ``the player to move is ahead in material.'' It fires on \(61.5\%\) of
winning positions and \(5.5\%\) of losing positions, a gap of
\(56\) percentage points. In general, \(\Delta(f)\) is the difference between a motif's firing rates on the winning and losing sets, and we retain only motifs with \(\Delta(f)>0.01\), i.e., more than one percentage point.

We also want the motifs to be different from one another. To check this, we measure \emph{containment} on the winning set. For a candidate motif, containment is its largest overlap with any existing motif: among winning positions where the candidate is \texttt{True}, what fraction also satisfy that existing motif? High containment indicates redundancy between motifs. We reject a candidate when containment exceeds \(70\%\), unless its \(\Delta(f)\) is larger than that of the overlapping motif; in that case, the candidate replaces the older motif.

\begin{algorithm}[h]
\small
\caption{BinProp motif discovery loop.}
\label{fig:motif-discovery}
\begin{algorithmic}[1]
\State $\mathcal{B} \gets \emptyset$ \Comment{Motif Set}
\State $H \gets \emptyset$ \Comment{Conversation History}
\For{$t=1,\ldots,100$}
    \State $f \gets \mathrm{LM}(\mathcal{B}, H)$
        \Comment{Propose one binary motif}
    \If{$f$ compiles and runs}
        \State $\Delta(f) \gets \Pr[f=1 \mid W] - \Pr[f=1 \mid L]$
            \Comment{Measure separation}
        \State $C \gets \{g \in \mathcal{B} :
            \operatorname{containment}_{W}(f,g) > \tau\}$
            \Comment{Find overlapping motifs}
        \If{$\Delta(f)>0.01$ and
            $\left(C=\emptyset \lor
            \Delta(f)>\max_{g\in C}\Delta(g)\right)$}
            \State $\mathcal{B} \gets (\mathcal{B}\setminus C)\cup\{f\}$
                \Comment{Replace weaker, overlapping motifs}
            \State \textbf{continue}
        \EndIf
    \EndIf
    \State $H \gets H \cup \operatorname{feedback}(f)$
\EndFor
\end{algorithmic}
\end{algorithm}

We run the iterative loop (Algorithm~\ref{fig:motif-discovery}) for 100 steps using each model's maximum reasoning setting, a 61,440-token reasoning budget, and a 4,096-token visible-output budget. Additional details, including the full prompts, are given in \S~\ref{appdx:prompt}.

\paragraph{Explaining Specific Moves.}
\label{sec:explaining_specifics}

Given an initial position, and a move we want to explain, we use the process detailed in \S \ref{sec:evaluation_explanations} to generate a set of positive and negative positions. Then, our goal is to find the most relevant motifs that explain the differences in positive and negative positions. For each motif, we ask two questions. First: how useful is the motif in general? This is measured by \(\Delta(f)\): motifs with larger \(\Delta(f)\) appear much more often in winning positions than losing ones. Second: how relevant is the motif for this particular move? We measure how often the motif is true on the given positives and how often it is false on the given negatives. Fit is the lower of these two percentages. We rank motifs by combining these two quantities, \(\Delta(f) \cdot \text{fit}(f)\), and return the top \(k\). Lastly, an additional LM call converts the motifs into natural language (\S~\ref{appdx:transcription}). 

\section{Results}

We evaluate four open-weight models, \texttt{Qwen3.5-397B-A17B} \citep{qwen2026qwen35}, \texttt{Kimi-K3} \citep{kimiteam2026k3}, \texttt{GLM-5.2} \citep{zeng2026glm}, and \texttt{DeepSeek-V4-Pro} \citep{deepseekv4pro}, and four proprietary models, \texttt{GPT-5.5} \citep{openai2026gpt55}, \texttt{GPT-5.6 Sol} \citep{openai2026gpt56}, \texttt{Opus 5} \citep{anthropic2026opus5}, and \texttt{Fable 5} \citep{anthropic2026fable5}. For each model, we compare direct prompting (Baseline) with BinProp. For each position, Baseline is prompted to propose three motifs, while BinProp selects three from its discovered set. Both methods follow the same evaluation protocol. For each position in the test set, we score up to three motifs with at least \(50\%\) support fit. The position's score is the arithmetic mean of these motif scores. If none qualifies, we score the highest-ranked valid motif; if none is valid, the position scores zero. Table~\ref{tab:chess_results} reports the average across all 100 test positions and shows that BinProp improves explainability for all eight models (bootstrap intervals in \S~\ref{app:bootstrap}). Even the lowest BinProp score, \(0.174\), exceeds the strongest baseline, \(0.173\); the highest BinProp score is \(0.296\). \begin{table}[ht]
\centering
\caption{Chess test results. Gain is BinProp minus Baseline, computed before rounding.}
\label{tab:chess_results}
\begin{tabular}{llcc}
\toprule
Model & Method & Mean Explainability & Gain \\
\midrule
\multirow{2}{*}{Qwen3.5-397B-A17B \citep{qwen2026qwen35}}
  & Baseline & 0.102 & \multirow{2}{*}{\textcolor{green!55!black}{+0.072}} \\
  & BinProp (ours)       & 0.174 & \\
\addlinespace[0.2em]
\multirow{2}{*}{GPT-5.6 Sol \citep{openai2026gpt56}}
  & Baseline & 0.062 & \multirow{2}{*}{\textcolor{green!55!black}{+0.144}} \\
  & BinProp (ours)       & 0.205 & \\
\addlinespace[0.2em]
\multirow{2}{*}{DeepSeek-V4-Pro \citep{deepseekv4pro}}
  & Baseline & 0.173 & \multirow{2}{*}{\textcolor{green!55!black}{+0.042}} \\
  & BinProp (ours)       & 0.214 & \\
\addlinespace[0.2em]
\multirow{2}{*}{Kimi-K3 \citep{kimiteam2026k3}}
  & Baseline & 0.079 & \multirow{2}{*}{\textcolor{green!55!black}{+0.136}} \\
  & BinProp (ours)       & 0.216 & \\
\addlinespace[0.2em]
\multirow{2}{*}{GLM-5.2 \citep{zeng2026glm}}
  & Baseline & 0.051 & \multirow{2}{*}{\textcolor{green!55!black}{+0.182}} \\
  & BinProp (ours)       & 0.233 & \\
\addlinespace[0.2em]
\multirow{2}{*}{GPT-5.5 \citep{openai2026gpt55}}
  & Baseline & 0.085 & \multirow{2}{*}{\textcolor{green!55!black}{+0.169}} \\
  & BinProp (ours)       & 0.254 & \\
\addlinespace[0.2em]
\multirow{2}{*}{Opus 5 \citep{anthropic2026opus5}}
  & Baseline & 0.049 & \multirow{2}{*}{\textcolor{green!55!black}{+0.242}} \\
  & BinProp (ours)       & 0.291 & \\
\addlinespace[0.2em]
\multirow{2}{*}{Fable 5 \citep{anthropic2026fable5}}
  & Baseline & 0.031 & \multirow{2}{*}{\textcolor{green!55!black}{+0.265}} \\
  & BinProp (ours)       & \textbf{0.296} & \\
\bottomrule
\end{tabular}
\end{table}

\begin{figure*}[t!]
    \centering
    \begin{minipage}[t]{0.463125\linewidth}
        \vspace{0pt}
        \includegraphics[width=\linewidth]{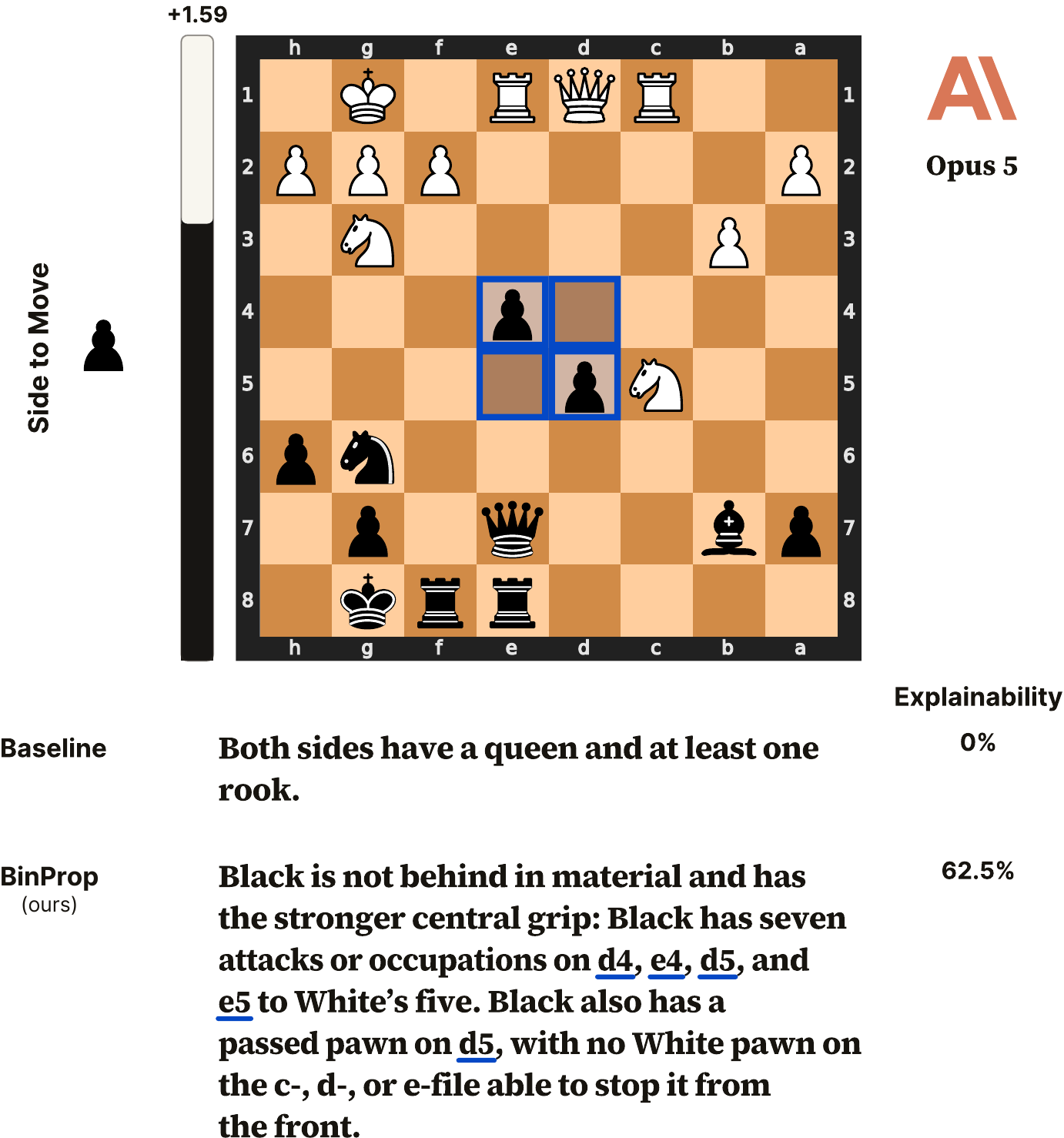}
    \end{minipage}\hfill
    \begin{minipage}[t]{0.463125\linewidth}
        \vspace{0pt}
        \includegraphics[width=\linewidth]{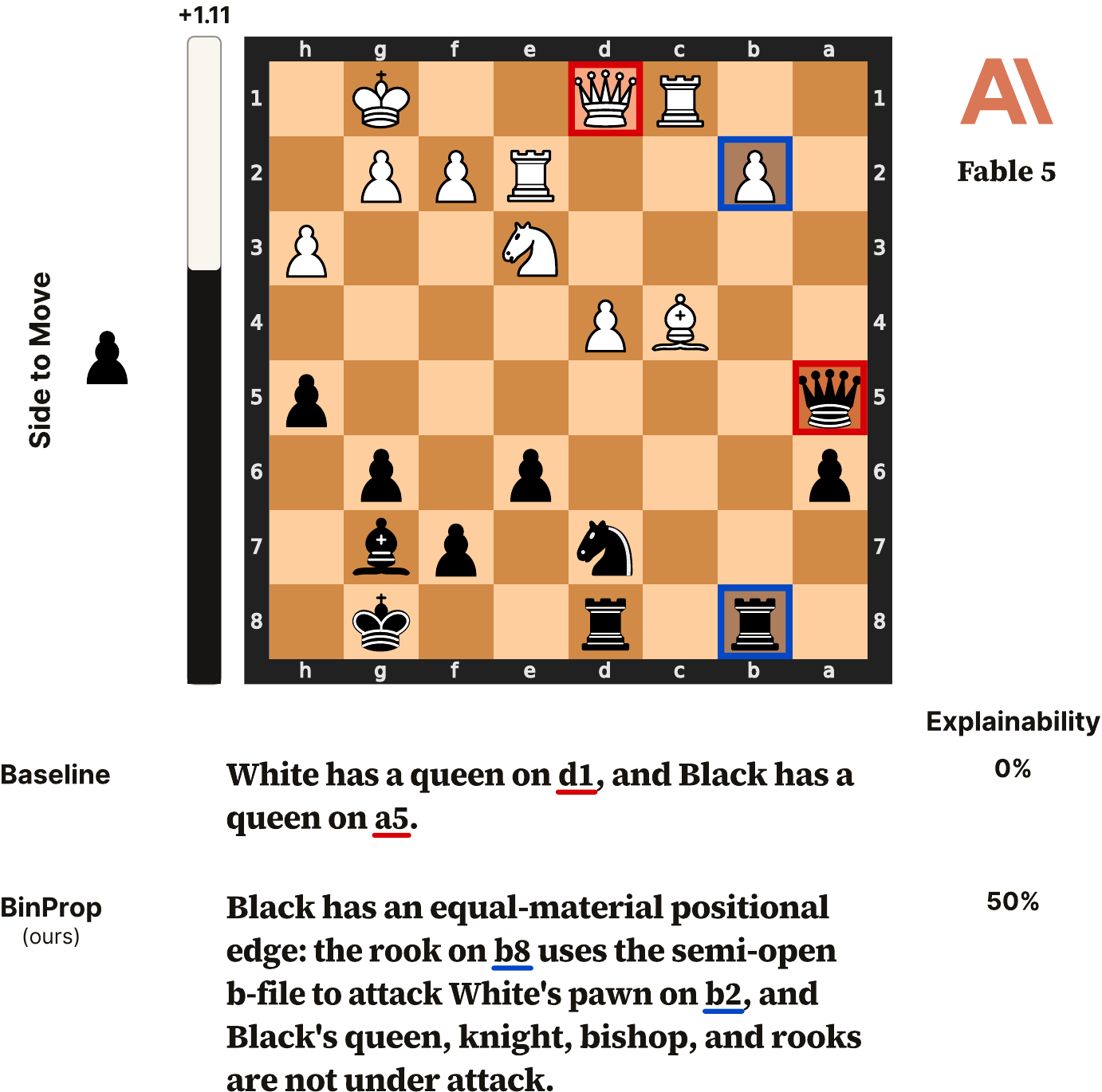}
    \end{minipage}
    \vspace{0.1em}

    \begin{minipage}[t]{0.463125\linewidth}
        \vspace{0pt}
        \includegraphics[width=\linewidth]{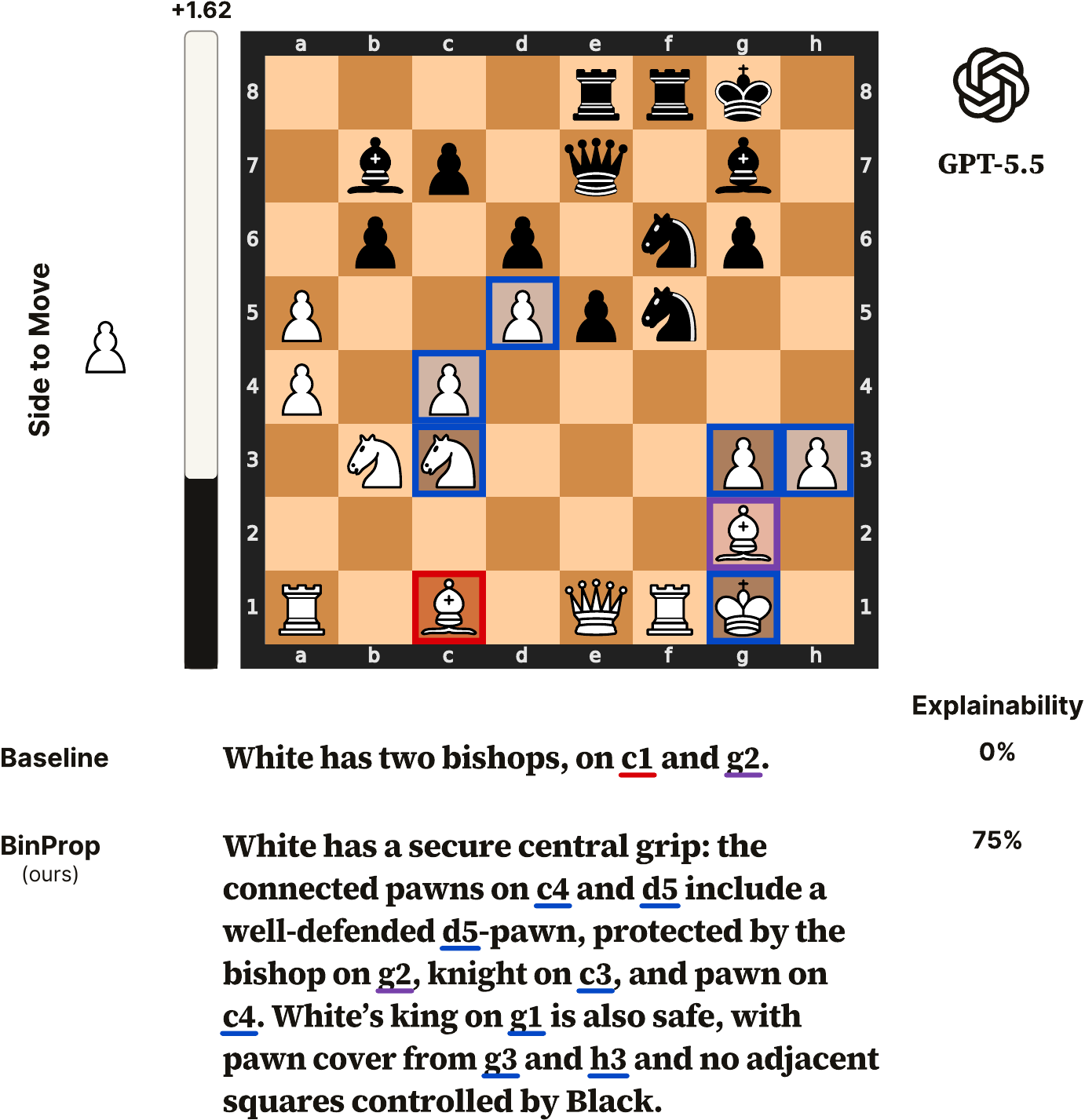}
    \end{minipage}\hfill
    \begin{minipage}[t]{0.463125\linewidth}
        \vspace{0pt}
        \includegraphics[width=\linewidth]{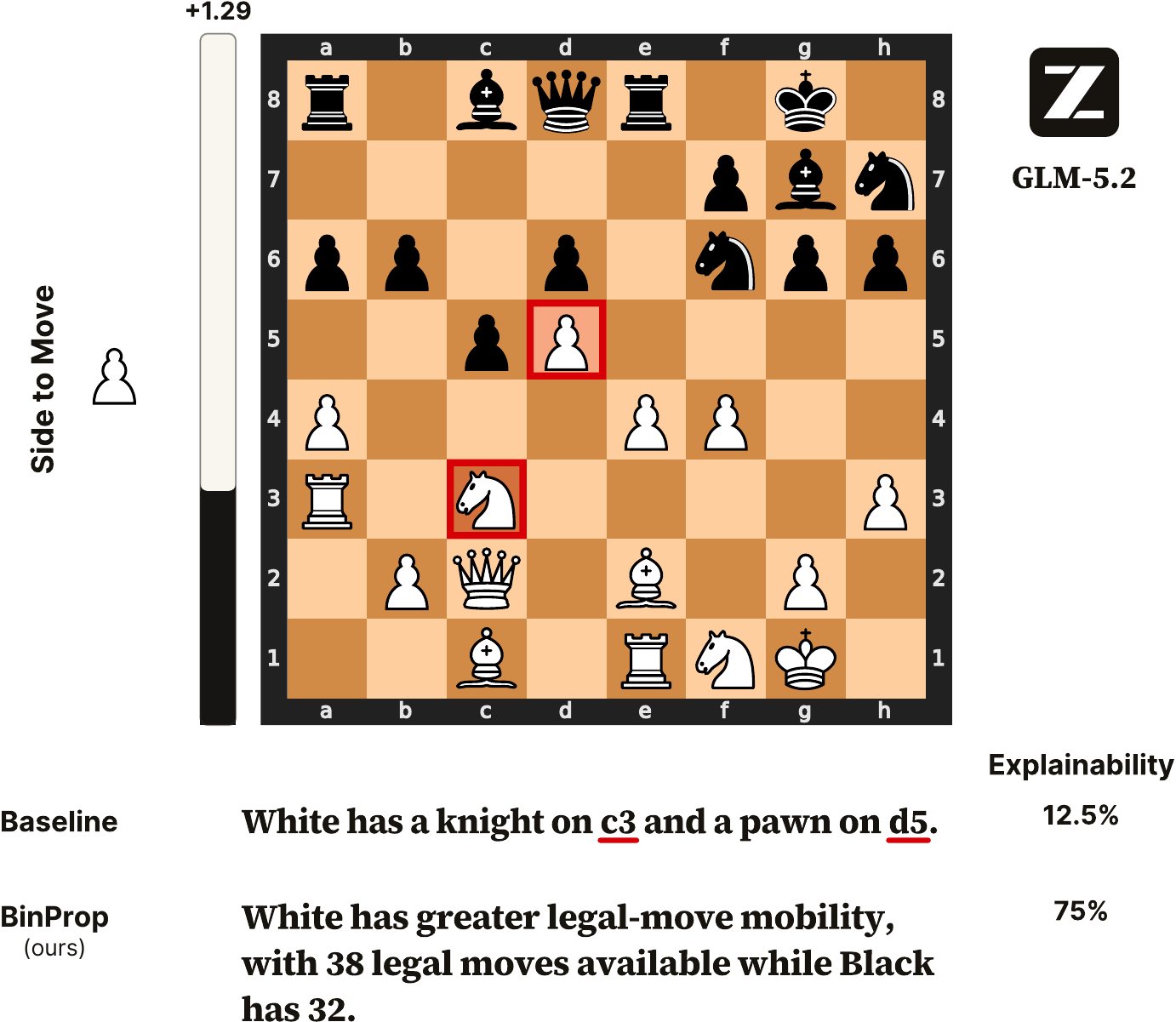}
    \end{minipage}
    \caption{Baseline and BinProp explanations for the top four models. Each panel shows one sampled positive position; explainability is measured over 24 held-out positions: eight positives, eight negatives, and eight nearby negatives. Evaluation bars show Stockfish's evaluation.}
    \label{fig:direct_bp_examples}
    
\end{figure*}

\paragraph{Baseline models struggle to generalize.} Even when prompted to identify general chess concepts, direct prompting mostly yields motifs that rely on shallow board details. In Fig.~\ref{fig:direct_bp_examples}, the \texttt{Opus 5} motif requires only that each side has a queen and at least one rook; \texttt{Fable 5} checks for White's queen on d1 and Black's on a5; and \texttt{GPT-5.5} checks for White's bishops on c1 and g2. All three fit the support examples but also hold on every nearby negative, so each scores zero. Still, a minority of baseline runs produce non-brittle motifs: one \texttt{Opus 5} motif requires a White pawn on the fifth rank or beyond and no Black bishop or knight in White's half. Another, from \texttt{GPT-5.5}, requires an advanced White bishop to attack a square next to Black's king, both achieving an explainability score of \(87.5\%\) .

\paragraph{BinProp.} BinProp discovers 192 motifs (\S~\ref{app:chess-motif-catalog}), ranging from 11 for \texttt{Kimi-K3} to 44 for \texttt{GPT-5.5}. Median \(\Delta(f)\) ranges from \(2.5\%\) for \texttt{Fable 5} to \(9.0\%\) for \texttt{Qwen}. The models also differ in how many chess ideas they combine within a motif. Most favor a single idea: 20 of \texttt{Opus 5}'s 30 motifs each focus on a single chess concept, such as material advantage or dangerous passed pawns. In contrast, \texttt{Fable 5} and \texttt{GPT-5.5} combine several ideas in 19 of 20 and 37 of 44 motifs, respectively. Such motifs can hold for different reasons in different positions. Fable's \emph{concrete target} motif, for example, can identify profitable captures, weak-pawn pressure, advanced passed pawns, or king attacks, subject to material and piece-safety conditions. Another example is \texttt{GPT-5.5}'s \emph{Material-parity forcing gain or mobile bishop pair}. 

\paragraph{Common Motifs.} Material advantage appears in most model motif sets. \texttt{Qwen}, \texttt{GPT-5.6 Sol}, and \texttt{Opus 5} require at least a pawn advantage; \texttt{GLM-5.2} requires two; \texttt{Kimi-K3} accepts a material lead or, at equal material, more attacks on enemy pieces. \texttt{GPT-5.5} requires a one-pawn material advantage, counting a \(0.4\)-pawn bonus for having two bishops. The bishop pair also appears in most motif sets: \texttt{Qwen}, \texttt{GLM-5.2}, \texttt{Kimi-K3}, and \texttt{DeepSeek-V4-Pro} require two bishops against at most one; \texttt{GPT-5.6 Sol} adds equal material and minor piece counts; \texttt{Fable 5} adds safe pieces and king; and \texttt{Opus 5} and \texttt{GPT-5.5} include it in broader motifs. 

\paragraph{Unique Motifs.}We measure uniqueness by applying motifs to the winning set and finding each motif's largest containment overlap with any motif from another model; lower overlap means more unique. 
Among motifs from \texttt{Opus 5}, \texttt{Fable 5}, \texttt{GPT-5.5}, and \texttt{GPT-5.6 Sol} with \(\Delta(f)\geq5\%\), the two most unique both come from \texttt{Opus 5}: \emph{Opponent pieces tied to passive defence} finds pieces forced to defend or block passed pawns, while \emph{Opponent saddled with a bad minor piece} finds a knight with few safe squares or a bishop trapped by its own pawns. Among these four models, the most unique accepted motif (\(\Delta(f)>1\%\)) is also from \texttt{Opus 5}: \emph{Line pressure with pins and x-rays} (\(\Delta=1.16\%\)), which compares the two sides' pins, x-rays, batteries, and potential discovered attacks.

\paragraph{Motif Selection.}
With standard selection, Baseline scores \(0.079\) and BinProp scores \(0.235\) (Table~\ref{tab:chess_results}). Selecting the highest-scoring motif among each model's up to three already-selected motifs, using test labels, gives \(0.164\) and \(0.356\). Choosing the best of these motifs across all eight models gives \(0.429\) and \(0.660\). Both oracles use the maximum motif score rather than the mean. The cross-model BinProp oracle (\(0.660\)) more than doubles the best individual BinProp result (\(0.296\)), suggesting room for further improvements through better ranking, stronger motifs, or both.

\section{Analysis}

\subsection{Explainability Tracks Objective Move Quality}
\label{sec:explainability-playing-strength}

\begin{wrapfigure}[20]{r}{0.475\textwidth}
    \centering
    \includegraphics[width=\linewidth]{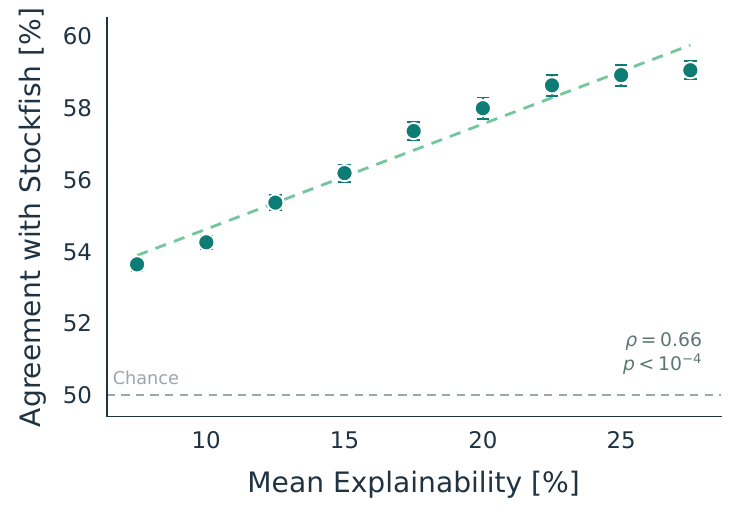}
    \vspace{-1em}
    \caption{Mean explainability versus Stockfish agreement. Points average independently sampled motif sets; error bars show 95\% confidence intervals.}
    \label{fig:explainability-performance}
\end{wrapfigure}

Does a higher explainability score correspond to better chess knowledge? If better explanations capture more useful chess concepts, those concepts should also help choose better moves. To test this, we randomly sample motif sets with varying explainability scores and use each set as features in a simple position evaluator. If explainability is indeed tied to objective chess strength, motif sets with higher explainability scores should agree more often with Stockfish.

We use motifs from the four strongest BinProp runs: \texttt{Fable 5}, \texttt{Opus 5}, \texttt{GPT-5.5}, and \texttt{GLM-5.2}. We repeatedly sample sets of 10 motifs whose mean explainability ranges from \(7.5\%\) to \(27.5\%\). Each set defines a 10-feature evaluator: feature \(j\) is one when motif \(j\) holds and zero otherwise. We fit a regularized linear model on the 200k winning and losing positions to predict Stockfish's centipawn evaluation, learning one weight per motif. The test set contains 100 starting positions with eight unseen continuation pairs each. Both continuations in a pair begin from the same position: one follows Stockfish's best move, while the other follows a lower-evaluated alternative. The evaluator agrees with Stockfish when it assigns the best continuation a higher score. As shown in Fig.~\ref{fig:explainability-performance}, explainability strongly correlates with Stockfish agreement (\(\rho=0.66\), \(p<10^{-4}\)).

\subsection{Comparison to Concept-Guided Chess Commentary}
\label{sec:ccc-comparison}

Concept-Guided Chess Commentary (CCC) \citep{kim2025bridging} is the closest reproducible prior method. It ranks 21 LeelaChessZero \citep{lc0} concept probes and conditions commentary on the top five.

We use \texttt{Fable 5}, the highest-scoring BinProp model, at maximum reasoning and give CCC the same eight support boards as the baseline and BinProp. For evaluation, \texttt{Fable 5} translates each commentary into up to three Boolean Python functions. Two additional \texttt{Fable 5} calls check that converting the commentary into code preserves its meaning. Functions approved by both checks undergo the same 50\% support-fit filter and scoring as BinProp. If no working function passes both checks, the commentary scores zero. On the test set, CCC scores \(0.051\), compared with \(0.031\) for the baseline and \(0.296\) for BinProp. Figure~\ref{fig:ccc-bp-example} shows example explanations from CCC and BinProp.

\afterpage{%
\begin{figure}[!t]
    \centering
    \includegraphics[width=0.855\linewidth,trim=0 18bp 0 3bp,clip]{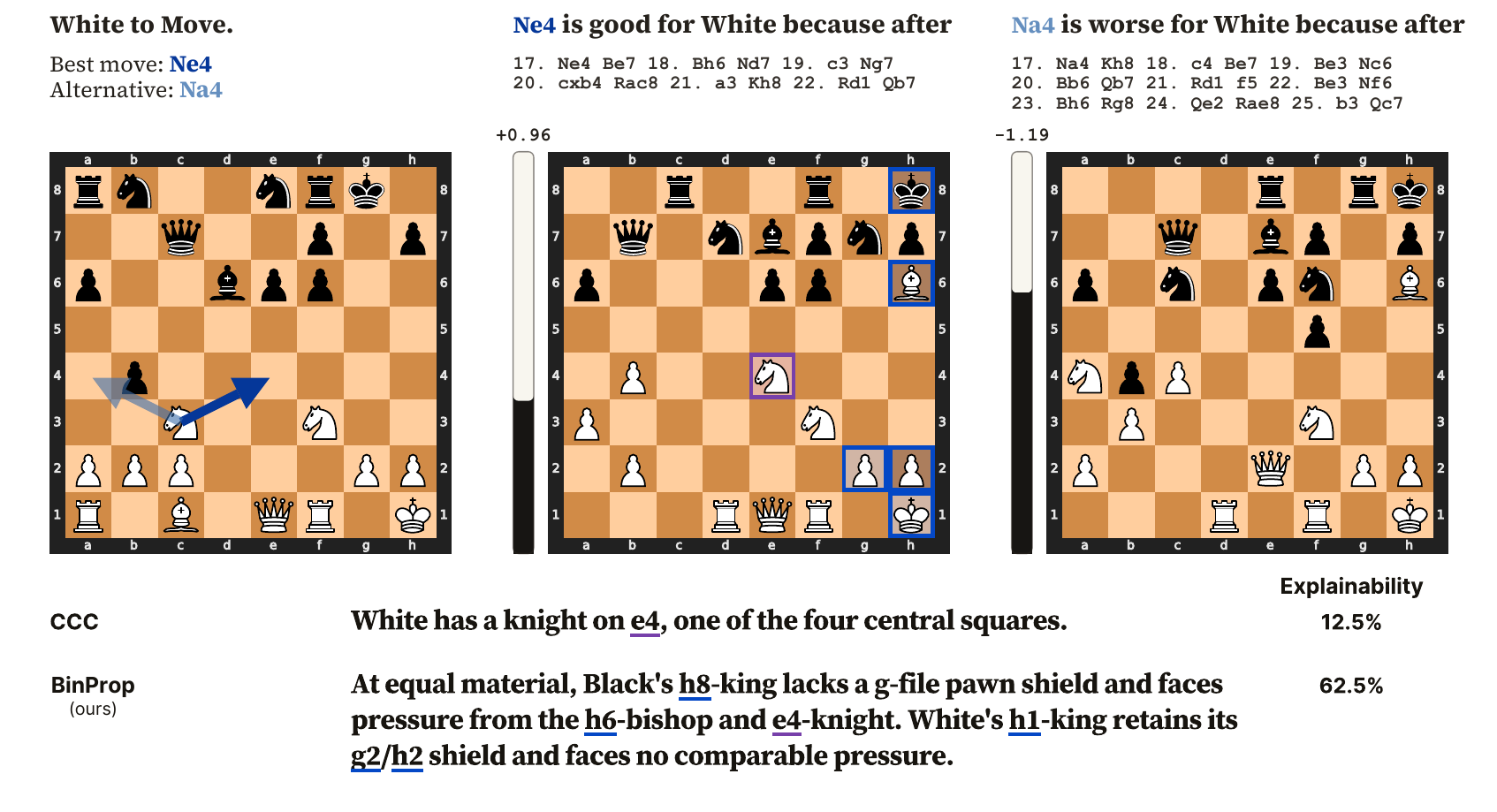}
    \caption{CCC and BinProp explanations from \texttt{Fable 5} for \texttt{Ne4} in Karpov--Ermenkov (1980). CCC identifies a centralized knight; BinProp (ours) combines material balance, king safety, and attacking pressure, earning a much higher explainability score.}
    \label{fig:ccc-bp-example}
\end{figure}
\afterpage{%
\begin{figure}[!t]
    
    \centering
    \begin{minipage}{0.81225\linewidth}
    \begin{minipage}[t]{0.49\linewidth}
        \vspace{0pt}
        \includegraphics[width=\linewidth]{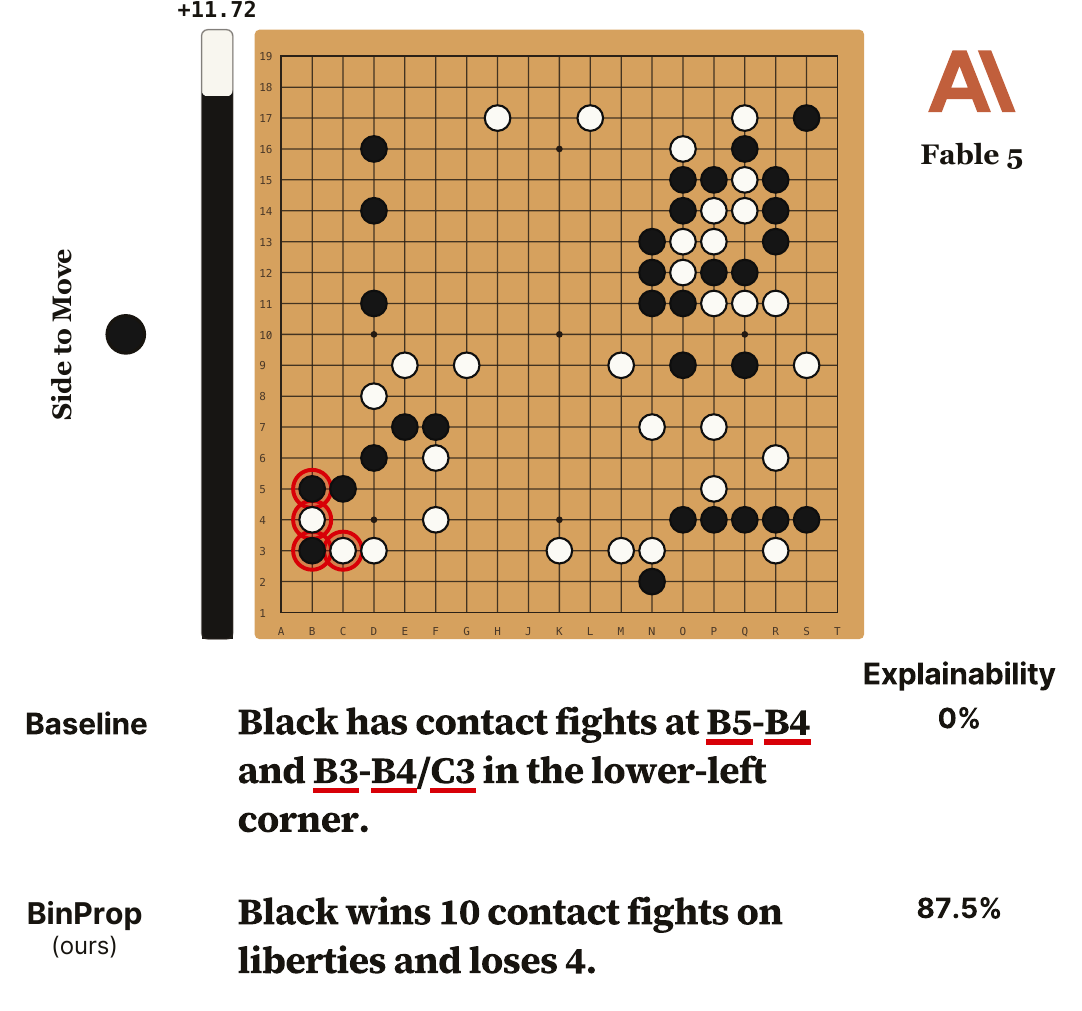}
    \end{minipage}\hfill
    \begin{minipage}[t]{0.49\linewidth}
        \vspace{0pt}
        \includegraphics[width=\linewidth]{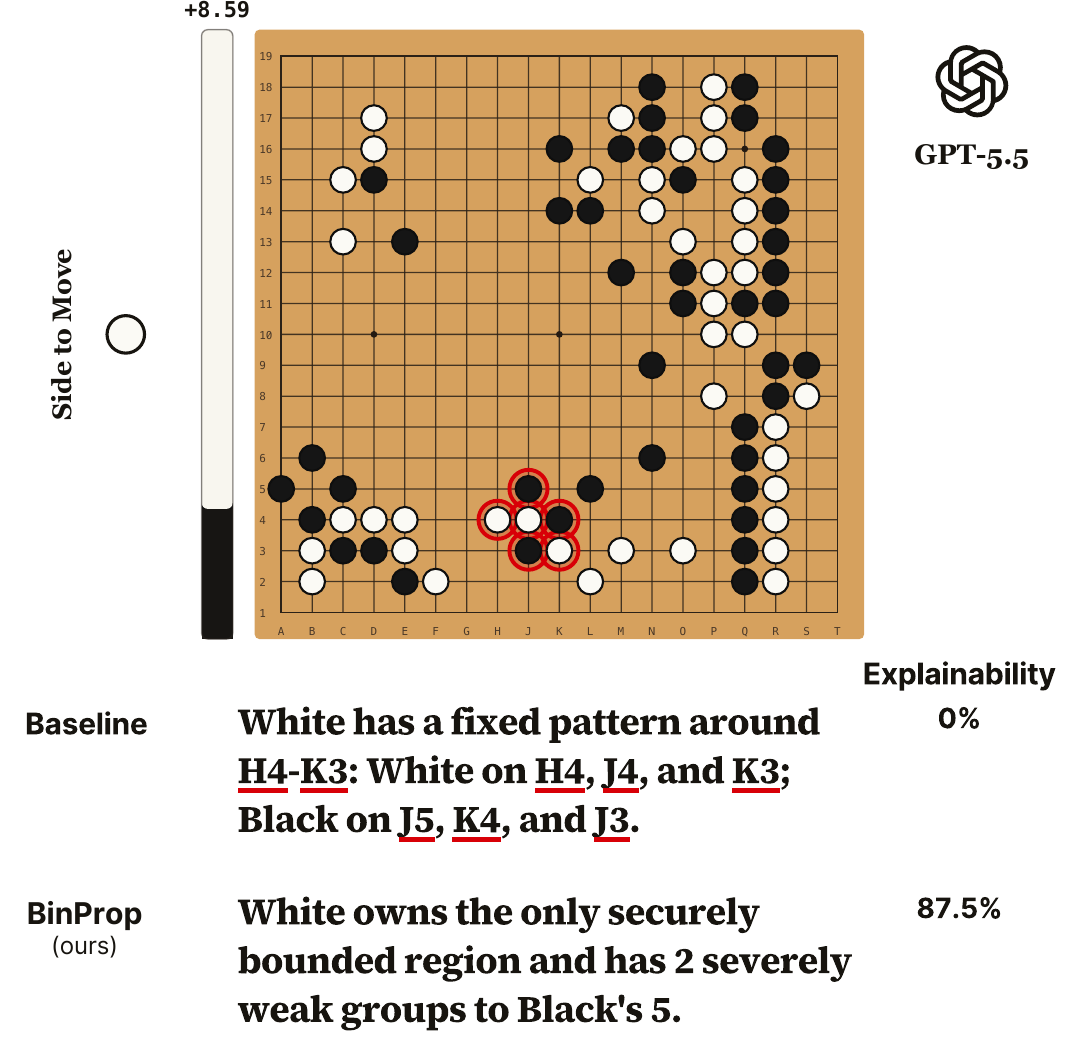}
    \end{minipage}
    \end{minipage}
    \caption{Baseline and BinProp explanations for the two highest-scoring Go models. Each panel shows one positive position. Explainability is measured over 24 held-out positions: eight positives, eight negatives, and eight nearby negatives. Bars show KataGo evaluations.}
    \label{fig:go-examples}
\end{figure}
}
}

\subsection{Using BinProp to Explain Go, Draughts, and Shogi}

We next ask whether BinProp extends beyond chess. We present our experiments on Go here, on draughts in \S~\ref{sec:appendix-checkers}, and on shogi in \S~\ref{sec:appendix-shogi}. For Go, we use 100k winning and 100k losing positions from JGDB~\citep{jgdb}, scored by KataGo~\citep{wu2019accelerating}, and evaluate on 100 held-out expert-game positions with positive, negative, and nearby-negative continuations (see \S~\ref{appdx:go_data_collection}). BinProp retains its chess discovery settings; engines, score bands, and continuation sampling are game-specific.

On Go, BinProp improves explainability scores for all eight models (Table~\ref{tab:go_results_full}). Like in chess, the few-shot baseline often produces brittle, coordinate-specific motifs: one \texttt{Qwen} motif requires three Black stones on a fixed board segment, while one \texttt{GLM-5.2} motif examines only a fixed \(3\times3\) region. BinProp instead finds reusable Go concepts (Fig.~\ref{fig:go-examples}). \texttt{Kimi-K3}'s \emph{local stone density territory} estimates territory from nearby stones, while \texttt{Opus 5}'s \emph{multi-scale box-majority area lead} combines local influence at two scales. Beyond territory, BinProp finds motifs about the risk of losing groups of stones. \texttt{GPT-5.5}'s \emph{net low-liberty attack swing} compares threatened and defended groups, while \texttt{GPT-5.6 Sol}'s \emph{low-liberty burden advantage} compares how many stones each side has in groups with only one or two liberties.

\begin{table}[H]
\centering
\caption{Mean Go test explainability for the top four BinProp models. Full results are in Table~\ref{tab:go_results_full}.}
\label{tab:go_results}
\small
\renewcommand{\arraystretch}{0.93}
\begin{tabular}{llcc}
\toprule
Model & Method & Mean Explainability & Gain \\
\midrule
\multirow{2}{*}{GPT-5.6 Sol \citep{openai2026gpt56}}
  & Baseline & 0.245 & \multirow{2}{*}{\textcolor{green!55!black}{+0.055}} \\
  & BinProp (ours)       & 0.300 & \\
\addlinespace[0.1em]
\multirow{2}{*}{Opus 5 \citep{anthropic2026opus5}}
  & Baseline & 0.190 & \multirow{2}{*}{\textcolor{green!55!black}{+0.111}} \\
  & BinProp (ours)       & 0.301 & \\
\addlinespace[0.1em]
\multirow{2}{*}{GPT-5.5 \citep{openai2026gpt55}}
  & Baseline & 0.181 & \multirow{2}{*}{\textcolor{green!55!black}{+0.122}} \\
  & BinProp (ours)       & 0.303 & \\
\addlinespace[0.1em]
\multirow{2}{*}{Fable 5 \citep{anthropic2026fable5}}
  & Baseline & 0.159 & \multirow{2}{*}{\textcolor{green!55!black}{+0.155}} \\
  & BinProp (ours)       & \textbf{0.314} & \\
\bottomrule
\end{tabular}
\end{table}

\section{Related Work}

\paragraph{Language-Based Model Explanations.}
Prior work explains classifier outputs with textual or multimodal justifications~\citep{hendricks2016generating,park2018multimodal}, counterexamples or absent features~\citep{gulshad2020explaining,dhurandhar2018explanations}, and sensitivity to user-defined concepts~\citep{kim2018interpretability}. Natural-language explanations can also provide executable supervision. BabbleLabble~\citep{hancock2018training} lets a user explain why a training example has a label, then converts that explanation into a program that labels new examples. For instance, the phrase \emph{due to} appearing between two entities becomes a rule for recognizing a causal relation. Related methods jointly learn language interpretation and classification or generate Python labeling functions with text-to-text models~\citep{srivastava2017joint,papanikolaou2021teach}. Anchors finds short conditions under which a classifier almost always keeps the same prediction, while LORE also describes nearby changes that would flip it~\citep{ribeiro2018anchors,guidotti2018local}.

\paragraph{Contrastive Concept Discovery.}
Closest in structure to our work are methods that propose and score natural-language differences between groups of samples. This pattern has been used for text distributions and goal-directed corpus analysis~\citep{zhong2022differences,zhong2023goal}, image sets~\citep{dunlap2024visdiff}, and language-model outputs~\citep{dunlap2025vibecheck}. A language model can also evolve class-specific visual attributes, retaining those that improve a CLIP classifier~\citep{chiquier2024evolving}. NLParam iteratively builds statistical models whose features are yes/no natural-language statements about individual examples~\citep{zhong2024datasets}. VidDiff and ADIFF instead describe fine-grained differences between paired videos or audio recordings~\citep{burgess2025viddiff,deshmukh2025adiff}. BinProp likewise proposes and tests descriptions, but compiles each one into a deterministic board function.

\paragraph{Explaining Chess.}
\citet{lee2022improving} generate commentary from an engine's evaluation of move quality and its suggested continuations. \citet{cui2026communicating} turn selected engine lines into written plans and test whether humans and language models can use them against different replies. In concurrent work, ACT-Eval decomposes commentary into atomic claims, checks factual claims with chess tools, and measures coverage against expert annotations~\citep{hebbar2026hallucinations}. BinProp automatically discovers motifs as Python functions and tests them on engine continuations.

\citet{mcgrath2022acquisition} train simple linear models to detect human-defined chess concepts, such as material advantage and the bishop pair, in AlphaZero's internal activations. They also look for concepts without predefined labels using matrix factorisation and correlations between activations and board features. This reveals patterns such as which squares an opponent's pieces can reach, although most factors in later layers remain uninterpreted. \citet{schut2025bridging} compare AlphaZero's internal representations along strong and weak search lines to find features that separate them. They keep features that a student agent can learn and that appear more clearly in AlphaZero games than in human games. They then show example positions to four grandmasters and test whether the players can infer these concepts. BinProp requires no engine activations or training: it iteratively discovers concepts that are associated more with winning positions than losing ones.

\section{Conclusion and Future Work}
We introduce BinProp, a training-free method for finding natural-language explanations for chess, Go, draughts, and shogi moves. BinProp iteratively finds concepts that appear more often in winning positions than in losing ones. Beyond games, BinProp could explain agent rollouts spanning hundreds or thousands of steps~\citep{press2025algotune,yang2026programbenchlanguagemodelsrebuild,datacurve2026deepswe,feng2026longcli}, showing why one trajectory succeeds and a similar one fails when manual review is impractical.

\bibliographystyle{references}
\bibliography{bib}

\newpage
\appendix

\section{Iterative Prompting Harness}
\label{appdx:prompt}

BinProp runs for 100 completed iterations, with one proposed motif per iteration.  Every call uses the provider's maximum reasoning setting, a 65,536-token total output cap, temperature \(1\), and top-\(p\) \(1\). Where the provider exposes separate budgets, we allocate 61,440 tokens to reasoning and 4,096 to the visible answer. The model receives one message containing fixed instructions, the 12 most recent model and evaluator messages, the current motif set ranked by \(\Delta(f)\), up to 12 examples from buckets that the current set does not separate, and four random positions from each of the winning and losing sets. Candidates are first screened on 20,000 positions per label and then scored on all 200,000 positions. A 512-position speed check requires a 95th-percentile runtime of at most 2\,ms per position.

\subsection{Prompt}

\begin{systembubble}
\begin{lstlisting}[style=macPython]
<instructions>
We have one bucket of winning chess positions and one comparison
bucket; the current prompt names the exact labels.
Your task is to build Python functions. Each function should return
True more often on winning positions than on comparison positions, and
should not overlap too much with existing functions.

Each position is a FEN; the FEN side to move is always the player
being evaluated.

This is an iterative loop: propose code, the evaluator compiles it,
runs it, checks the dataset, then gives concrete feedback.
Use evaluator feedback as ground truth. Do not repeat failures from
the transcript.

Return Python code every turn with all three required items:
1. MOTIF_NAME = 'descriptive motif name'
2. MOTIF_DESCRIPTION = 'one clear sentence'
3. def motif(fen: str) -> bool:

Do not include self-test generators. The evaluator checks your motif
on the dataset.

Use one fenced Python code block. ```python and plain ``` fences are
both accepted.

Rules:
- You may import chess and math only.
- True should mean the side to move has the chess feature your
  function checks.
- Use board.turn for the side to move and not board.turn for the
  opponent.
- The function must still make sense after swapping White and Black:
  use side-to-move ideas, not White-only or Black-only ideas.
- Keep code short: no docstrings, minimal comments, and total source
  under 12000 characters.
- No engines, files, network access, randomness, global mutable state,
  recursion, classes, or unsafe dynamic calls.
- while loops and try/except are allowed only when finite and fast;
  smoke checks time out after about 1 second.

python-chess API cheat sheet:
- board = chess.Board(fen); stm = board.turn; opp = not stm.
- board.pieces(chess.KNIGHT, stm) returns the matching squares.
- board.piece_at(sq) returns a Piece or None; use piece.piece_type and
  piece.color.
- board.attacks(sq) returns attacked squares from the piece on sq;
  board.attackers(color, sq) returns squares of color attacking sq.
- Use chess.square_file(sq), chess.square_rank(sq),
  chess.square(file, rank), and chess.square_distance(a, b).
- There is no chess.BitBoard, Board.is_defended_by, or
  Board.square_color; use board.attackers(color, sq) and
  (file+rank)%2.
- board.is_capture(move) takes a legal Move, not a target square.
- No prose outside Python code block(s).
</instructions>

<history>
Model:
Proposal source omitted for brevity.

candidate=level-material king safety edge
status=scored

System:
Evaluator feedback for your previous proposal:
candidate = level-material king safety edge
status = scored
P(True|winning) = 0.050
P(True|losing) = 0.007
delta = +0.043
reason = accepted
</history>

<current>
Your answer must include one Python code block containing:

MOTIF_NAME = 'descriptive motif name'
MOTIF_DESCRIPTION = 'one clear sentence'
def motif(fen: str) -> bool:
    ...

Current iteration: 23

Winning bucket:
eval_cp_stm is +125 to +250cp, so the side to move is clearly better.

Losing bucket:
eval_cp_stm is -250 to -125cp, so the side to move is clearly worse.

Acceptance:
- delta must be at least 0.010;
- winning-set containment must be below 0.700, unless the candidate
  has higher delta and replaces every overlapping motif;
- the function must compile, run, and satisfy the speed limit.

Current motifs ranked by winning-minus-comparison delta:
1. Net material advantage:
   delta=0.5605
   desc=The side to move has a higher total material value than the
   opponent.
2. Healthy mobile wing majority:
   delta=0.1766
   desc=With balanced non-pawn forces and no advanced passer, the
   side to move has a healthier connected wing majority.
3. Level-material mobility dominance:
   delta=0.0904
   desc=With material values level, the side to move has substantially
   greater effective mobility for its non-pawn pieces.
...

Prompt hints:
Mixed buckets where the current motifs do not separate winning and
comparison positions:
...

If you need inspiration, here are four randomly sampled winning
positions and four randomly sampled losing positions. Do not
hard-code them. Acceptance still depends on the full dataset.
...
</current>
\end{lstlisting}
\end{systembubble}

The listing shows the chess specialization; FEN examples are elided. Go uses the same history, ranking, hints, acceptance test, and feedback loop. Only the position contract and API hints change:

\begin{systembubble}
\begin{lstlisting}[style=macPython]
Each position is a 19x19 board. The evaluated side is passed as color,
either 'B' or 'W'.

def motif(board_rows: list[str], color: str) -> bool:

- You may import math or from sgfmill import boards only.
- True should mean the side named by color has the Go feature.
- board_rows is 19 strings of length 19 using '.', 'B', and 'W'.
- Use color for the evaluated side and the opposite color for the
  opponent; never assume the evaluated side is always B.
- Use orthogonal neighbors, with explicit board bounds, for groups
  and liberties.
\end{lstlisting}
\end{systembubble}

For Go, the winning bucket has a \(+7.5\) to \(+15.0\)-point score from the evaluated side's perspective, while the comparison bucket has a \(-15.0\) to \(-7.5\)-point score.

\subsection{Proposal Format}

A valid proposal contains a name, a one-sentence description, and one side-to-move function.

\begin{agentbubble}
\begin{lstlisting}[style=macPython]
import chess

MOTIF_NAME = 'Net material advantage'
MOTIF_DESCRIPTION = (
    'The side to move has a higher total material value '
    'than the opponent.'
)

def motif(fen: str) -> bool:
    board = chess.Board(fen)
    stm = board.turn
    values = {
        chess.PAWN: 1,
        chess.KNIGHT: 3,
        chess.BISHOP: 3,
        chess.ROOK: 5,
        chess.QUEEN: 9,
    }
    own = sum(
        values[p] * len(board.pieces(p, stm))
        for p in values
    )
    opp = sum(
        values[p] * len(board.pieces(p, not stm))
        for p in values
    )
    return own > opp
\end{lstlisting}
\end{agentbubble}

\begin{systembubble}
\begin{lstlisting}[style=macPython]
candidate = "Net material advantage"
status = "scored"

P(candidate=True | winning) = 0.615
P(candidate=True | losing) = 0.055
delta = +0.561

reason = "accepted"
\end{lstlisting}
\end{systembubble}

\subsection{Low-Separation Feedback}

When \(\Delta(f)<0.01\), the system reports the activation rate on the winning set and the false-positive rate on the losing set, as well as three losing positions where the candidate incorrectly returned \texttt{True}. Each example includes the FEN and up to 25 executed source lines.

\begin{systembubble}
Rejected: not enough winning-vs-losing gap.

\begin{lstlisting}[style=macPython]
candidate = "Material-rebalancing tactical shot"

P(candidate=True | winning) = 0.043
P(candidate=True | losing) = 0.061

delta = -0.018
required >= 0.010
\end{lstlisting}

Losing position 1:
\begin{lstlisting}[style=macPython]
fen = "r2q1rk1/1pp1bppp/1p2p1b1/4P3/3P4/P1N1nN2/1P2QKPP/2R4R w - - 0 17"
color = "white"
label = "losing"
eval_cp_stm = -220
\end{lstlisting}

Candidate trace:
\begin{lstlisting}[style=macPython]
L039>> hit = False
L040>> if is_capture:
L041>>     gain = cv - av if not safe else cv + bonus
L042>>     favorable = cv > av or (cv >= 3 and safe)
L043>>     hit = favorable and balance + gain >= -2
...
L061>> board.pop()
L062>> if hit:
L063>>     return True
\end{lstlisting}

Losing position 2:
\begin{lstlisting}[style=macPython]
fen = "r4rk1/3N1ppp/bn3q2/p7/2pPPB2/2P2P2/6PP/RQ2KB1R b KQ - 0 18"
color = "black"
label = "losing"
eval_cp_stm = -217
\end{lstlisting}

Losing position 3:
\begin{lstlisting}[style=macPython]
fen = "2kr1bnr/1b4pp/p3p3/1pq5/P4N2/3p2P1/1P1P1P1P/R1B1QRK1 w - - 0 17"
color = "white"
label = "losing"
eval_cp_stm = -203
\end{lstlisting}
\end{systembubble}

\subsection{Overlap Feedback}

The following candidate is fully contained in \texttt{Net material advantage} on its winning activations. Because its \(\Delta\) is lower, it cannot replace the existing motif.

\begin{systembubble}
Rejected: winning-set overlap is too high and the candidate has lower separation.

\begin{lstlisting}[style=macPython]
candidate = "Compensated non-pawn force surplus"
existing = "Net material advantage"

candidate True on winning = 10870
existing True on winning = 61523
both True on winning = 10870

P(existing=True | candidate=True, winning) = 1.000
P(candidate=True | existing=True, winning) = 0.177

winning-set containment = 1.000
overlap threshold = 0.700

candidate delta = 0.097
existing delta = 0.561
decision = reject
\end{lstlisting}

Overlapping winning position 1:
\begin{lstlisting}[style=macPython]
fen = "2r4r/1b2kp2/p3p2b/1p3p1p/7P/1B1N2P1/PPP1N3/R4RK1 w - - 3 20"
color = "white"
label = "winning"
eval_cp_stm = 209
\end{lstlisting}

Overlapping winning position 2:
\begin{lstlisting}[style=macPython]
fen = "rnr3k1/pp2ppbp/4b1p1/1NP5/8/2P1B3/PP2BPPP/R3N1K1 b - - 6 15"
color = "black"
label = "winning"
eval_cp_stm = 138
\end{lstlisting}

Overlapping winning position 3:
\begin{lstlisting}[style=macPython]
fen = "4kb1r/5ppp/4pn2/2pP4/5Pb1/1P1PN3/1Pr3PP/R1B2K1R b k - 3 19"
color = "black"
label = "winning"
eval_cp_stm = 158
\end{lstlisting}

Candidate trace:
\begin{lstlisting}[style=macPython]
L017>> rook_diff = count(chess.ROOK, stm) - count(chess.ROOK, opp)
L018>> minor_diff = (
L019>>     count(chess.BISHOP, stm) + count(chess.KNIGHT, stm)
L020>>     - count(chess.BISHOP, opp) - count(chess.KNIGHT, opp)
L021>> )
L022>> pawn_diff = count(chess.PAWN, stm) - count(chess.PAWN, opp)
L024>> exchange_edge = (
L025>>     rook_diff == 1 and minor_diff == -1
L026>>     and -1 <= pawn_diff <= 0
L027>> )
L028>> extra_minor = (
L029>>     rook_diff == 0 and minor_diff == 1
L030>>     and -2 <= pawn_diff <= -1
L031>> )
L032>> return bool(exchange_edge or extra_minor)

candidate value = True
\end{lstlisting}

Existing motif trace:
\begin{lstlisting}[style=macPython]
L007>> board = chess.Board(fen)
L008>> stm = board.turn
L009>> values = {
L010>>     chess.PAWN: 1, chess.KNIGHT: 3,
L011>>     chess.BISHOP: 3, chess.ROOK: 5, chess.QUEEN: 9
L012>> }
L013>> own = sum(
L014>>     values[p] * len(board.pieces(p, stm))
L015>>     for p in values
L016>> )
L017>> opp = sum(
L018>>     values[p] * len(board.pieces(p, not stm))
L019>>     for p in values
L020>> )
L021>> return own > opp

existing value = True
\end{lstlisting}
\end{systembubble}

\subsection{Compile and Runtime Feedback}

For parse, compile, and runtime failures, the evaluator includes only the relevant source excerpt. It omits rejected source for score and overlap failures because the model already produced that source in the preceding message.

\begin{systembubble}
Rejected: runtime error.

The candidate used an unsupported \texttt{python-chess} API:
\begin{lstlisting}[style=macPython]
board.is_empty(sq)
\end{lstlisting}

Use:
\begin{lstlisting}[style=macPython]
piece = board.piece_at(sq)
if piece is None:
    ...
\end{lstlisting}
\end{systembubble}

\clearpage
\section{Turning Python Functions into Text}
\label{appdx:transcription}

Each motif is a binary Python function: it takes a board position and returns \texttt{True} or \texttt{False}. Explainability is computed from these return values. The natural-language explanations in our figures are written afterward and are not used to compute the score.

For each displayed motif, we run the function on the shown positive and negative boards. We save the branch that returned \texttt{True}, the values computed on that branch, and the board facts it used, such as piece locations, attacks, defenders, legal moves, material, and controlled squares. If the return value cannot be tied to concrete board facts, we do not display the motif.

\begin{figure}[ht]
    \centering
    \begin{minipage}[t]{0.57\linewidth}
        \vspace{0pt}
\begin{lstlisting}[
    style=paperPython,
    basicstyle=\ttfamily\tiny,
    frame=single,
    numbers=left,
    numberstyle={\tiny\color{gray}\ifnum\value{lstnumber}=16\relax\llap{\textcolor{red!75!black}{\large$\bullet$}\hspace{0.25em}}\fi},
    numbersep=5pt,
    xleftmargin=2.0em,
    framexleftmargin=1.8em,
    framesep=2pt,
    aboveskip=0pt,
    belowskip=0pt,
    breaklines=true,
    breakatwhitespace=true,
    columns=fullflexible,
    keepspaces=true,
    showstringspaces=false
]
def motif(fen: str) -> bool:
    b = chess.Board(fen)
    us = b.turn
    them = not us
    if b.king(us) is None or b.king(them) is None:
        return False
    if b.is_check():
        return False
    if _mat(b, us) < _mat(b, them) - 3:
        return False
    pu, wu, mu = _report(b, us, them)
    pt, wt, mt = _report(b, them, us)
    if mu < 2:
        return False
    if wu >= 1 and wu > wt and pu > pt:
        return True
    return pu >= 4 and pu - pt >= 4 and wu >= wt
\end{lstlisting}
    \end{minipage}
    \hspace{0.035\linewidth}
    \begin{minipage}[t]{0.30\linewidth}
        \vspace{0pt}
        \centering
        \includegraphics[width=\linewidth]{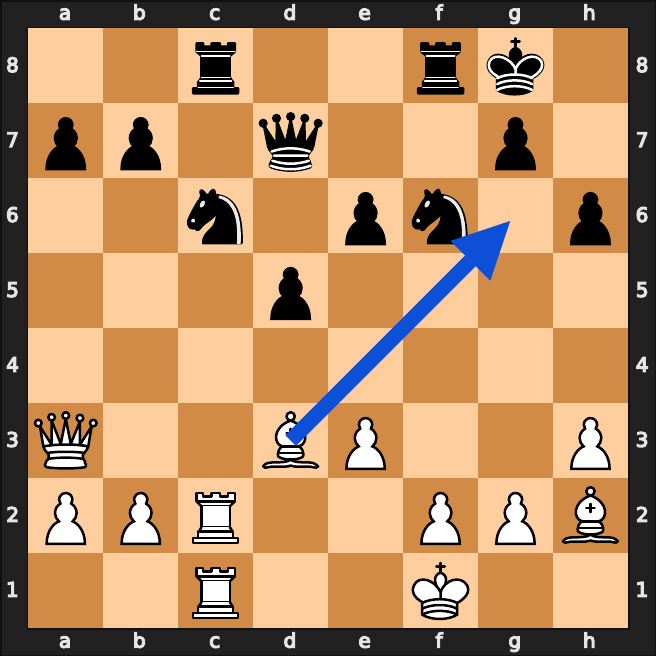}
    \end{minipage}
    \caption{An \texttt{Opus 5} motif from Carlsen--Erigaisi (2022). The highlighted branch combines material balance with an edge in king-zone pressure and weak squares. The arrow shows one contribution to White's pressure; the red dot marks the executed return.}
    \label{fig:transcription-example}
\end{figure}

In Figure~\ref{fig:transcription-example}, the motif finds
that White's bishop on d3 attacks g6 and h7, while its queen
on a3 attacks f8. This gives White two attacking pieces and
three attacks near Black's king, including the undefended
square g6. Black has no corresponding attacks near White's
king. Together with equal material and White not being in
check, these facts satisfy the highlighted condition,
so the motif returns \texttt{True}.

A Python script produces an initial explanation by filling
in sentence patterns with the recorded facts. For example,
``[Color]'s bishop on [square] attacks [targets]'' becomes
``White's bishop on d3 attacks g6 and h7.''
The script checks stated piece locations, attacks, defenders,
and material counts against the board.
\texttt{GPT-5.6 Sol} then rewrites the explanation for
readability. A Python checker checks statements it recognizes
against the board, such as piece locations and material counts.
If a check fails, the model tries again. If no rewrite passes,
the initial explanation is used.

\clearpage
\section{Data Collection}

\subsection{Chess Data Collection}
\label{appdx:chess_data_collection}

\paragraph{Winning and losing sets.}
For BinProp motif discovery, we use positions sampled from the Lichess Elite database \citep{lichessEliteDatabase}. Our local mirror contains monthly PGN files from January through November 2025. We evaluate positions with Stockfish 18 using \texttt{depth=12}. The final pool contains 200k positions: 100k winning and 100k losing, balanced by side to move. A position is winning if its evaluation for the side to move is between \(+125\) and \(+250\) centipawns, and losing if it is between \(-250\) and \(-125\). We discard mate scores.

\paragraph{Validation and test positions.}
Validation and test positions come from a separate PGN Mentor master-game corpus \citep{pgnmentor}. Each split contains 100 starting positions from distinct games, with no game shared between the splits. At each candidate position, we run Stockfish 18 with \texttt{MultiPV=8} for \texttt{80ms}. Scores are measured for the player to move in the starting position. We retain a starting position if the best move evaluates between \(+75\) and \(+125\) centipawns, the second-best move evaluates between \(-125\) and \(+50\), and their gap is at least \(75\) centipawns. Exactly one candidate move may evaluate at or above \(+75\). Positions with mate scores are discarded.

\paragraph{Quiet positions.}
\label{appdx:chess_quiet_positions}
A position is quiet if neither king is attacked and there is no visible material tactic: no pawn attacks a queen, rook, bishop, or knight; no attacked pawn or tactical piece is undefended; no undefended queen is attacked by a rook, bishop, or knight; and no undefended rook is attacked by a bishop or knight.

\paragraph{Continuation sampling.}
For each starting position, let \(c_1\) be Stockfish's best move and \(c_2\) the selected negative move. The latter is the second-best move, or another of the eight candidate moves that lies in the negative evaluation band and within \(50\) centipawns of the second-best move. After forcing \(c_1\) or \(c_2\), we repeatedly play Stockfish's top-ranked move for both sides until reaching the first quiet position after at least 6 plies, stopping if none is found by ply 16. We call this position the branch anchor.

From each anchor, we search up to 12 additional plies with \texttt{MultiPV=6}. On the original player's turns, candidate moves must remain within \(50\) centipawns of the principal continuation; on the opponent's turns, they must remain within \(50\) centipawns of the locally best move. We retain quiet, non-prefix-duplicate endpoints with the same side to move as the starting position. Endpoints descended from \(c_1\) form the positive pool, and endpoints descended from \(c_2\) form the negative pool. Because the anchor and endpoint depths vary, complete continuations are not restricted to 6--16 plies; in the published splits they span 8--26 plies.

Nearby negatives are generated on the \(c_1\) branch. At a later turn of the original player, the positive continuation plays Stockfish's top-ranked move and its nearby-negative sibling plays a lower-ranked move; both then play Stockfish's top reply. The pair therefore shares its entire earlier path and differs only in this final two-ply suffix. Neither suffix contains a capture or promotion, and both endpoints are quiet. After construction, we re-evaluate every retained starting position and endpoint with Stockfish 18 at depth 18. Scores are measured for the player to move in the starting position. Positive endpoints must evaluate at or above \(+75\) centipawns and remain within \(50\) centipawns of the starting position's best-move evaluation; negatives and nearby negatives must evaluate between \(-125\) and \(+50\). The weakest positive must be at least \(75\) centipawns better than the strongest negative or nearby negative. Each accepted starting position retains 12 positives, 12 negatives, and 8 nearby negatives.

Four of the 12 positives and four of the 12 negatives are shown to the model. The remaining eight positives, eight negatives, and all eight nearby negatives are scored. Positive siblings used to construct nearby negatives need not belong to the 12-positive pool and do not form an additional scored pool.

\clearpage
\subsection{Chess Bootstrap Confidence Intervals}
\label{app:bootstrap}

For the chess results in Table~\ref{tab:chess_results}, we estimate uncertainty by drawing 100 test starting positions with replacement, keeping Baseline and BinProp results for all eight models together at each position. We repeat this 100,000 times and use the 2.5th and 97.5th percentiles of the gains as 95\% confidence limits. We first average the selected motif scores for each test position, then average across the 100 positions; failures score zero. Table~\ref{tab:bootstrap} reports BinProp's gain over the baseline for each model and across all eight models.

\begin{table}[ht]
\centering
\caption{Chess explainability gains (BinProp minus Baseline). All values are in percentage points. Each row reports a 95\% bootstrap confidence interval.}
\label{tab:bootstrap}
\begin{tabular}{lcc}
\toprule
Model & Gain & 95\% confidence interval \\
\midrule
Qwen3.5-397B-A17B & +7.21 & [4.04, 10.31] \\
GPT-5.6 Sol & +14.38 & [10.50, 18.42] \\
DeepSeek-V4-Pro & +4.19 & [-0.40, 8.75] \\
Kimi-K3 & +13.65 & [8.94, 18.56] \\
GLM-5.2 & +18.17 & [14.67, 21.75] \\
GPT-5.5 & +16.90 & [13.19, 20.63] \\
Opus 5 & +24.19 & [21.00, 27.52] \\
Fable 5 & +26.52 & [22.00, 31.08] \\
\midrule
Eight-model average & +15.65 & [13.53, 17.90] \\
\bottomrule
\end{tabular}
\end{table}

\clearpage
\section{Go}
\label{sec:appendix-go}
\label{appdx:go_results}

\subsection{Data Collection}
\label{appdx:go_data_collection}

\paragraph{Winning and losing sets.}
For BinProp motif discovery, we use positions sampled from the JGDB Go game dataset \citep{jgdb}. We keep \(19\times19\), no-handicap games in which both players are at least 7 dan. Positions are evaluated from the side-to-move perspective with KataGo \citep{wu2019accelerating}, using the \texttt{kata1-b18c384nbt-s9996604416-d4316597426} network. The final pool contains 200k positions: 100k winning and 100k losing, balanced by side to move. A position is winning if its score lead is between \(+7.5\) and \(+15.0\) points, and losing if it is between \(-15.0\) and \(-7.5\).

\paragraph{Validation and test positions.}
Validation and test use JGDB games not used for motif discovery. Each split contains 100 starting positions from distinct games, with no game shared between the splits. We store the source game and move for every starting position.

At each candidate position, KataGo uses \(512\) visits to propose up to ten moves. We then independently analyze the child position after each proposed move with \(128\) visits and rank the moves by their rescored value for the player to move in the starting position. We retain a starting position if the best move evaluates between \(+7.5\) and \(+12.5\) points, the second-best move evaluates between \(-12.5\) and \(+5.0\), and their gap is at least \(7.5\) points. Exactly one candidate move may evaluate at or above \(+7.5\).

\paragraph{Continuation sampling.}
For each accepted starting position, we first choose two moves from the candidates rescored above. Let \(c_1\) be the rescored best move. Let \(c_2\) be the negative seed: it is either the rescored second-best move or another of the ten candidate moves whose score lies in the negative band and within \(5\) points of the second-best move. These two moves start separate continuation trees from the same starting position. All scores below are measured from the perspective of the player to move in the starting position.

From each seed, we search continuations ending 8--12 plies after the starting position. At every intermediate position, KataGo proposes up to four moves, and we independently analyze each resulting child with \(128\) visits before expanding it further. To avoid pruning a promising path too early, an intermediate node may lie up to \(5\) points outside the final acceptance bands. This tolerance is used only during search; it never applies to a retained endpoint. We retain only endpoints with the same player to move as in the starting position, and hence at an even distance from it.

After the search, we apply the strict endpoint filters. An endpoint in the \(c_1\) tree is eligible as a positive only if it evaluates at or above \(+7.5\) points and lies within \(5\) points of the starting position's rescored best-move value. An endpoint in the \(c_2\) tree is eligible as a negative only if it evaluates between \(-12.5\) and \(+5.0\). Nearby negatives, constructed separately below from the \(c_1\) tree, must satisfy this same negative band. Thus, the initial branch determines an endpoint's candidate label, but its final score determines whether it is retained.

Finally, we apply a separation check across all three pools for each starting position: the weakest retained positive must evaluate at least \(7.5\) points above the strongest retained negative or nearby negative. We accept only starting positions with enough qualifying endpoints, retaining exactly 12 positives, 12 negatives, and 8 nearby negatives from each.

Nearby negatives are generated on the \(c_1\) branch. From the positive paths, we select a late turn of the player to move in the starting position, no earlier than ply 7. We independently rescore candidate moves at that position: the positive sibling plays KataGo's best move, while the nearby-negative sibling plays a lower-ranked move, and each is followed by KataGo's best reply for the opponent. The pair therefore shares its entire earlier path and differs only in this final two-ply suffix. We retain the nearby sibling only if it enters the negative band and remains at least \(7.5\) points below its positive sibling.

Four positives and four negatives are shown to the model. The remaining eight positives, eight negatives, and all eight nearby negatives are held out for evaluation.

\clearpage
\subsection{Results}
Results for all eight models are reported in Table~\ref{tab:go_results_full}. Each BinProp run uses 100 proposal steps: invalid code counts as a step, while provider failures and empty responses are retried. The final motif sets contain 35 motifs for \texttt{Opus 5} and 30 for \texttt{Kimi-K3}.
\begin{table}[!ht]
\centering
\caption{Mean explainability on Go test positions for all eight models, using maximum reasoning for both Baseline and BinProp explanations.}
\label{tab:go_results_full}
\small
\renewcommand{\arraystretch}{0.93}
\begin{tabular}{llcc}
\toprule
Model & Method & Mean Explainability & Gain \\
\midrule
\multirow{2}{*}{DeepSeek-V4-Pro \citep{deepseekv4pro}}
  & Baseline & 0.051 & \multirow{2}{*}{\textcolor{green!55!black}{+0.098}} \\
  & BinProp (ours)       & 0.149 & \\
\addlinespace[0.1em]
\multirow{2}{*}{Qwen3.5-397B-A17B \citep{qwen2026qwen35}}
  & Baseline & 0.109 & \multirow{2}{*}{\textcolor{green!55!black}{+0.154}} \\
  & BinProp (ours)       & 0.264 & \\
\addlinespace[0.1em]
\multirow{2}{*}{Kimi-K3 \citep{kimiteam2026k3}}
  & Baseline & 0.141 & \multirow{2}{*}{\textcolor{green!55!black}{+0.134}} \\
  & BinProp (ours)       & 0.274 & \\
\addlinespace[0.1em]
\multirow{2}{*}{GLM-5.2 \citep{zeng2026glm}}
  & Baseline & 0.141 & \multirow{2}{*}{\textcolor{green!55!black}{+0.158}} \\
  & BinProp (ours)       & 0.299 & \\
\addlinespace[0.1em]
\multirow{2}{*}{GPT-5.6 Sol \citep{openai2026gpt56}}
  & Baseline & 0.245 & \multirow{2}{*}{\textcolor{green!55!black}{+0.055}} \\
  & BinProp (ours)       & 0.300 & \\
\addlinespace[0.1em]
\multirow{2}{*}{Opus 5 \citep{anthropic2026opus5}}
  & Baseline & 0.190 & \multirow{2}{*}{\textcolor{green!55!black}{+0.111}} \\
  & BinProp (ours)       & 0.301 & \\
\addlinespace[0.1em]
\multirow{2}{*}{GPT-5.5 \citep{openai2026gpt55}}
  & Baseline & 0.181 & \multirow{2}{*}{\textcolor{green!55!black}{+0.122}} \\
  & BinProp (ours)       & 0.303 & \\
\addlinespace[0.1em]
\multirow{2}{*}{Fable 5 \citep{anthropic2026fable5}}
  & Baseline & 0.159 & \multirow{2}{*}{\textcolor{green!55!black}{+0.155}} \\
  & BinProp (ours)       & \textbf{0.314} & \\
\bottomrule
\end{tabular}
\end{table}

\clearpage
\subsection{Qualitative examples}
\begin{figure}[H]
\centering
\resizebox*{!}{0.80\textheight}{%
\begin{tabular}{@{}c@{\hspace{12pt}}c@{}}
\includegraphics[width=246pt]{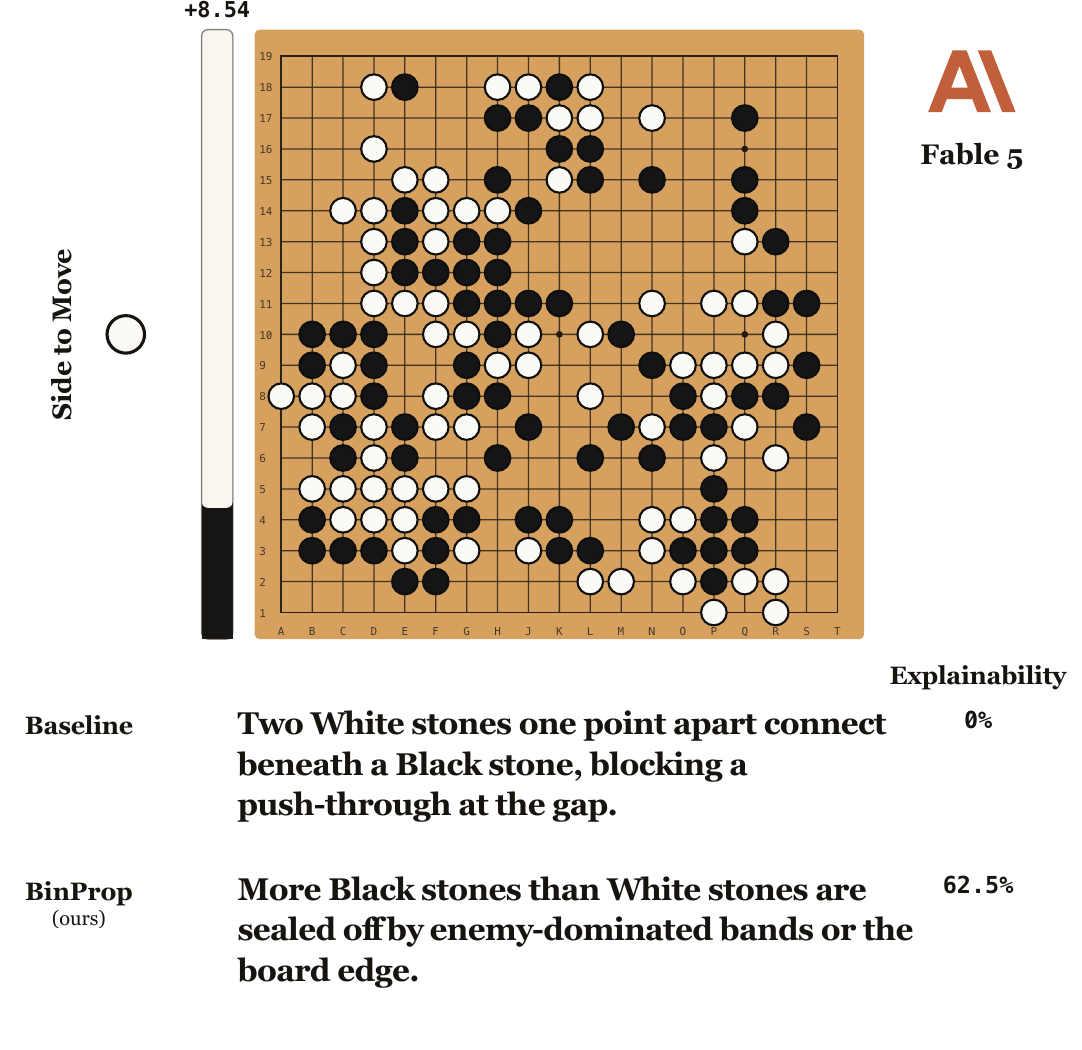} &
\includegraphics[width=246pt]{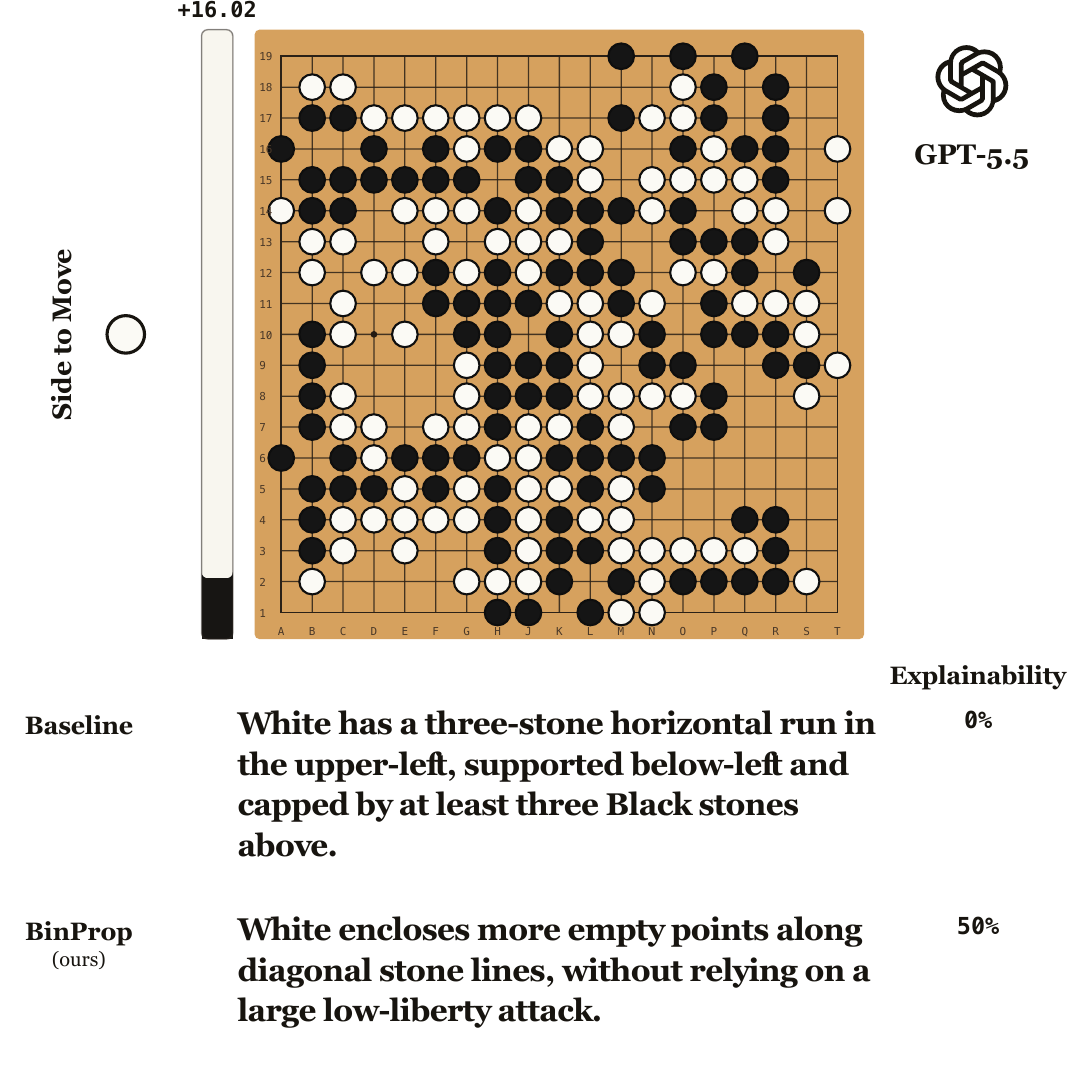} \\[6pt]
\includegraphics[width=246pt]{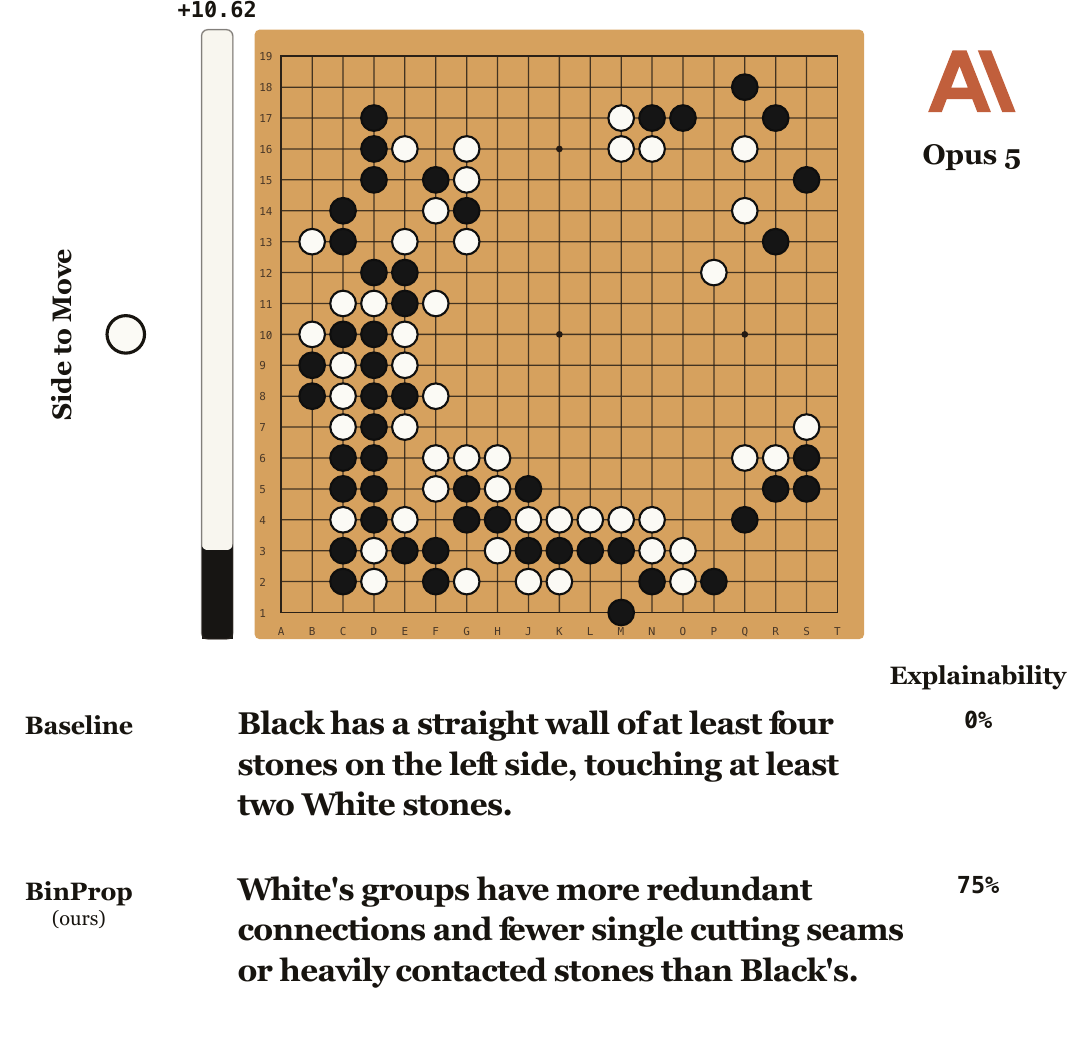} &
\includegraphics[width=246pt]{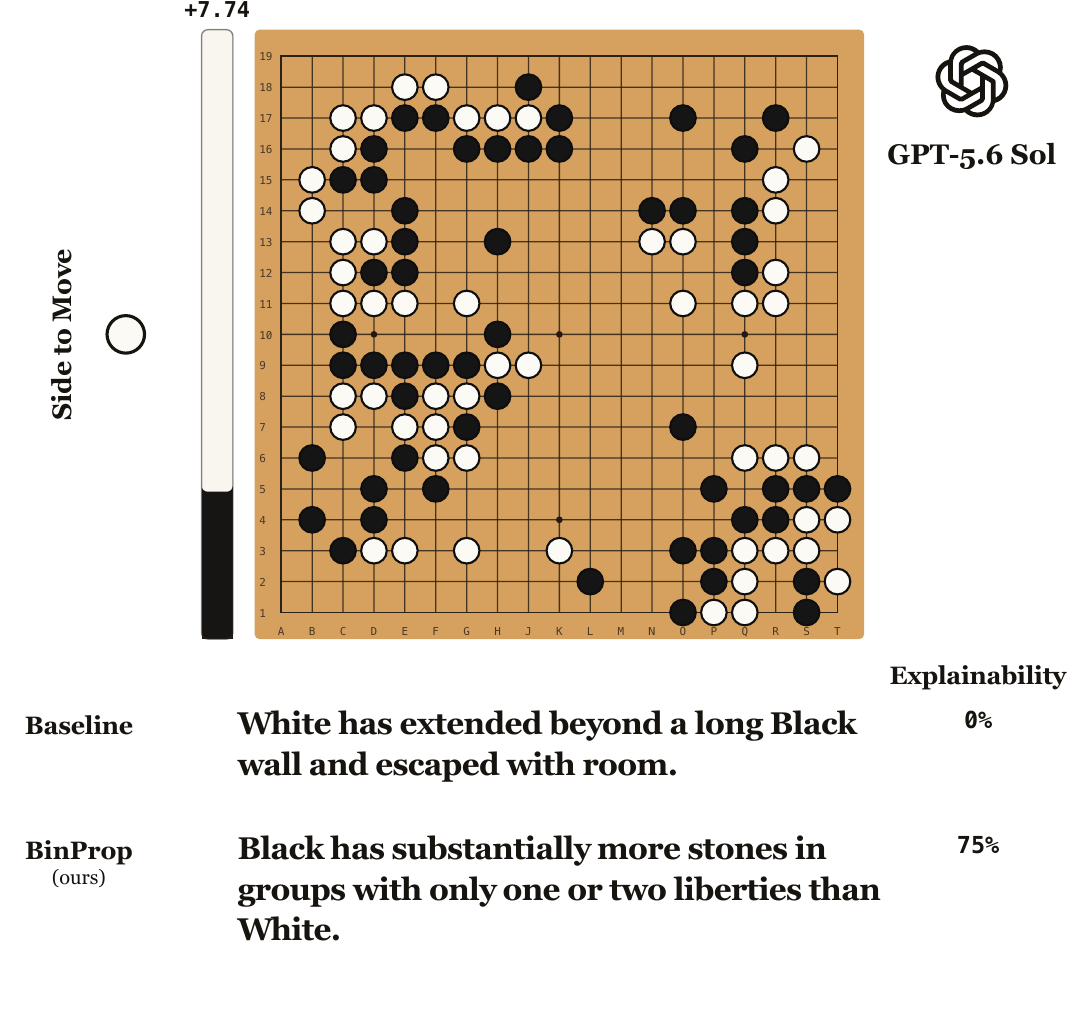} \\[6pt]
\includegraphics[width=246pt]{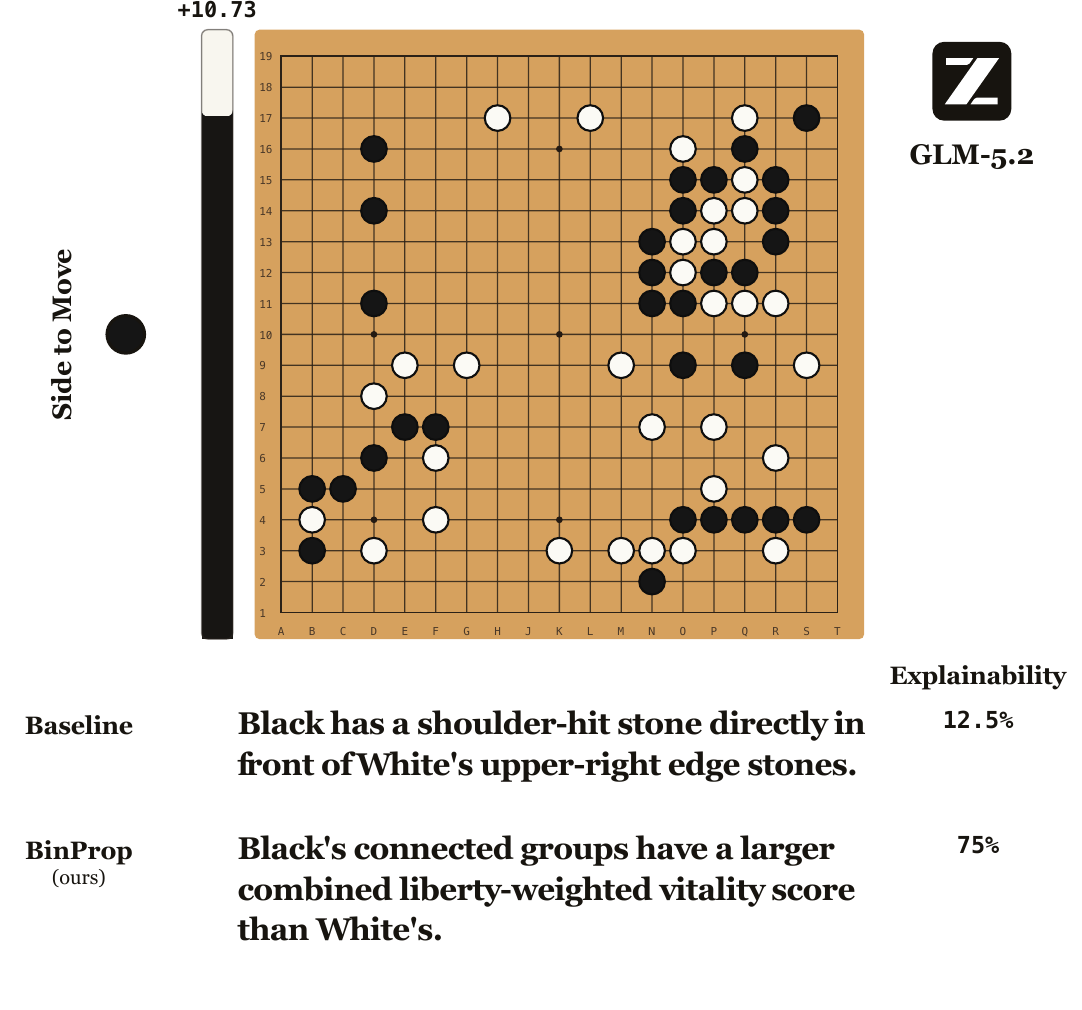} &
\includegraphics[width=246pt]{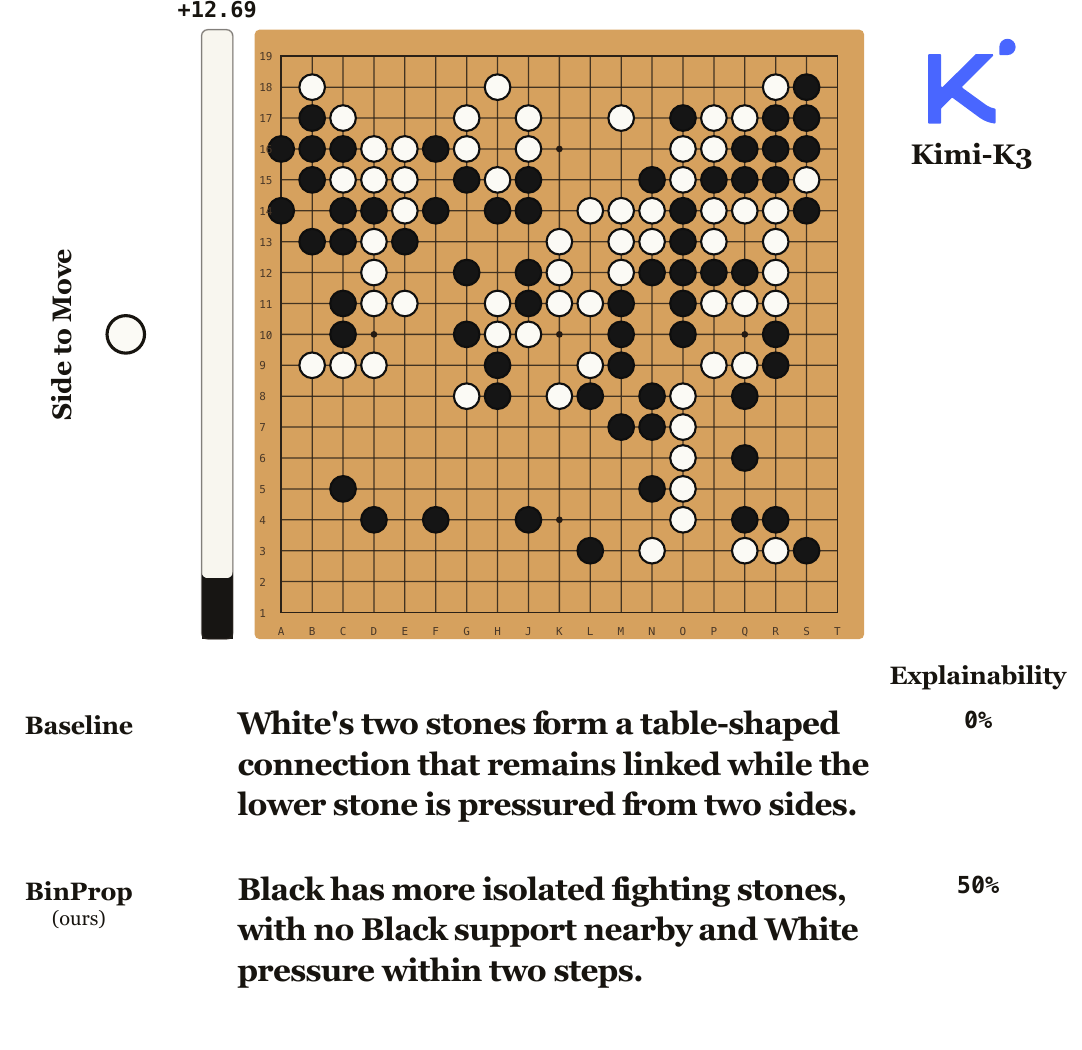}
\end{tabular}}
\caption{Baseline and BinProp explanations for the six highest-scoring Go models. Each panel shows one held-out positive position; explainability is measured over eight positive, eight negative, and eight nearby-negative positions. Bars show KataGo's evaluation.}
\label{fig:go-top-six}
\end{figure}

Figure~\ref{fig:go-top-six} compares Baseline and BinProp explanations for Go.

\clearpage
\subsection{AlphaGo's Move 37}
\label{appdx:go-move37}
\begin{figure}[H]
\centering
\includegraphics[width=\linewidth,height=0.76\textheight,keepaspectratio]{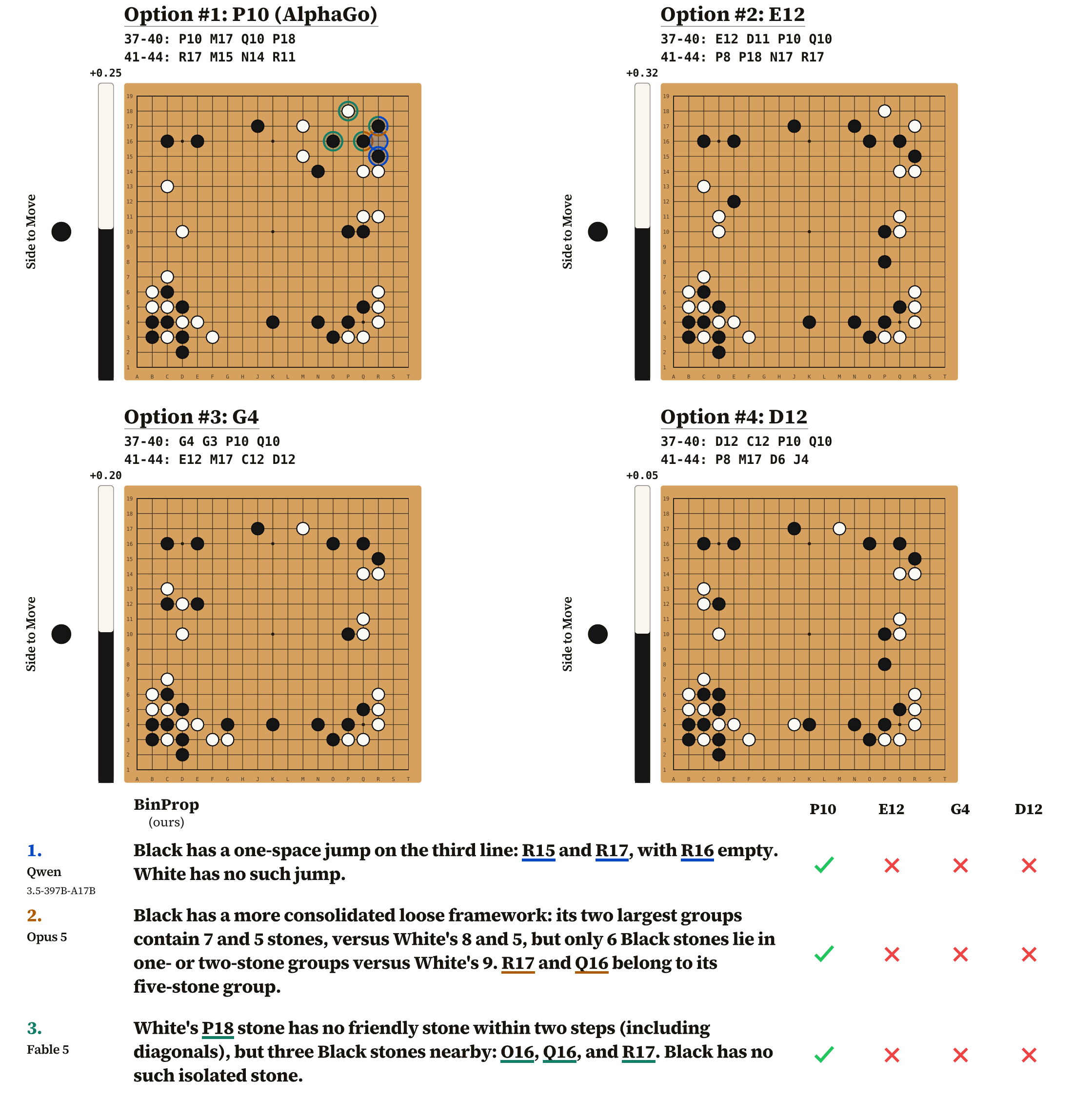}
\caption{Eight-ply continuations after P10, E12, G4, and D12, with three BinProp explanations. Bars show KataGo's estimated lead for Black immediately after each candidate move.}
\label{fig:alphago-move37-a}
\label{fig:alphago-move37-b}
\end{figure}

We compare AlphaGo's move 37 (P10) in its second game against Lee Sedol~\citep{hassabis2016alphago} with KataGo's three highest-ranked alternatives. Figure~\ref{fig:alphago-move37-a} shows eight-ply continuations and three existing BinProp properties that hold after P10 but not after the alternatives. 

\clearpage
\section{Draughts (American Checkers)}
\label{sec:appendix-checkers}
\subsection{Data Collection}
\label{appdx:checkers_data_collection}
\paragraph{Source games and engine.}
We study the standard $8\times8$ game known as draughts or American checkers~\citep{wcdfRules2012}. Its two piece types are uncrowned pieces, conventionally called men, and kings. We use 22{,}621 recorded games from Open Checkers Archive 2.0~\citep{openCheckersArchive2}. Ponder~\citep{ponderCheckers}, with forced-capture continuations preserved, evaluates positions at depth 8. We divide its scores by 256, so one unit below is the value of one uncrowned piece.

\paragraph{Winning and losing sets.}
We collect 8{,}000 winning and 8{,}000 losing positions, balanced by side to move. Winning positions evaluate from $+1.25$ to $+2.5$ uncrowned-piece equivalents and losing positions from $-2.5$ to $-1.25$. The sets used for motif discovery, validation, and test are mutually exclusive.

\paragraph{Validation and test positions.}
Validation and test each contain 100 starting positions from distinct held-out games. Ponder ranks candidate moves at depth 8. We retain positions whose best move scores $+0.75$ to $+1.25$ uncrowned-piece equivalents, whose second-best move scores $-1.25$ to $+0.5$, and whose best--second-best gap is at least $0.75$. Only one move may score at least $+0.75$. All continuation scores below use the perspective of the player to move in the starting position.

\paragraph{Continuation sampling.}
Positive continuations begin with the best move; negative continuations begin with the second-best move or another move within $0.5$ of its score that satisfies the same score-band and gap requirements. From each seed, we follow Ponder's best moves for at least six plies, until no capture is available, then branch over up to four moves within $0.5$ of the current player's best move. Continuation moves are searched at depth 6 and retained endpoints are evaluated at depth 8. We stop at 16 plies from the starting position and retain only endpoints with no available captures and the same player to move. Positives must score at least $+0.75$ and remain within $0.5$ of the initial best-move value; negatives must score $-1.25$ to $0$.

Nearby negatives share at least the first six plies of a positive continuation. At the shared decision point, the positive line plays Ponder's best move and the nearby-negative line plays a lower-ranked move. Each move is followed by Ponder's best reply, so the two endpoints differ only in their final two plies and have the same player to move. Both endpoints must have no available captures and lie within 16 plies of the starting position. We retain the pair only if the positive satisfies the filters above and the nearby negative scores $-1.25$ to $0$.

Every positive must exceed every negative or nearby negative by at least $0.75$ uncrowned-piece equivalents. Each episode contains 12 positives, 12 negatives, and eight nearby negatives. Four positives and four negatives are shown; the remaining eight positions in each group are held out. Motif selection and scoring follow the chess protocol (\S~\ref{sec:evaluation_explanations}), including nearby-negative specificity.

\subsection{Results}

Results are reported in Table~\ref{tab:checkers-results}.
\begin{table}[ht]
\centering
\caption{Mean explainability on the draughts test set for Baseline and BinProp explanations.}
\label{tab:checkers-results}
\begin{tabular}{llcc}
\toprule
Model & Method & Mean Explainability & Gain \\
\midrule
\multirow{2}{*}{Fable 5 \citep{anthropic2026fable5}}
  & Baseline & 0.355 & \multirow{2}{*}{\textcolor{red!70!black}{-0.114}} \\
  & BinProp (ours) & 0.241 & \\
\addlinespace[0.2em]
\multirow{2}{*}{DeepSeek-V4-Pro \citep{deepseekv4pro}}
  & Baseline & 0.130 & \multirow{2}{*}{\textcolor{green!55!black}{+0.222}} \\
  & BinProp (ours) & 0.352 & \\
\addlinespace[0.2em]
\multirow{2}{*}{Qwen3.5-397B-A17B \citep{qwen2026qwen35}}
  & Baseline & 0.378 & \multirow{2}{*}{\textcolor{green!55!black}{+0.023}} \\
  & BinProp (ours) & 0.401 & \\
\addlinespace[0.2em]
\multirow{2}{*}{Opus 5 \citep{anthropic2026opus5}}
  & Baseline & 0.425 & \multirow{2}{*}{\textcolor{green!55!black}{+0.058}} \\
  & BinProp (ours) & 0.483 & \\
\addlinespace[0.2em]
\multirow{2}{*}{GPT-5.5 \citep{openai2026gpt55}}
  & Baseline & 0.371 & \multirow{2}{*}{\textcolor{green!55!black}{+0.126}} \\
  & BinProp (ours) & 0.497 & \\
\addlinespace[0.2em]
\multirow{2}{*}{GPT-5.6 Sol \citep{openai2026gpt56}}
  & Baseline & 0.393 & \multirow{2}{*}{\textcolor{green!55!black}{+0.109}} \\
  & BinProp (ours) & 0.502 & \\
\addlinespace[0.2em]
\multirow{2}{*}{GLM-5.2 \citep{zeng2026glm}}
  & Baseline & 0.359 & \multirow{2}{*}{\textcolor{green!55!black}{+0.148}} \\
  & BinProp (ours) & 0.506 & \\
\addlinespace[0.2em]
\multirow{2}{*}{Kimi-K3 \citep{kimiteam2026k3}}
  & Baseline & 0.415 & \multirow{2}{*}{\textcolor{green!55!black}{+0.113}} \\
  & BinProp (ours) & \textbf{0.528} & \\
\bottomrule
\end{tabular}
\end{table}

\clearpage
\subsection{Qualitative examples}
\begin{figure}[H]
\centering
\includegraphics[width=0.98\textwidth,height=0.80\textheight,keepaspectratio]{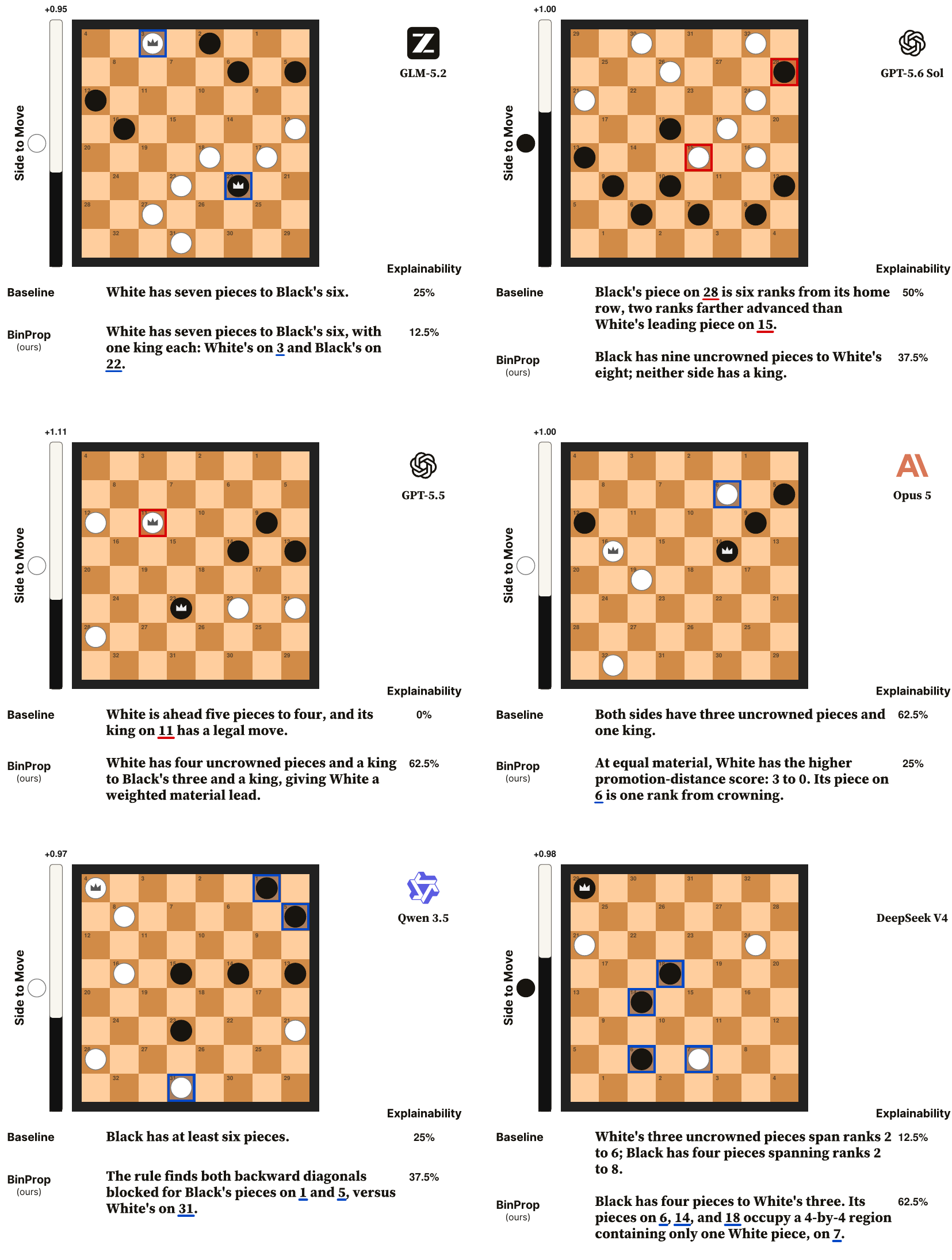}
\caption{Baseline and BinProp explanations on held-out draughts positions. Examples are sampled without ranking their held-out scores. Bars show Ponder's evaluation in uncrowned-piece equivalents.}
\label{fig:checkers-top-six}
\end{figure}

Figure~\ref{fig:checkers-top-six} compares Baseline and BinProp explanations for draughts.

\clearpage

\clearpage
\section{Shogi}
\label{sec:appendix-shogi}
\subsection{Data Collection}
\label{appdx:shogi_data_collection}
\paragraph{Source games and engine.}
We use recorded CSA games from the official WDoor/Floodgate 2024 and 2025 archives~\citep{wdoorFloodgate}. YaneuraOu~\citep{yaneuraou}, using the published Suisho5 NNUE parameters~\citep{suisho5}, evaluates every retained position.

\paragraph{Winning and losing sets.}
We collect 100{,}000 winning and 100{,}000 losing positions, balanced by side to move. Their side-to-move evaluations are $+800$ to $+1600$ and $-1600$ to $-800$ centipawns, respectively. The sets used for motif discovery, validation, and test are mutually exclusive.

\paragraph{Validation and test positions.}
Validation and test each contain 100 starting positions from distinct held-out games. YaneuraOu searches up to eight candidate moves with 10{,}000 nodes per position. We retain positions whose best move scores $+800$ to $+1300$ centipawns, whose second-best move scores $-1300$ to $+500$, and whose best--second-best gap is at least $800$. Only one searched move may score at least $+800$. All continuation scores below use the perspective of the player to move in the starting position.

\paragraph{Continuation sampling.}
We grow separate trees after the best move and after qualifying alternatives. An alternative must score $-1300$ to $+500$ centipawns, lie within $500$ of the second-best move, and trail the best move by at least $800$. At each subsequent position, YaneuraOu searches up to six moves with 6{,}000 nodes. We retain endpoints 8--12 plies from the starting position, with the same player to move. Positive endpoints must score at least $+800$ and remain within $500$ centipawns of the initial best-move value. Negatives must score $-1300$ to $0$ and remain within $500$ of the second-best move's initial score. Search paths may temporarily leave these bands, but retained endpoints must satisfy them.

Nearby negatives share at least the first six plies of a positive continuation. At the shared decision point, the positive line plays YaneuraOu's best move and the nearby-negative line plays a lower-ranked move. Each move is followed by YaneuraOu's best reply, so the two endpoints differ only in their final two plies and have the same player to move. Both must end 8--12 plies from the starting position. We retain the pair only if the positive satisfies the filters above and the nearby negative scores $-1300$ to $0$ centipawns.

Every positive must exceed every negative or nearby negative by at least $800$ centipawns. Each episode contains 12 positives, 12 negatives, and eight nearby negatives. Four positives and four negatives are shown; the remaining eight positions in each group are held out. Motif selection and scoring follow the chess protocol (\S~\ref{sec:evaluation_explanations}), including nearby-negative specificity.

\subsection{Results}

Results are reported in Table~\ref{tab:shogi-results}.
\begin{table}[ht]
\centering
\caption{Mean explainability on the shogi test set for Baseline and BinProp explanations.}
\label{tab:shogi-results}
\begin{tabular}{llcc}
\toprule
Model & Method & Mean Explainability & Gain \\
\midrule
\multirow{2}{*}{Qwen3.5-397B-A17B \citep{qwen2026qwen35}}
  & Baseline & 0.204 & \multirow{2}{*}{\textcolor{green!55!black}{+0.029}} \\
  & BinProp (ours) & 0.233 & \\
\addlinespace[0.2em]
\multirow{2}{*}{DeepSeek-V4-Pro \citep{deepseekv4pro}}
  & Baseline & 0.086 & \multirow{2}{*}{\textcolor{green!55!black}{+0.207}} \\
  & BinProp (ours) & 0.293 & \\
\addlinespace[0.2em]
\multirow{2}{*}{GPT-5.5 \citep{openai2026gpt55}}
  & Baseline & 0.297 & \multirow{2}{*}{\textcolor{green!55!black}{+0.016}} \\
  & BinProp (ours) & 0.313 & \\
\addlinespace[0.2em]
\multirow{2}{*}{Kimi-K3 \citep{kimiteam2026k3}}
  & Baseline & 0.253 & \multirow{2}{*}{\textcolor{green!55!black}{+0.072}} \\
  & BinProp (ours) & 0.326 & \\
\addlinespace[0.2em]
\multirow{2}{*}{Opus 5 \citep{anthropic2026opus5}}
  & Baseline & 0.300 & \multirow{2}{*}{\textcolor{green!55!black}{+0.028}} \\
  & BinProp (ours) & 0.327 & \\
\addlinespace[0.2em]
\multirow{2}{*}{Fable 5 \citep{anthropic2026fable5}}
  & Baseline & 0.269 & \multirow{2}{*}{\textcolor{green!55!black}{+0.058}} \\
  & BinProp (ours) & 0.327 & \\
\addlinespace[0.2em]
\multirow{2}{*}{GPT-5.6 Sol \citep{openai2026gpt56}}
  & Baseline & 0.275 & \multirow{2}{*}{\textcolor{green!55!black}{+0.054}} \\
  & BinProp (ours) & 0.329 & \\
\addlinespace[0.2em]
\multirow{2}{*}{GLM-5.2 \citep{zeng2026glm}}
  & Baseline & 0.231 & \multirow{2}{*}{\textcolor{green!55!black}{+0.105}} \\
  & BinProp (ours) & \textbf{0.336} & \\
\bottomrule
\end{tabular}
\end{table}

\clearpage
\subsection{Qualitative examples}
\begin{figure}[H]
\centering
\includegraphics[width=0.98\textwidth,height=0.80\textheight,keepaspectratio]{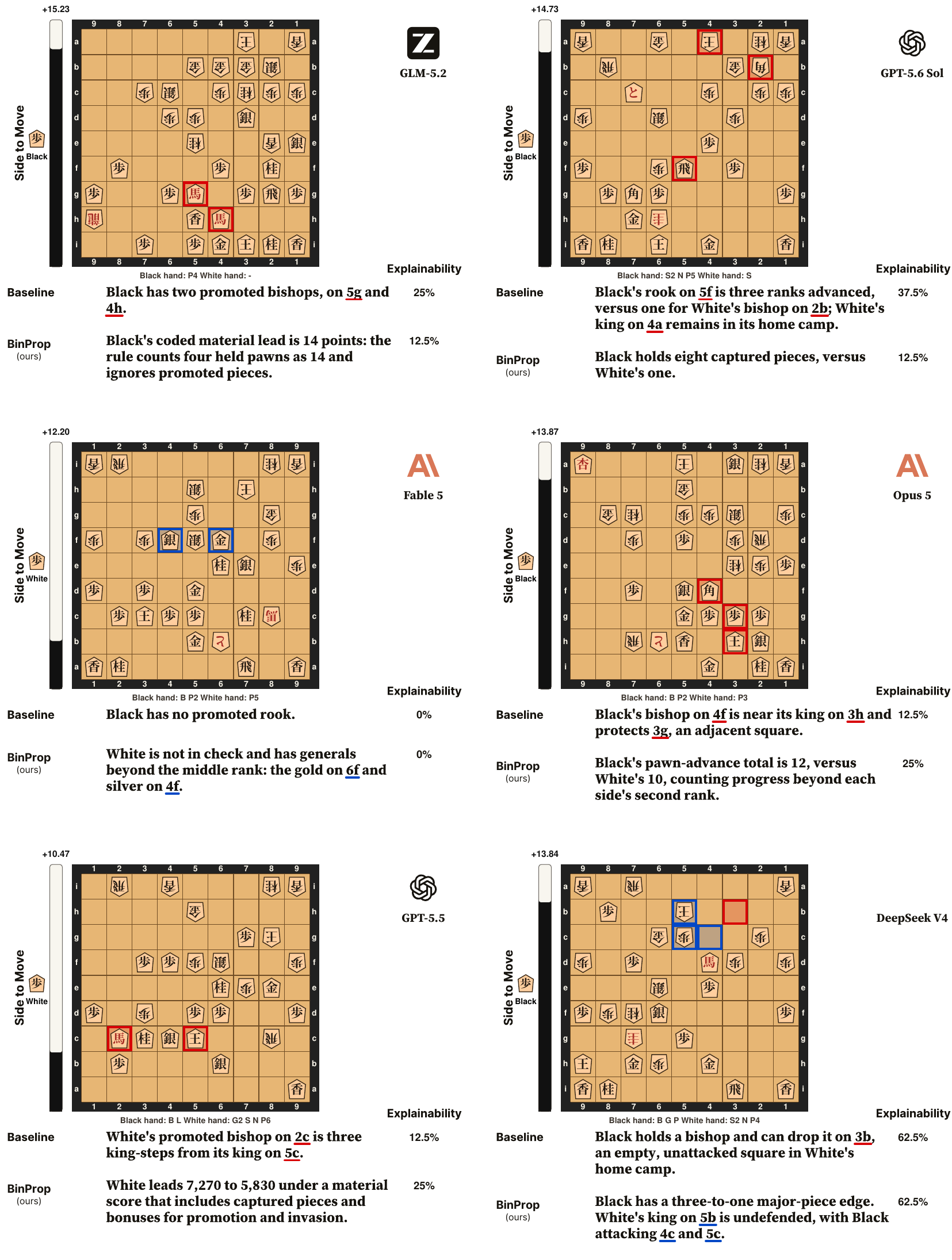}
\caption{Baseline and BinProp explanations on held-out shogi positions. Examples are sampled without ranking their held-out scores. Bars show YaneuraOu's evaluation in pawns.}
\label{fig:shogi-top-six}
\end{figure}

Figure~\ref{fig:shogi-top-six} compares Baseline and BinProp explanations for shogi.

\clearpage

\clearpage
\section{Explaining Moves from Expert Games}
\label{appdx:chess_game_examples}
We run BinProp with \texttt{Opus 5} for a further 150 iterations, adding 22 motifs for a total of 48. Using Stockfish, we identify mistakes by world champions and other top players in the PGN Mentor game collection~\citep{pgnmentor}. We use BinProp with this extended motif set to explain nine such decisions (Figures~\ref{fig:website-opus250-1}--\ref{fig:website-opus250-9}), comparing the move played with Stockfish's preferred move.

\begingroup
\setlength{\parskip}{0pt}
\setlength{\intextsep}{1pt}
\setlength{\abovecaptionskip}{1pt}
\setlength{\belowcaptionskip}{0pt}
\captionsetup{font=footnotesize}

\begin{figure}[H]
\centering
\includegraphics[width=0.64\linewidth]{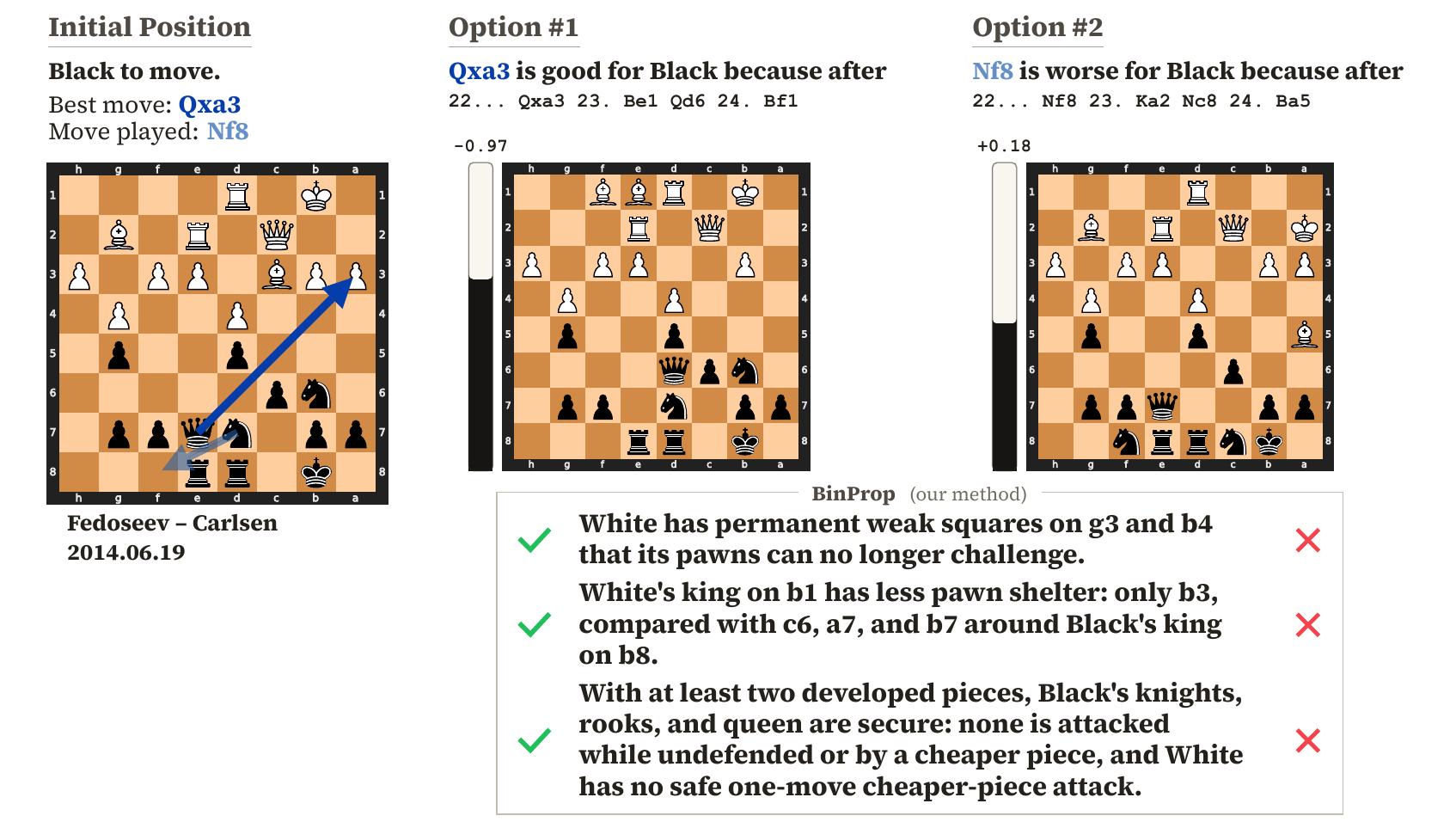}
\caption{Fedoseev--Carlsen (2014).}
\label{fig:website-opus250-1}
\end{figure}
\vfill
\begin{figure}[H]
\centering
\includegraphics[width=0.64\linewidth]{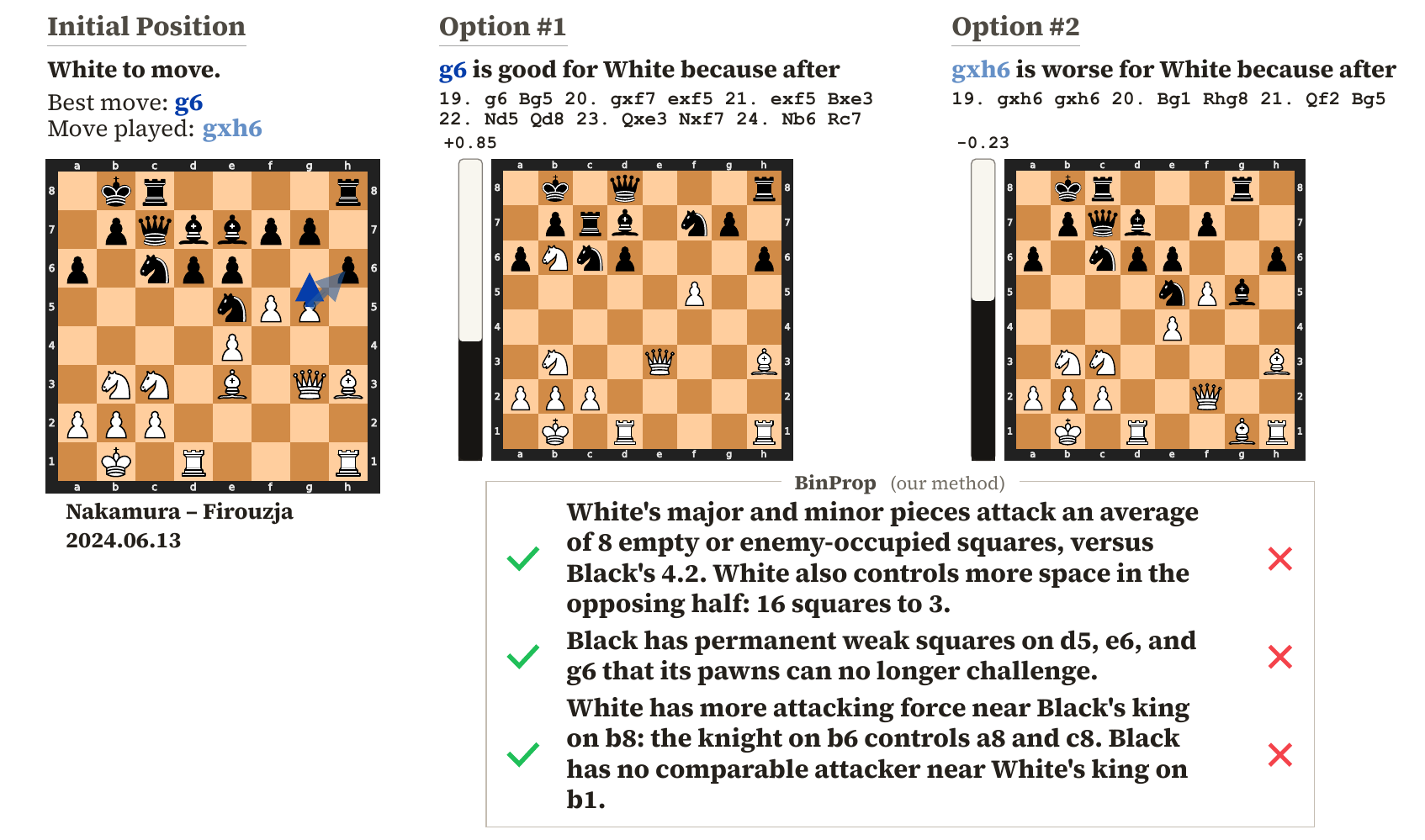}
\caption{Nakamura--Firouzja (2024).}
\label{fig:website-opus250-2}
\end{figure}
\vfill
\begin{figure}[H]
\centering
\includegraphics[width=0.64\linewidth]{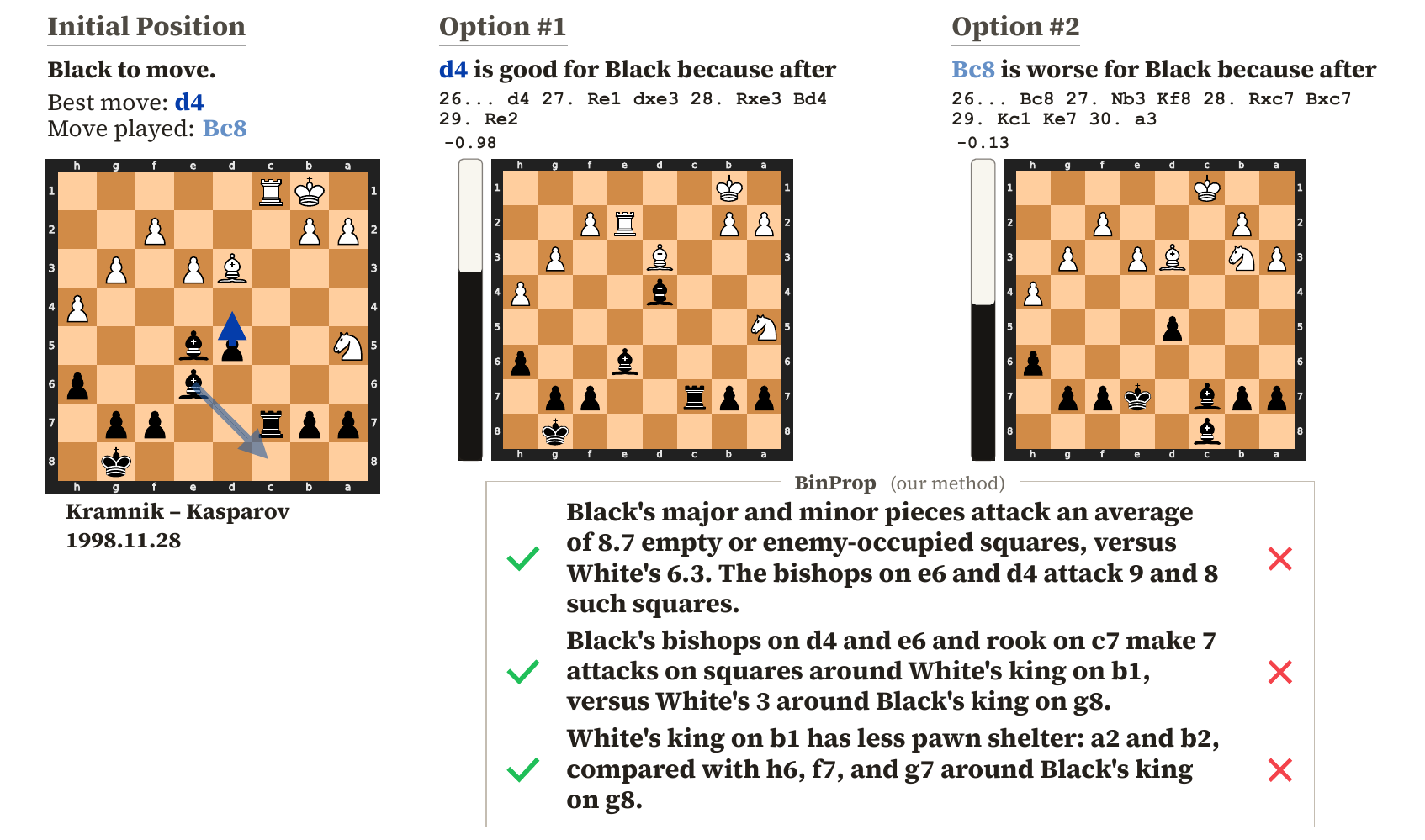}
\caption{Kramnik--Kasparov (1998).}
\label{fig:website-opus250-3}
\end{figure}

\clearpage

\begin{figure}[H]
\centering
\includegraphics[width=0.64\linewidth]{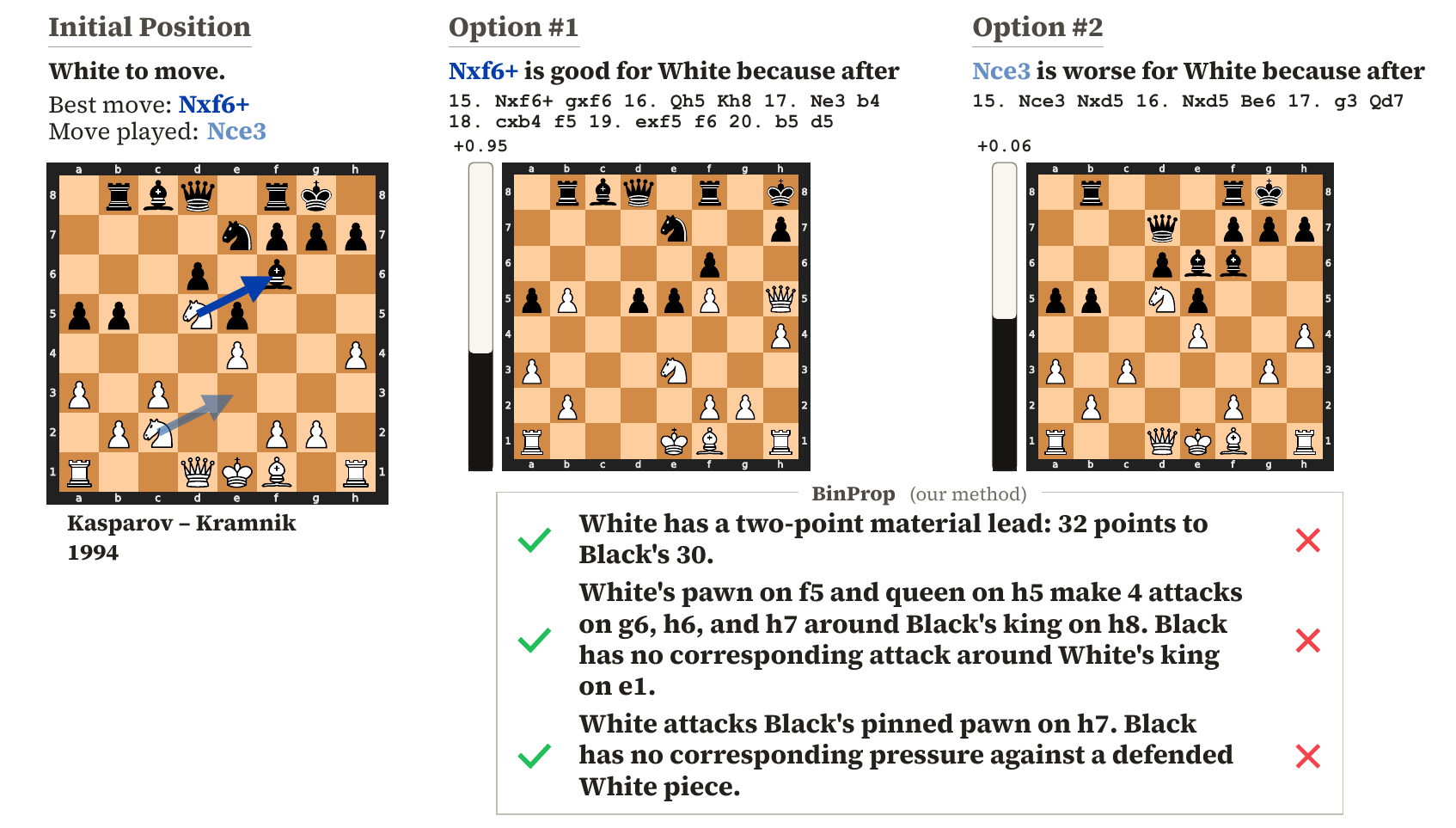}
\caption{Kasparov--Kramnik (1994).}
\label{fig:website-opus250-4}
\end{figure}
\vfill
\begin{figure}[H]
\centering
\includegraphics[width=0.64\linewidth]{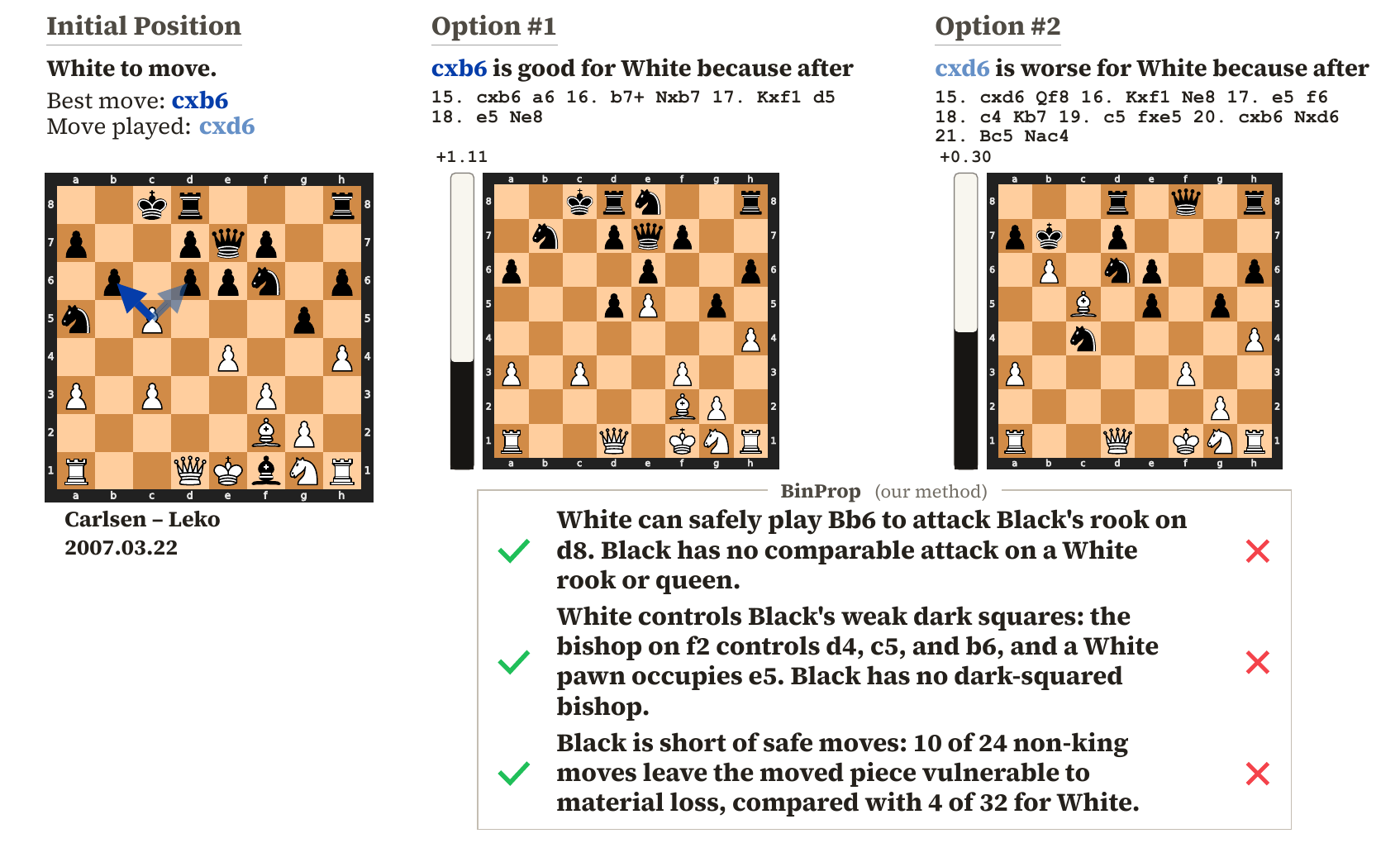}
\caption{Carlsen--Leko (2007).}
\label{fig:website-opus250-5}
\end{figure}
\vfill
\begin{figure}[H]
\centering
\includegraphics[width=0.64\linewidth]{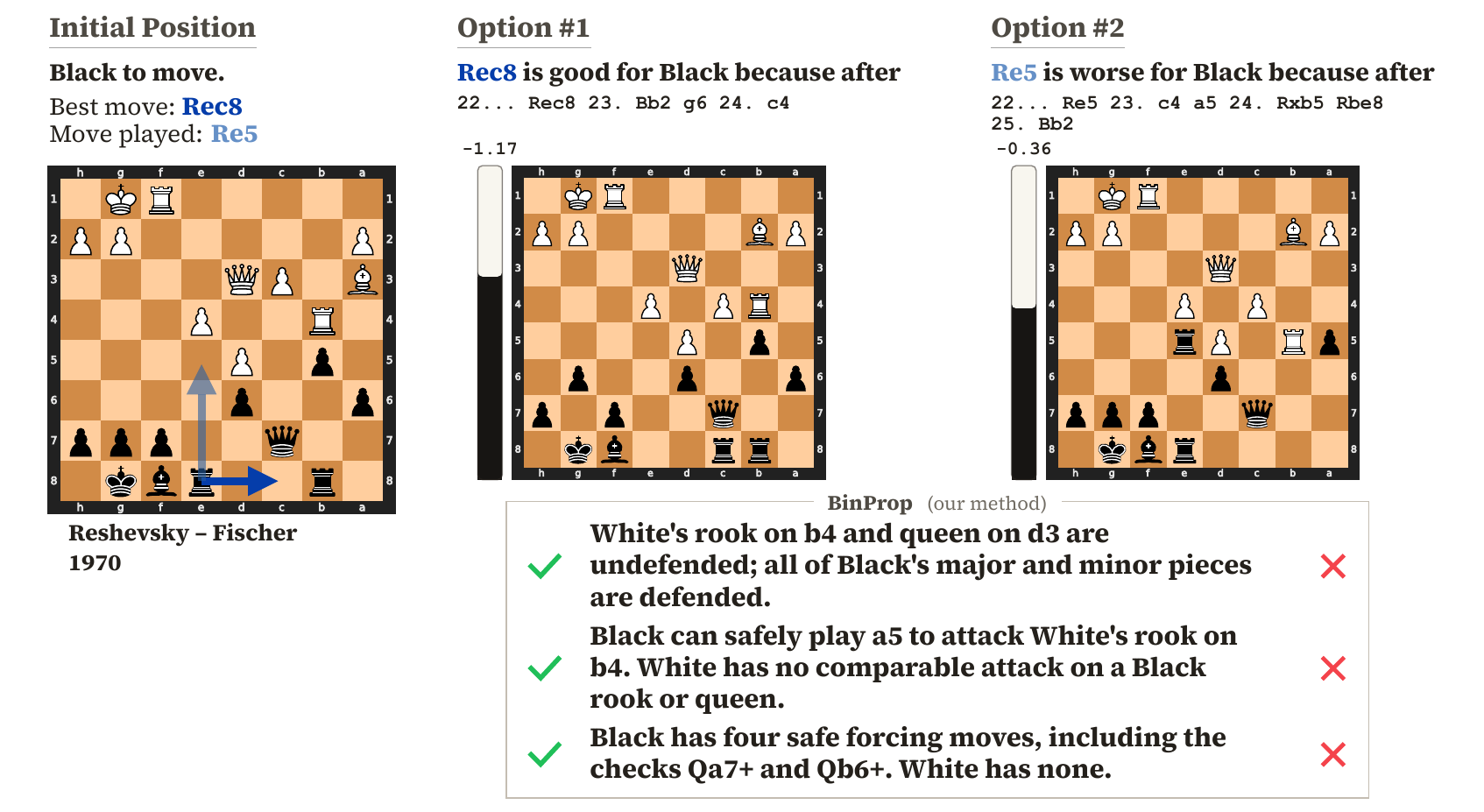}
\caption{Reshevsky--Fischer (1970).}
\label{fig:website-opus250-6}
\end{figure}

\clearpage

\begin{figure}[H]
\centering
\includegraphics[width=0.64\linewidth]{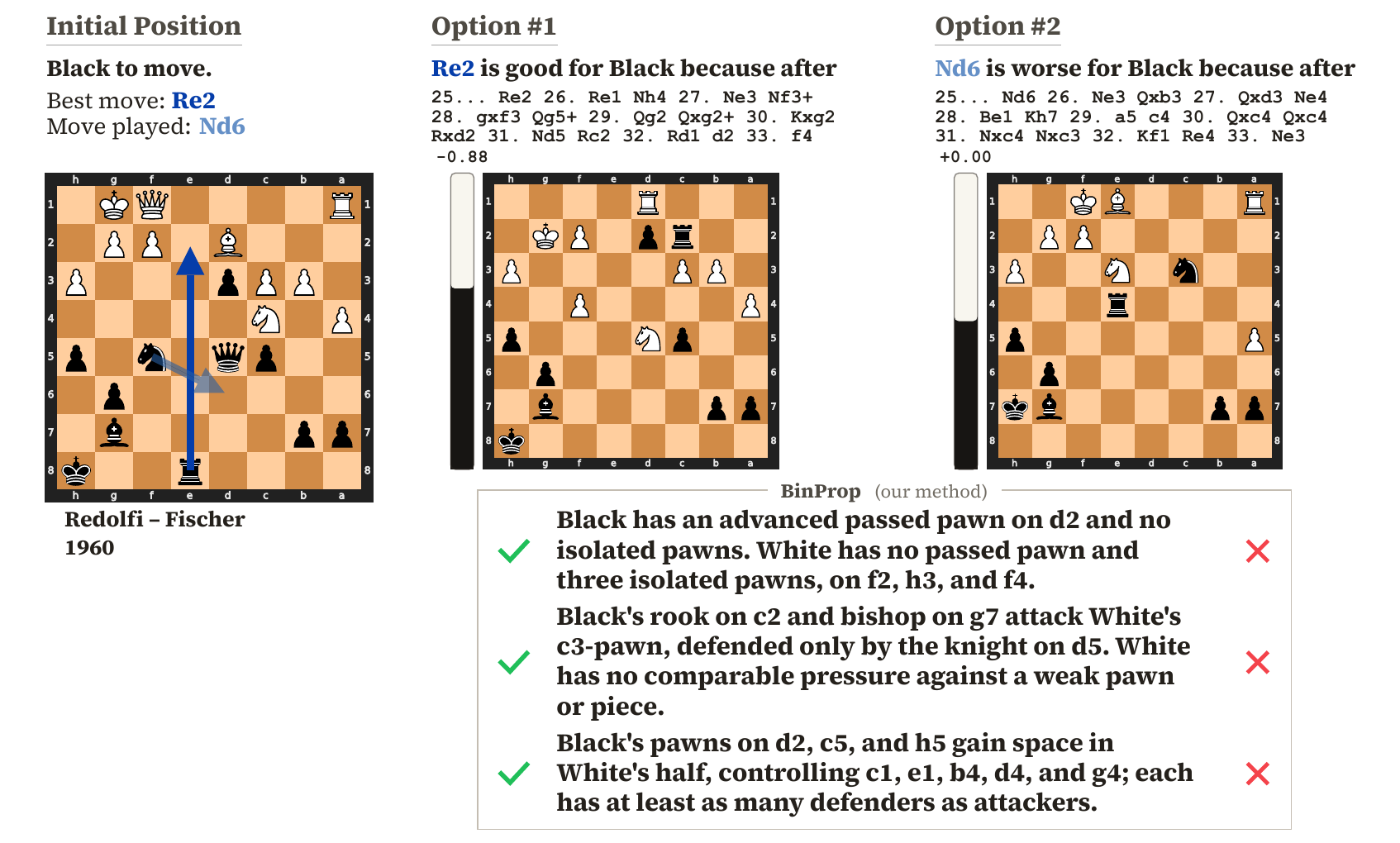}
\caption{Redolfi--Fischer (1960).}
\label{fig:website-opus250-7}
\end{figure}
\vfill
\begin{figure}[H]
\centering
\includegraphics[width=0.64\linewidth]{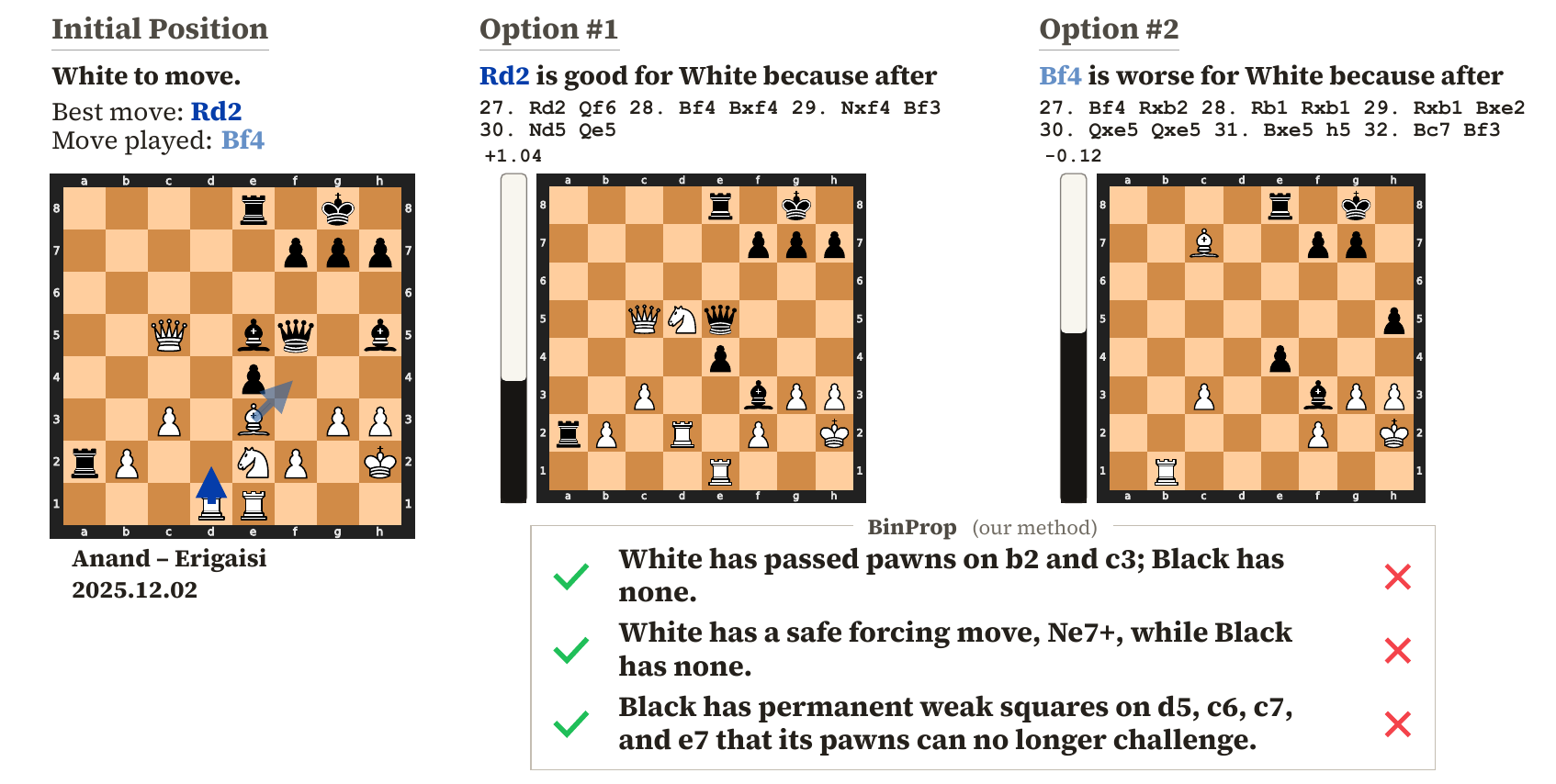}
\caption{Anand--Erigaisi (2025).}
\label{fig:website-opus250-8}
\end{figure}
\vfill
\begin{figure}[H]
\centering
\includegraphics[width=0.64\linewidth]{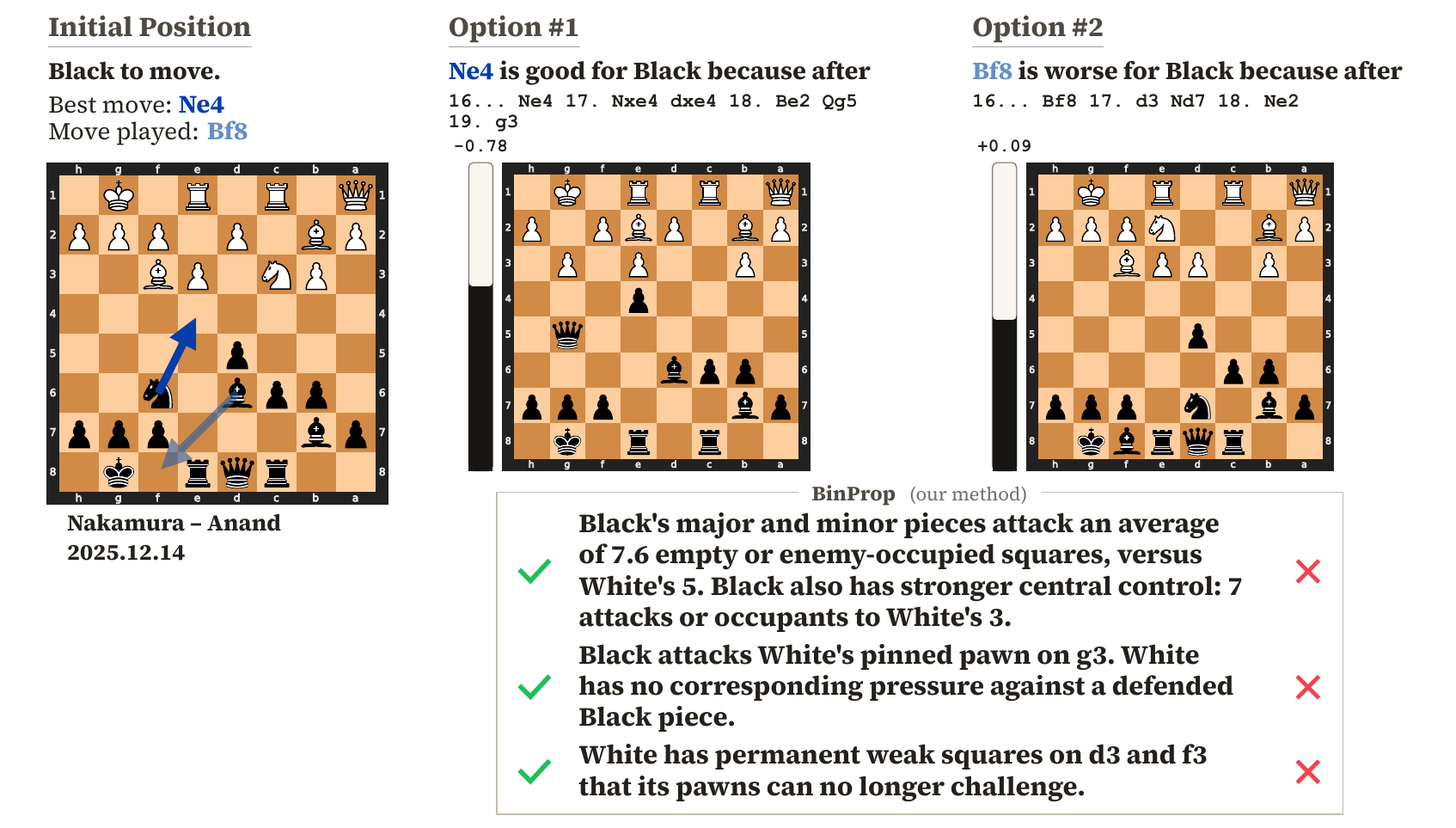}
\caption{Nakamura--Anand (2025).}
\label{fig:website-opus250-9}
\end{figure}

\endgroup

\clearpage

\section{Chess Motif Sets}
\label{app:chess-motif-catalog}
The following tables list all 192 BinProp motifs used for the chess results in Table~\ref{tab:chess_results}. Names and descriptions are model-generated. All descriptions refer to the side to move. $\Delta(f)$ is the difference in how often a motif holds on winning and losing positions, in percentage points.
\begingroup
\small
\setlength{\tabcolsep}{4pt}
\renewcommand{\arraystretch}{1.12}
\setlength{\LTleft}{0pt}
\setlength{\LTright}{0pt plus 1fill}
\setlength{\LTpre}{5pt}
\setlength{\LTpost}{0pt}
\setlength{\emergencystretch}{1.5em}
\setlength{\LTcapwidth}{\textwidth}
\captionsetup[longtable]{singlelinecheck=false,justification=raggedright,width=\textwidth}
\subsection{Fable 5}
\label{app:motifs-fable5}
\begin{longtable}{@{}>{\raggedright\arraybackslash}p{0.285\linewidth}>{\raggedright\arraybackslash}p{\dimexpr0.715\linewidth-4\tabcolsep-34pt\relax}>{\raggedleft\arraybackslash}p{34pt}@{}}
\caption{Chess motifs discovered by \texttt{Fable 5}.}\label{tab:motifs-fable5}\\
\toprule
Motif & Description & $\Delta(f)$ \\
\midrule
\endfirsthead
\multicolumn{3}{l}{\texttt{Fable 5} (continued)}\\
\toprule
Motif & Description & $\Delta(f)$ \\
\midrule
\endhead
\midrule
\multicolumn{3}{r}{\footnotesize Continued on next page}\\
\endfoot
\bottomrule
\endlastfoot
Material advantage or hanging enemy piece & The side to move is at least a pawn ahead in material, or material is level while an enemy minor piece or better hangs to a free or favorable capture. & 56.1 \\
\addlinespace[0.4em]
Consolidated queenless material lead or dual margin-trump parity edge & Not in check, with no unit hanging or attacked by a cheaper piece, a calm own king ring and no advanced enemy passer, the side to move is either a clean point or more up in a fully queenless position (a single-pawn edge requiring one comparative trump) or at exact material parity holding two or more margin trumps among passers, pawn health, rook activity, bishop pair, outposts, space, development and king quality. & 16.3 \\
\addlinespace[0.4em]
Concrete target with safe camp at material parity or better & Not behind in material, not in check, and with none of its own pieces profitably capturable, the side to move owns a concrete target: a static-exchange-winning capture, a besieged weak enemy pawn, exclusive heavy-piece infiltration, a safe advanced passer, a trapped ventured enemy minor, or a dominant attack on the enemy king ring. & 16.2 \\
\addlinespace[0.4em]
Positional edge at level material & With no material advantage, the side to move holds a clear positional plus from safe piece mobility, pressure on the enemy king zone, passed pawns, open rook files, king shelter and the bishop pair, while none of its own pieces are loose. & 12.0 \\
\addlinespace[0.4em]
Space, development and initiative dominance near material parity & With material within a fraction of a pawn, the side to move holds a broad concrete edge combining space, central control, development lead, pressure on enemy pieces, king safety, passed pawns and rook placement. & 10.2 \\
\addlinespace[0.4em]
Pawn health and piece activity superiority at equal material & Material is exactly equal but the side to move has clearly healthier pawns, better passed pawns, more active rooks and outposts, a safer king and more pressure on the enemy king zone. & 8.8 \\
\addlinespace[0.4em]
Calm parity squeeze against a structurally burdened opponent & At exact material parity, with none of its pieces hanging or attacked by lesser men and its own king ring calm, the side to move holds a durable plus such as an exclusive open-file heavy battery, an advanced protected pawn duo, a rim-stranded enemy knight, a blocked-pawn-buried enemy bishop, shut-in enemy rooks against its own free rook, or a far healthier pawn structure. & 7.1 \\
\addlinespace[0.4em]
Sole bishop pair at exact material equality & With material exactly level, not in check, none of its pieces hanging and its king zone not swarmed, the side to move owns two bishops on opposite color complexes while the opponent lacks the bishop pair. & 4.4 \\
\addlinespace[0.4em]
Dynamic compensation without material edge & With material equal or slightly down and its own pieces basically safe, the side to move has concrete compensation such as a safe fork or double attack, a favorably capturable or pinned enemy piece, an advanced supported passer, a manned assault on an exposed enemy king, a development lead against a stranded central king, or a broad activity and structure edge. & 4.3 \\
\addlinespace[0.4em]
Parity bind: bad enemy bishop, pawn clamp, or queenless outside majority & At exact material parity with a calm camp, no development lag and no advanced enemy passer, the side to move either faces a hemmed-in bad enemy bishop while owning a good minor, or clamps space with three secure pawns beyond the middle against none, or in a queenless position holds a clean pawn majority on a wing far from the enemy king. & 2.6 \\
\addlinespace[0.4em]
Entrenched minor-piece outpost at level material or clean check refutation & With material exactly level, a safe camp and king, the side to move has a knight or bishop on an advanced pawn-proof supported outpost that the opponent cannot match, or it is in check but can capture the lone checker without loss and stand one to three points ahead. & 2.4 \\
\addlinespace[0.4em]
Parity shelter or file grip on enemy king, or queenless grind edge & At exact material parity with a safe camp, the side to move either keeps a clearly healthier king shelter than the enemy with queens on, or aims a heavy piece down a file near the enemy king that holds no defending pawns while its own shelter is better, or in a queenless position holds a clear combined edge in passers, king activity, rook activity and pawn health. & 1.8 \\
\addlinespace[0.4em]
Temporary material deficit repaid with interest by winning capture & The side to move is a pawn or more behind on raw material, yet owns a legal capture whose static exchange gain exceeds the deficit (or exactly repays it while a second exchange-winning target hangs), and the opponent has no bigger counter-capture elsewhere. & 1.7 \\
\addlinespace[0.4em]
Mid-exchange full recapture restoring the balance with an edge & Temporarily down material mid-exchange, the side to move has a capture whose static-exchange value fully repays the deficit to roughly level, after which the opponent has no profitable capture and the mover keeps the better structure, king pressure, passers, rook files or bishops. & 1.7 \\
\addlinespace[0.4em]
Cornered intruder roundup or minor-army imbalance & Not ahead in raw material and not in check, the side to move either besieges an enemy piece stranded deep in its camp with no safe retreat so it can be rounded up for a net gain, or at level count owns two extra minor pieces against rook-and-pawn style compensation or a full piece against three pawns. & 1.6 \\
\addlinespace[0.4em]
Wing king behind fully open file assaulted at exact parity & At exact material parity, not in check, with its queen on and no friendly unit profitably capturable, the side to move keeps own pawns ahead of its king on every adjacent file while the enemy king sits on a wing back rank behind at least one adjacent file holding no enemy pawn in front of it, and strictly more mover pieces or a thorn pawn bear on the enemy king field than enemy pieces bear on the movers king. & 1.4 \\
\addlinespace[0.4em]
Asymmetric shelter breach attack at exact material parity & With material exactly level, queens on for the mover, both kings on their back two ranks and none of its pieces loose, the side to move keeps a fully intact pawn shelter while the enemy wing king is missing a near shelter pawn on at least one adjacent file and faces at least as many, and at least two, of the movers pieces bearing on its king zone. & 1.3 \\
\addlinespace[0.4em]
Stranded enemy king with queens on or refutable check & Near material parity the side to move faces an enemy king stranded on its back rank without castling rights or wandered up the board while queens stay on and its own king is sheltered, or it can win a trapped enemy queen, or it stands materially fine after a check whose giver hangs to a favorable capture. & 1.3 \\
\addlinespace[0.4em]
Queen-led siege of a castling-less central enemy king & At material parity or down at most one pawn, the side to move keeps its queen and a secure king while the enemy king is stuck on the d, e or f file near its home rank with no castling rights, backed by an adjacent half-open file, king-ring pressure or a development lead. & 1.1 \\
\addlinespace[0.4em]
Wrecked enemy king shelter with attacker surplus near parity & Near material parity with its queen still on board and its own king ring not swarmed, the side to move besieges an enemy king whose pawn shelter is destroyed or clearly inferior, holding strictly more distinct attackers on the enemy king ring than the opponent has on its own. & 1.0 \\
\addlinespace[0.4em]
\end{longtable}
\clearpage
\subsection{Opus 5}
\label{app:motifs-opus5}
\begin{longtable}{@{}>{\raggedright\arraybackslash}p{0.285\linewidth}>{\raggedright\arraybackslash}p{\dimexpr0.715\linewidth-4\tabcolsep-34pt\relax}>{\raggedleft\arraybackslash}p{34pt}@{}}
\caption{Chess motifs discovered by \texttt{Opus 5}.}\label{tab:motifs-opus5}\\
\toprule
Motif & Description & $\Delta(f)$ \\
\midrule
\endfirsthead
\multicolumn{3}{l}{\texttt{Opus 5} (continued)}\\
\toprule
Motif & Description & $\Delta(f)$ \\
\midrule
\endhead
\midrule
\multicolumn{3}{r}{\footnotesize Continued on next page}\\
\endfoot
\bottomrule
\endlastfoot
Material advantage for the side to move & The side to move has a net material edge of at least one pawn using standard piece values. & 48.4 \\
\addlinespace[0.4em]
Positional dominance without material deficit & The side to move wins the overall balance of positional features (piece mobility, space, central control, passed pawns, pawn health, bishop pair, rook files, outposts) while not being behind in material. & 34.8 \\
\addlinespace[0.4em]
Permanent holes in the enemy camp & The side to move controls clearly more permanent weak squares deep in the opponent half - squares that no enemy pawn can ever attack again and that it already covers with at least as many units as the defender - than the opponent controls in its own half, while not being down material. & 18.8 \\
\addlinespace[0.4em]
Superior pawn structure and passed pawn edge & The side to move has a clearly healthier pawn skeleton than the opponent - more or further advanced passed pawns and fewer isolated, doubled, backward or scattered pawns - while not being down significant material. & 15.4 \\
\addlinespace[0.4em]
Enemy king stranded without shelter & With plenty of pieces still on the board and no material deficit for the side to move, the opponent's king is far worse sheltered than the king of the side to move: it sits on or near the central files, has lost the right to castle without reaching a corner, has drifted off its home rank, is missing the pawns that should stand in front of it, stands beside open or semi-open files, and/or is surrounded by enemy pieces. The side to move owns real attacking material (a queen or rook plus other pieces), so the exposed king is a concrete target. & 13.1 \\
\addlinespace[0.4em]
Safe piece infiltration in enemy territory & The side to move has clearly more securely posted active pieces inside the opponent half - units that no enemy pawn can hit now or after one push, that cannot be profitably captured, and that still have targets or free squares - than the opponent has in its own half, while not being down material. & 10.8 \\
\addlinespace[0.4em]
Space-gaining pawn wedge in the centre & The side to move owns a clear space advantage: several of its pawns (at least one on a central file) stand on the fourth rank or beyond and grip squares in the opponent half, none of these advanced pawns is over-attacked, and it is not behind in material. & 10.3 \\
\addlinespace[0.4em]
Opponent starved of forcing resources & The side to move owns concrete forcing resources - a capture that wins material, a safe check, a safe pawn push that hits an enemy piece, or a promotion - while the opponent, even if handed a free move, would have none or far fewer of them, so the whole initiative belongs to the side to move. & 10.1 \\
\addlinespace[0.4em]
Own camp free of enemy pressure & The side to move has nothing concrete to worry about at home: none of its pieces is hanging, harassed by a cheaper unit, over-attacked or pinned, no enemy piece or far advanced enemy pawn sits unchallenged inside its own half, and at most one enemy piece eyes the squares around its king. & 9.6 \\
\addlinespace[0.4em]
Opponent lags in mobilization and king shelter & With plenty of pieces still on the board and without being down material, the side to move is clearly better mobilized than the opponent: fewer minor pieces still on the first rank, less back-rank clutter, no corner-boxed rook, and a king already tucked away or free to castle while the opponent king sits in the centre or has wandered out. & 9.5 \\
\addlinespace[0.4em]
Pressure on weak squares around the enemy king & The side to move hits more completely undefended squares in the ring and front zone of the opponent king, with at least two attacking units and more total pressure there than the opponent generates around its own king. & 9.5 \\
\addlinespace[0.4em]
Pressure on the opponent weak pawns & The side to move puts more pressure on the opponent structurally weak or under-defended pawns (and harasses more enemy pieces with its own pawns) than the opponent does in return, while not being materially down. & 8.6 \\
\addlinespace[0.4em]
Opponent saddled with a bad minor piece & The opponent owns at least one clearly poor knight or bishop - almost no safe squares, still stuck on its home rank, rim-bound, or a bishop buried behind its own same-coloured and blocked pawns - while every minor piece of the side to move is healthy, with material roughly balanced. & 8.0 \\
\addlinespace[0.4em]
Winning capture with no counter-strike & The side to move has a capture whose full exchange sequence nets material and leaves it at least level in material, while the opponent, even if handed a free move, has no capture that wins as much back. & 7.3 \\
\addlinespace[0.4em]
Opponent pieces tied to passive defence & The opponent has clearly more pieces stuck in passive defensive duty than the side to move - non-pawn units that must guard an attacked friendly pawn, act as the only guard of an attacked unit, or sit in front of a passed pawn as blockader - while the side to move is not down material. & 6.6 \\
\addlinespace[0.4em]
Dangerous advanced passed pawn & The side to move owns a clearly more dangerous passed pawn than the opponent - advanced, unblockaded, with a clear or well controlled promotion path and pawn, neighbour-passer or heavy-piece support - while not being down material. & 5.0 \\
\addlinespace[0.4em]
Opponent moves keep walking into loss of material & A far larger share of the opponent possible moves would leave the moved piece capturable at a profit than of the side-to-move own moves, while the opponent has no capture that wins more material than the side to move can win, so the opponent is in a bind with hardly any safe option. & 4.8 \\
\addlinespace[0.4em]
Checking initiative against the enemy king & The side to move can deliver more safe checks (checking moves whose piece cannot be profitably captured) than the opponent could deliver if it were their turn, so it owns the forcing initiative. & 4.6 \\
\addlinespace[0.4em]
Enemy pawn over-extended into our camp & The opponent has a pawn that has pushed past the middle into the side-to-move half where no neighbouring enemy pawn can support it, and the side to move attacks that pawn while either blockading it or outnumbering its defenders, whereas the side to move has no such over-extended pawn itself and is not behind in material. & 4.4 \\
\addlinespace[0.4em]
Opponent burdened with two different kinds of weakness & The opponent camp holds targets of at least two different kinds - structurally weak pawns on half-open files, pieces under fire or about to be kicked by a pawn push, and a draughty or centralised king - and clearly more such targets in total than the side to move has at home, while the side to move is not down material. & 3.8 \\
\addlinespace[0.4em]
Favorable piece imbalance without a material lead & The side to move has no lead in raw material (standard 1/3/3/5/9 count is level within half a pawn), yet owns the structurally better army: the bishop pair in an open position, minor pieces traded against enemy pawns while pieces still dominate, two minors against a rook, knights in a locked pawn structure, or bishops in an open one. The armies must differ in composition - identical material on both sides never qualifies. & 3.6 \\
\addlinespace[0.4em]
Grip on the opponent weak colour complex & With material roughly level, the side to move keeps a bishop on a colour complex where the opponent has no bishop and few pawns, and grips the resulting unguarded squares of that colour in the opponent camp far more than the opponent grips the other complex. & 3.4 \\
\addlinespace[0.4em]
True material edge hidden without a visible lead & The raw count shows no clear material lead for the side to move, yet it is really the one ahead: either it owns the stronger pieces in a material imbalance while standing slightly up on points, or an enemy piece has strayed into the half of the board belonging to the side to move where it has no safe square left and can be attacked more times than it is defended. & 3.1 \\
\addlinespace[0.4em]
Enemy queen harassed with tempo & The side to move can gain a tempo by safely attacking the opposing queen with a lesser piece, while its own queen cannot be harassed the same way and enjoys at least as many safe squares. & 3.1 \\
\addlinespace[0.4em]
Stronger double attack threat with nothing hanging & The side to move can make a safe move that sets up a bigger threat package (a fork, a hit on a more valuable or undefended enemy piece, or a threat combined with check) than the best threat the opponent could set up, while none of its own pieces are hanging or attacked by a cheaper unit. & 2.6 \\
\addlinespace[0.4em]
Enemy camp riddled with concrete pressure points & With material level or at most a pawn ahead, the side to move creates far more concrete problems in the opponent camp - hanging or pawn-harassed pieces, over-attacked units, pinned targets and pieces held only by a single overloaded or unstable guard - than the opponent creates against it. & 2.6 \\
\addlinespace[0.4em]
Opponent pieces cornered and hanging & Without holding more than a one-pawn material lead, the side to move troubles the opponent pieces far more than its own are troubled: more enemy knights, bishops, rooks and queens have at most one safe square available, and more of them stand attacked while undefended, hit by a cheaper unit, or attacked more times than defended. & 2.0 \\
\addlinespace[0.4em]
Tempo-gaining threat creation dominance & With material roughly level, the side to move has clearly more quiet moves that safely create a brand-new attack on an enemy unit (hitting it with a cheaper piece or hitting an undefended one) than the opponent can create in reply, so it keeps making threats with tempo. & 1.8 \\
\addlinespace[0.4em]
More active king in a simplified position & With the queens traded and only a few pieces left, the side to move has the far more active king - more centralized, off its own back rank and closer to the pawns on the board - than the opponent, while not being behind in material. & 1.6 \\
\addlinespace[0.4em]
Line pressure with pins and x-rays & The side to move owns more line power against the opponent - absolute pins, x-ray alignments where a more valuable enemy unit stands behind a lesser one, sliders stacked in batteries and latent discovered-attack setups - than the opponent has in return. & 1.2 \\
\addlinespace[0.4em]
\end{longtable}
\clearpage
\subsection{GPT-5.5}
\label{app:motifs-gpt55}
\begin{longtable}{@{}>{\raggedright\arraybackslash}p{0.285\linewidth}>{\raggedright\arraybackslash}p{\dimexpr0.715\linewidth-4\tabcolsep-34pt\relax}>{\raggedleft\arraybackslash}p{34pt}@{}}
\caption{Chess motifs discovered by \texttt{GPT-5.5}.}\label{tab:motifs-gpt55}\\
\toprule
Motif & Description & $\Delta(f)$ \\
\midrule
\endfirsthead
\multicolumn{3}{l}{\texttt{GPT-5.5} (continued)}\\
\toprule
Motif & Description & $\Delta(f)$ \\
\midrule
\endhead
\midrule
\multicolumn{3}{r}{\footnotesize Continued on next page}\\
\endfoot
\bottomrule
\endlastfoot
Material surplus & The side to move has at least a pawn of material advantage, with a small bonus for the bishop pair. & 48.3 \\
\addlinespace[0.4em]
Material-parity king hook pressure & With material roughly balanced, the side to move has pressure against a home-rank central king or a castled king shelter hook. & 8.7 \\
\addlinespace[0.4em]
Balanced castling lead and invader tempo & Without a large material edge, the side to move has a castling/development initiative, a safe major-piece invader, or a protected tempo on an overextended queen. & 8.7 \\
\addlinespace[0.4em]
Secure central grip and pawn targets & With material roughly balanced, the side to move has a safe king and either a supported central pawn grip or clearly better pressure on weak enemy pawns. & 8.4 \\
\addlinespace[0.4em]
Overloaded central shield and structural edge & The side to move has a coordinated attack on a pinned central king shield, a safe knight fork, or a queenless pawn-structure edge at material parity. & 7.1 \\
\addlinespace[0.4em]
Wing pawn tempo and central knight wedge & The side to move has a protected wing-pawn tempo on an advanced piece, a complementary fifth-rank knight wedge, or a queen-and-pawn phalanx against a central king. & 7.0 \\
\addlinespace[0.4em]
Supported central pawn tempo versus sleepy pieces & With material close, the side to move has a safe king and either a development bind from enemy back-rank minors or a supported central pawn break that gains time on a piece. & 6.8 \\
\addlinespace[0.4em]
Under-material checking spear against exposed king & Without a material surplus, the side to move has a coordinated pawn, minor, or rook spear against an exposed enemy king. & 6.6 \\
\addlinespace[0.4em]
Castling reserve versus loosened king & The side to move has castling reserve or a safer flank king while the opponent has a loose flank shelter or an exposed central king, with material close enough for king safety to matter. & 6.4 \\
\addlinespace[0.4em]
Wing pawn roller and queen net & The side to move has a safe king and either a supported advanced wing-pawn roller aimed at the enemy king or a lower-piece move that traps or wins the enemy queen. & 5.9 \\
\addlinespace[0.4em]
Bad bishop bind and active rook conversion & With material close, the side to move exploits a color-complex bad bishop, superior simplified rook activity, or a safe en-passant resource. & 5.8 \\
\addlinespace[0.4em]
Safe queen middlegame activity edge & With material roughly level and queens on, the side to move has a safe flank king and a clear activity or target edge. & 5.8 \\
\addlinespace[0.4em]
Compensated central bind and rook-pair edge & Without a large raw material edge, the side to move has a compensated positional bind from an exposed central king, an active rook-pair imbalance, or a central pawn grip against unchallenged bishops. & 5.7 \\
\addlinespace[0.4em]
Balanced file battery against an uncastled king & With material close, the side to move has a major-piece file battery, active bishops, or a protected spear aimed at an uncastled opponent king. & 5.3 \\
\addlinespace[0.4em]
Material-neutral mobility and development bind & With no pawn-up material edge, the side to move has a clear mobility, space, development, or active-rook bind. & 5.3 \\
\addlinespace[0.4em]
Balanced safe advanced knight post & The side to move is not relying on a large material edge and has the better safe central knight advanced into the opponent half. & 4.9 \\
\addlinespace[0.4em]
Balanced queen-zone storm or central pin & With material still roughly balanced, the side to move has queen-zone pressure, a pawn storm, or a pinning slider against a poorly sheltered opposing king. & 4.5 \\
\addlinespace[0.4em]
Piece quality targets or stranded central king & With material roughly balanced, the side to move has either active pieces for pawns with a loose target or a safe flank king against a stranded central enemy king. & 3.8 \\
\addlinespace[0.4em]
Protected queen chase and deep spear & Without a large material edge, the side to move has a protected lesser-piece tempo on the enemy queen, a deep safe queen invasion, or an immediate favorable capture. & 3.6 \\
\addlinespace[0.4em]
Material-parity forcing gain or mobile bishop pair & With no existing material edge, the side to move has either an immediate material-winning tactic or an active bishop pair against a side without one. & 3.5 \\
\addlinespace[0.4em]
Queenless rook pressure and loose outpost target & With no large material edge, the side to move has queenless rook-file activity or safe-king pressure on an overextended loose enemy piece. & 3.2 \\
\addlinespace[0.4em]
Material-parity initiative targets & The side to move is not materially behind or clearly ahead, but has a concrete initiative such as loose-piece pressure, a strong knight, an advanced passer, or a seventh-rank invasion. & 3.1 \\
\addlinespace[0.4em]
Legal castling reserve and twin outpost squeeze & At material parity or a pawn down, the side to move can castle against a lagging central king or has twin protected advanced knights creating concrete threats. & 3.1 \\
\addlinespace[0.4em]
Queen imbalance or advanced pawn lever & The side to move has a favorable queen-for-material imbalance or a dangerous advanced pawn that is passed or lodged near the enemy king. & 2.6 \\
\addlinespace[0.4em]
Balanced space clamp with loose-major tactics & With material not clearly unfavorable, the side to move has a safe major-piece tactic or a protected central space clamp against lagging development. & 2.5 \\
\addlinespace[0.4em]
Long-range compensation without material edge & The side to move is not up material but has active long-range compensation from bishops, rook lifts, or queen pressure on a central king. & 2.4 \\
\addlinespace[0.4em]
Loose-piece gain or central pawn bind & The side to move is not materially lost and has a concrete target through a profitable capture, a trapped corner knight, or a central pawn clamp against an uncastled king. & 2.3 \\
\addlinespace[0.4em]
Overextended pawn cleanup and backed spear & Without a material surplus, the side to move has a safe cleanup of an overextended pawn or a heavily backed pawn/rook squeeze. & 2.2 \\
\addlinespace[0.4em]
Balanced space phalanx and king-defender hit & With material not already decisive, the side to move has a defended space-development clamp or a safe minor/diagonal hit on a key defender near the enemy king. & 2.1 \\
\addlinespace[0.4em]
Parity x-ray pawn target and passer pickup & With no material surplus and a safe king, the side to move can safely remove an overextended enemy pawn or x-ray a deep weak pawn to a major piece or king. & 2.1 \\
\addlinespace[0.4em]
Rook-backed pawn center and knight-pair shelter edge & With material close, the side to move has the safer king plus either a rook-backed connected center, a hard knight outpost, or knight-pair pressure against a knightless weak shelter. & 2.1 \\
\addlinespace[0.4em]
Exchange clamp and backed outpost pressure & With material not already decisive, the side to move has a safe exchange clamp, pawn-backed outpost, or coordinated queen pressure against restricted enemy pieces. & 1.9 \\
\addlinespace[0.4em]
Tempo net on queen or invader & With material close, the side to move has a safe developing queen tempo, a queen sortie into a weakened shelter, or a pawn net around an overadvanced minor. & 1.8 \\
\addlinespace[0.4em]
Material-balanced exposed king pressure & Without a material surplus, the side to move has concentrated piece pressure against an exposed or overburdened opponent king. & 1.8 \\
\addlinespace[0.4em]
Deep spear pins and checking reserves & At near material balance, the side to move has a deep queen spear, a strong slider pin, multiple safe checks, or bishop pressure against knight-only defenders. & 1.7 \\
\addlinespace[0.4em]
Safe queen-minor flank-king battery & The side to move has a safe king and a queen-minor or back-rank rook battery against an exposed flank king without relying on a large material edge. & 1.6 \\
\addlinespace[0.4em]
Enemy-camp wedge and deep minor raider & Without a large material edge, the side to move has a protected far-advanced pawn wedge or minor piece lodged deep in the enemy camp. & 1.6 \\
\addlinespace[0.4em]
Queenless small-edge endgame conversion & With queens off and material only slightly favorable, the side to move has a clean simplified edge, a central knight, connected advanced pawns, or a safe major-piece pickup. & 1.4 \\
\addlinespace[0.4em]
Minor swarm imbalance and clean major tactic & Without a material surplus, the side to move has a favorable minor-piece swarm, a queenless chase of a central king, or a concrete safe major-piece tactic. & 1.3 \\
\addlinespace[0.4em]
Supported queen lift and central overload & With material close and a safe king, the side to move has a supported queen lift against a castled shelter, an open major-piece overload near a central king, or a queenless bishop-pair bind. & 1.2 \\
\addlinespace[0.4em]
Central file pressure with loose wedge pickup & With material close, the side to move has central-file rook pressure, a queen-rook battery on a weak target, or a safe pickup of an overadvanced enemy wedge. & 1.2 \\
\addlinespace[0.4em]
Compensating minor net and rook-lift spear & Without a large material edge, the side to move has a safe king with a minor-piece net, deep pawn wedge, or rook-lift spear aimed at the enemy king. & 1.2 \\
\addlinespace[0.4em]
Sub-pawn exchange and sixth-rank activity & Without a pawn-sized material surplus, the side to move has a fractional exchange edge, a safe sixth-rank rook, a no-bishop knight storm, or a concrete low-piece capture of a trapped major piece. & 1.1 \\
\addlinespace[0.4em]
Dual-reserve central king squeeze & With material close, the side to move keeps flexible castling or coordinated slider pressure while the opposing king is still stuck near the center. & 1.0 \\
\addlinespace[0.4em]
\end{longtable}
\clearpage
\subsection{GLM-5.2}
\label{app:motifs-glm52}
\begin{longtable}{@{}>{\raggedright\arraybackslash}p{0.285\linewidth}>{\raggedright\arraybackslash}p{\dimexpr0.715\linewidth-4\tabcolsep-34pt\relax}>{\raggedleft\arraybackslash}p{34pt}@{}}
\caption{Chess motifs discovered by \texttt{GLM-5.2}.}\label{tab:motifs-glm52}\\
\toprule
Motif & Description & $\Delta(f)$ \\
\midrule
\endfirsthead
\multicolumn{3}{l}{\texttt{GLM-5.2} (continued)}\\
\toprule
Motif & Description & $\Delta(f)$ \\
\midrule
\endhead
\midrule
\multicolumn{3}{r}{\footnotesize Continued on next page}\\
\endfoot
\bottomrule
\endlastfoot
Minority pawn majority lever & The side to move has more pawns than the opponent on either the same-file center pair (d+e files) or on both wings combined (a-c plus f-h files), with total pawn count at least equal, indicating a lever-ready structural edge. & 34.6 \\
\addlinespace[0.4em]
Material advantage two or more & The side to move has at least two points more total piece value (pawn=1, knight/bishop=3, rook=5, queen=9) than the opponent. & 15.3 \\
\addlinespace[0.4em]
Pawn island advantage & The side to move has fewer pawn-island groups (maximal sets of pawns on contiguous files) than the opponent, indicating a more cohesive pawn structure. & 15.3 \\
\addlinespace[0.4em]
Mobility advantage over opp & The side to move has at least five more legal moves than the opponent. & 13.0 \\
\addlinespace[0.4em]
Central piece value lead & The side to move has at least two more total piece value on the central 4x4 box (files c through f, ranks 3 through 6) than the opponent. & 10.9 \\
\addlinespace[0.4em]
Development edge advantage & The side to move has strictly fewer undeveloped (back-rank) knights or bishops than the opponent, indicating superior piece development and initiative. & 5.4 \\
\addlinespace[0.4em]
Central pawn density pressure advantage & The side to move has more pawns and pieces on files c through f within its attack zone than the opponent has on the corresponding files, indicating central pressure and space, with a non-pawn attacker of the centre. & 4.8 \\
\addlinespace[0.4em]
Dual check sources & The side to move has legal checking moves from at least two different pieces, indicating coordinated pressure on the opponent king. & 3.8 \\
\addlinespace[0.4em]
King pawn shield advantage & The side to move has a complete pawn shield in front of its back-rank king while the opponent king lacks such a shield. & 3.6 \\
\addlinespace[0.4em]
Royal fork check targets major piece & The side to move has a legal checking move that either captures an enemy rook or queen or, after the move, attacks an enemy rook or queen with the moved piece. & 3.5 \\
\addlinespace[0.4em]
Nonpawn king zone attack ratio & The side to move has at least two non-pawn pieces attacking the opponent king's 3x3 zone while the opponent has at most one non-pawn piece attacking the side-to-move king's 3x3 zone. & 3.1 \\
\addlinespace[0.4em]
Bishop pair advantage & The side to move has the bishop pair while the opponent has at most one bishop. & 2.6 \\
\addlinespace[0.4em]
Favorable trade capture & The side to move has a legal capture of an enemy non-pawn piece that wins at least two points of material in the exchange or is undefended, bringing the static material balance to at least even. & 2.5 \\
\addlinespace[0.4em]
Supported minor attacks opp king zone & The side to move has a knight or bishop that is defended by a friendly pawn and that attacks a square adjacent to the opponent king, indicating a stable outpost applying pressure near the enemy king. & 2.4 \\
\addlinespace[0.4em]
Central pawn phalanx advantage & The side to move has a more advanced pair of friendly pawns on adjacent central files (c-d, d-e, or e-f) than any such opponent pawn phalanx, indicating central tension and space control. & 2.4 \\
\addlinespace[0.4em]
Attacks hanging opp king neighbor & The side to move has at least one legal, non-sacrificial check or capture on a square adjacent to the opponent king. & 2.2 \\
\addlinespace[0.4em]
Attacks opp strong piece asymmetry & The side to move has at least one safe attack by a non-pawn piece on an enemy non-pawn piece while the opponent has no safe non-pawn attack on the side-to-move non-pawn pieces. & 1.5 \\
\addlinespace[0.4em]
Central knight outpost & The side to move has a knight on a central square (files c-f, ranks 4-5) defended by a friendly pawn and not attackable by any enemy pawn. & 1.5 \\
\addlinespace[0.4em]
Rook uncontested on halfopen file & The side to move has a rook on a file with no friendly pawns where the opponent has no rook or queen on that file. & 1.4 \\
\addlinespace[0.4em]
Backward pawn advantage & The side to move has no backward pawns while the opponent has at least one backward pawn. & 1.3 \\
\addlinespace[0.4em]
Active major in opp back two ranks & The side to move has a rook or queen in the opponent's back two ranks attacking an enemy pawn while the opponent has no rook or queen in the side-to-move's back two ranks attacking side-to-move pawns. & 1.1 \\
\addlinespace[0.4em]
\end{longtable}
\clearpage
\subsection{Kimi-K3}
\label{app:motifs-kimi3}
\begin{longtable}{@{}>{\raggedright\arraybackslash}p{0.285\linewidth}>{\raggedright\arraybackslash}p{\dimexpr0.715\linewidth-4\tabcolsep-34pt\relax}>{\raggedleft\arraybackslash}p{34pt}@{}}
\caption{Chess motifs discovered by \texttt{Kimi-K3}.}\label{tab:motifs-kimi3}\\
\toprule
Motif & Description & $\Delta(f)$ \\
\midrule
\endfirsthead
\multicolumn{3}{l}{\texttt{Kimi-K3} (continued)}\\
\toprule
Motif & Description & $\Delta(f)$ \\
\midrule
\endhead
\midrule
\multicolumn{3}{r}{\footnotesize Continued on next page}\\
\endfoot
\bottomrule
\endlastfoot
Material edge with activity tiebreak & The side to move has more total piece value than the opponent, or with equal material attacks more enemy pieces than the opponent attacks. & 65.2 \\
\addlinespace[0.4em]
Healthier pawn structure & The side to move has a clearly sounder pawn structure, with at least two fewer isolated, doubled, or backward pawns than the opponent. & 7.2 \\
\addlinespace[0.4em]
Development lead on even terms & With material roughly equal, the side to move has a clear development lead: at least two more units of development (minor pieces off their back rank plus a king tucked off the central file) than the opponent. & 5.8 \\
\addlinespace[0.4em]
Enemy-half pawn space edge & The side to move's pawns attack at least three more distinct squares on the opponent's half of the board than the opponent's pawns attack on the side-to-move's half, reflecting a clear space advantage. & 3.0 \\
\addlinespace[0.4em]
Accumulated positional pluses score & The side to move has at least three points of positional advantages such as the bishop pair, pinned enemy pieces, advanced passed pawns, rooks on open files, and the safer king. & 2.9 \\
\addlinespace[0.4em]
Bishop pair advantage & The side to move owns the bishop pair (two or more bishops) while the opponent has at most one bishop, a lasting positional edge. & 2.6 \\
\addlinespace[0.4em]
Check based double attack & With material not already favorable, the side to move has mate in one or a checking move that also leaves an enemy non-pawn piece with more attackers than defenders. & 2.0 \\
\addlinespace[0.4em]
Activity compensation without material edge & With equal or less material, the side to move controls many more squares than the opponent, showing superior piece activity. & 1.8 \\
\addlinespace[0.4em]
Enemy king stranded in center & With queens still on the board, the opponent king is stranded on a central file without safe castling escape while the side to move has its own king tucked away on a flank. & 1.3 \\
\addlinespace[0.4em]
King ring siege edge & The side to move has at least three pieces attacking squares adjacent to the enemy king, at least two more than the opponent has around the side-to-move king. & 1.1 \\
\addlinespace[0.4em]
Tactical pressure on loose enemy pieces & With material nearly even, the side to move creates more tactical pressure, in the form of enemy non-pawn pieces that are en prise, underdefended, or attacked by pawns, than it suffers on its own pieces. & 1.1 \\
\addlinespace[0.4em]
\end{longtable}
\clearpage
\subsection{DeepSeek-V4-Pro}
\label{app:motifs-deepseekv4}
\begin{longtable}{@{}>{\raggedright\arraybackslash}p{0.285\linewidth}>{\raggedright\arraybackslash}p{\dimexpr0.715\linewidth-4\tabcolsep-34pt\relax}>{\raggedleft\arraybackslash}p{34pt}@{}}
\caption{Chess motifs discovered by \texttt{DeepSeek-V4-Pro}.}\label{tab:motifs-deepseekv4}\\
\toprule
Motif & Description & $\Delta(f)$ \\
\midrule
\endfirsthead
\multicolumn{3}{l}{\texttt{DeepSeek-V4-Pro} (continued)}\\
\toprule
Motif & Description & $\Delta(f)$ \\
\midrule
\endhead
\midrule
\multicolumn{3}{r}{\footnotesize Continued on next page}\\
\endfoot
\bottomrule
\endlastfoot
King safety pawn storm & The side to move has more pawns on the wing where the opponent king is located than the opponent has on the side-to-move king wing, indicating a pawn storm advantage. & 16.8 \\
\addlinespace[0.4em]
King pawn shield intact & The side to move has more pawns on the files adjacent to its own king than the opponent has near theirs. & 12.2 \\
\addlinespace[0.4em]
King attack pressure & The side to move attacks more squares adjacent to the opponent king than the opponent attacks adjacent to the side-to-move king, with a margin of at least 2. & 12.0 \\
\addlinespace[0.4em]
Dominant center control & The side to move controls more central squares (d4, e4, d5, e5) than the opponent. & 9.5 \\
\addlinespace[0.4em]
Development lead & The side to move has more pieces (knights and bishops) off their back rank than the opponent, indicating a development advantage. & 7.5 \\
\addlinespace[0.4em]
Fewer pawn islands & The side to move has fewer pawn islands than the opponent, indicating a healthier pawn structure. & 6.9 \\
\addlinespace[0.4em]
Knight attacking king zone & The side to move has a knight attacking squares near the opponent king while the opponent has no equivalent attack. & 6.8 \\
\addlinespace[0.4em]
Restricted opponent pieces & The side to move has more opponent pieces (queen, rooks, bishops, knights) on ranks 1-2 from their perspective than the opponent has restricted pieces on the side-to-move side, indicating a space advantage cramping the opponent. & 6.7 \\
\addlinespace[0.4em]
Centralized active piece advantage & The side to move has a knight or bishop on or beyond the fourth rank (from its perspective) that attacks at least one central square, while the opponent has no such active minor piece. & 6.4 \\
\addlinespace[0.4em]
Advanced pawn chain advantage & The side to move has more pawns on the fifth rank or beyond than the opponent, indicating superior space and advanced structure. & 5.8 \\
\addlinespace[0.4em]
Rook on open file & The side to move has a rook on a file with no pawns of either color, while the opponent has no such rook. & 5.8 \\
\addlinespace[0.4em]
Safe pawn break threat & The side to move threatens a pawn break that would open lines against the opponent king, while the opponent has no comparable threat. & 5.6 \\
\addlinespace[0.4em]
Bishop blocked by own pawns & The side to move has a bishop on the same color as its own pawns on that color's central squares, without the opponent having the same restriction. & 4.9 \\
\addlinespace[0.4em]
Connected rooks & The side to move has two rooks connected (on same rank or file with no pieces between them) while the opponent does not. & 4.2 \\
\addlinespace[0.4em]
Bishop attacking king zone & The side to move has a bishop attacking squares adjacent to the opponent king while the opponent has no such bishop attack on the side-to-move king. & 3.9 \\
\addlinespace[0.4em]
Knight outpost attacking center & The side to move has a knight on a protected outpost square (immune to opponent pawns, defended by a friendly pawn) that attacks at least one central square (d4,e4,d5,e5), while the opponent has no such knight. & 3.7 \\
\addlinespace[0.4em]
Bishop on weak color complex & The side to move has a bishop and the opponent has no bishop that can guard the squares on the bishop color, because all their pawns are on squares of that bishop color, while the opponent does not have such a bishop-versus-pawns color weakness to exploit. & 3.6 \\
\addlinespace[0.4em]
Knight vs bishop control imbalance & The side to move has more knights on outpost squares (protected by pawns, not attackable by opponent pawns) than the opponent has bishops on open diagonals (diagonals with no friendly pawns blocking), indicating a positional piece-type advantage. & 3.6 \\
\addlinespace[0.4em]
Queenside pawn majority advantage & The side to move has more pawns on the queenside (files a-d) than the opponent, with a margin of at least 2, indicating a favorable pawn majority for creating passed pawns. & 3.2 \\
\addlinespace[0.4em]
Passed pawn protected & The side to move has a passed pawn that is defended by a friendly pawn, while the opponent has no such supported passed pawn. & 2.8 \\
\addlinespace[0.4em]
Pawn storm against kingside castled king & The side to move has at least two pawns on the files in front of the opponent castled king (f/g/h files for kingside castle), while the opponent has fewer such pawns against the side-to-move king. & 2.7 \\
\addlinespace[0.4em]
Bishop pair advantage & The side to move has both bishops while the opponent has at most one bishop. & 2.6 \\
\addlinespace[0.4em]
Knight on outpost & The side to move has a knight on a square that cannot be attacked by opponent pawns and is protected by a friendly pawn. & 2.1 \\
\addlinespace[0.4em]
Queen attacking king zone without opponent & The side to move has a queen attacking squares adjacent to the opponent king while the opponent has no queen attack on the side-to-move king zone. & 1.7 \\
\addlinespace[0.4em]
Rook invading opponent territory & The side to move has a rook on the opponent side of the board while the opponent has no rook on the side-to-move side. & 1.6 \\
\addlinespace[0.4em]
Unstoppable passed pawn & The side to move has a passed pawn that cannot be caught by the opponent. & 1.4 \\
\addlinespace[0.4em]
Rook behind passed pawn & The side to move has a rook behind its own passed pawn while the opponent has no rook behind their own passed pawn. & 1.2 \\
\addlinespace[0.4em]
Piece attacking king defended target & The side to move has a piece attacking an opponent piece that is only defended by the opponent king, while the opponent has no such attack, indicating a tactical vulnerability. & 1.2 \\
\addlinespace[0.4em]
\end{longtable}
\clearpage
\subsection{GPT-5.6 Sol}
\label{app:motifs-sol56}
\begin{longtable}{@{}>{\raggedright\arraybackslash}p{0.285\linewidth}>{\raggedright\arraybackslash}p{\dimexpr0.715\linewidth-4\tabcolsep-34pt\relax}>{\raggedleft\arraybackslash}p{34pt}@{}}
\caption{Chess motifs discovered by \texttt{GPT-5.6 Sol}.}\label{tab:motifs-sol56}\\
\toprule
Motif & Description & $\Delta(f)$ \\
\midrule
\endfirsthead
\multicolumn{3}{l}{\texttt{GPT-5.6 Sol} (continued)}\\
\toprule
Motif & Description & $\Delta(f)$ \\
\midrule
\endhead
\midrule
\multicolumn{3}{r}{\footnotesize Continued on next page}\\
\endfoot
\bottomrule
\endlastfoot
Net material advantage & The side to move has a higher total material value than the opponent. & 56.1 \\
\addlinespace[0.4em]
Healthy mobile wing majority & With balanced non-pawn forces and no advanced passer, the side to move has a healthier connected wing majority than the opponent's best countermajority. & 17.7 \\
\addlinespace[0.4em]
Level-material mobility dominance & With material values level, the side to move has substantially greater effective mobility for its non-pawn pieces. & 9.0 \\
\addlinespace[0.4em]
Safe forcing check against an exposed central king & With material roughly level, the side to move can give a non-sacrificial forcing check to an exposed king on the central files. & 4.5 \\
\addlinespace[0.4em]
Unopposed bishop pair at nominal parity & At equal conventional material with matched minor-piece counts, the side to move has bishops on both color complexes while the opponent does not. & 4.2 \\
\addlinespace[0.4em]
Durable material-reversing capture & With no nominal material edge and differing force inventories, the side to move can capture into a material lead that survives every immediate countercapture. & 3.8 \\
\addlinespace[0.4em]
Matched-force positional coordination & With matched non-pawn forces and at most a one-pawn investment, the side to move has a pronounced combined edge in space, piece placement, and king security. & 3.4 \\
\addlinespace[0.4em]
Targetable pawn-structure superiority & With identical material inventories, the side to move has substantially fewer isolated, doubled, backward, and loose pawns than the opponent. & 2.8 \\
\addlinespace[0.4em]
Secure advanced passed-pawn superiority & At near material parity, the side to move has a secure advanced passed pawn substantially stronger than any opposing passer. & 2.7 \\
\addlinespace[0.4em]
Wing-king pressure superiority at material parity & At equal material with queens and both kings on the wings, the side to move has concentrated pressure around the opposing king that clearly exceeds the opponent's attack and pawn-cover safety. & 2.7 \\
\addlinespace[0.4em]
Castled king versus central king & At equal material, the side to move has a pawn-sheltered castled king while the opponent king remains in the center. & 2.6 \\
\addlinespace[0.4em]
Queen-led multi-piece king-zone encirclement & At near material parity, the side to move has a coordinated queen-led attack around the opposing king that substantially exceeds the counterattack. & 2.1 \\
\addlinespace[0.4em]
Uncontested connected-rook file control & With identical material inventories, the side to move has connected rooks controlling a safe semi-open forward lane while the opposing rooks lack comparable coordination. & 2.1 \\
\addlinespace[0.4em]
Productive secure invasion superiority & Without a material edge, the side to move has a clearly superior secure invasion by productive non-pawn pieces in the opponent's half. & 2.0 \\
\addlinespace[0.4em]
Shelter-asymmetric safe wing check & At near material parity, the side to move has better pawn shelter and a non-sacrificial direct check against a less-sheltered wing king. & 1.6 \\
\addlinespace[0.4em]
Clean capture from identical material & With identical material inventories, the side to move can capture a more valuable piece or make a capture that has no immediate legal recapture. & 1.4 \\
\addlinespace[0.4em]
Ready castling versus a stranded central king & At near material parity, the side to move can castle immediately while the opponent has permanently lost castling rights with its king still in the center. & 1.3 \\
\addlinespace[0.4em]
Protected advanced knight outpost superiority & The side to move has a pawn-supported advanced knight outpost whose activity and stability clearly exceed any opposing knight outpost. & 1.2 \\
\addlinespace[0.4em]
Pawn-anchored central space domination & At nominal material parity, the side to move has a substantially stronger secure central space complex built around protected advanced pawns. & 1.2 \\
\addlinespace[0.4em]
Reinforced minor coalition against rook & The side to move has two extra minor pieces against an extra enemy rook, balanced by one enemy pawn or strengthened at equal pawns by a bishop pair or two pawn-supported minors. & 1.2 \\
\addlinespace[0.4em]
Coordinated deployment without a material edge & With matched non-pawn forces and at most a one-pawn deficit, the side to move has a pronounced deployment lead through active minors or a safe supported advanced rook. & 1.1 \\
\addlinespace[0.4em]
\end{longtable}
\clearpage
\subsection{Qwen3.5-397B-A17B}
\label{app:motifs-qwen35}
\begin{longtable}{@{}>{\raggedright\arraybackslash}p{0.285\linewidth}>{\raggedright\arraybackslash}p{\dimexpr0.715\linewidth-4\tabcolsep-34pt\relax}>{\raggedleft\arraybackslash}p{34pt}@{}}
\caption{Chess motifs discovered by \texttt{Qwen3.5-397B-A17B}.}\label{tab:motifs-qwen35}\\
\toprule
Motif & Description & $\Delta(f)$ \\
\midrule
\endfirsthead
\multicolumn{3}{l}{\texttt{Qwen3.5-397B-A17B} (continued)}\\
\toprule
Motif & Description & $\Delta(f)$ \\
\midrule
\endhead
\midrule
\multicolumn{3}{r}{\footnotesize Continued on next page}\\
\endfoot
\bottomrule
\endlastfoot
Material advantage & Side to move has more total material value than the opponent. & 56.1 \\
\addlinespace[0.4em]
Pawn majority on wings & Side to move has a pawn majority on at least one wing (queenside or kingside) compared to opponent. & 15.5 \\
\addlinespace[0.4em]
Fewer pawn islands & Side to move has fewer pawn islands than the opponent, indicating better pawn structure. & 15.3 \\
\addlinespace[0.4em]
Mobility advantage & Side to move has more legal moves available than the opponent. & 14.7 \\
\addlinespace[0.4em]
Center control & Side to move attacks more center squares (d4, d5, e4, e5) than the opponent. & 11.6 \\
\addlinespace[0.4em]
Rook open or semi open file & Side to move has at least one rook on a file with no opponent pawns (open or semi-open file). & 9.8 \\
\addlinespace[0.4em]
Castled with pawn shield & Side to move has castled and their king has at least 2 pawns on adjacent files for protection. & 9.7 \\
\addlinespace[0.4em]
Opponent king exposure & Side to move is winning because the opponent king is exposed or lacks a pawn shield. & 9.4 \\
\addlinespace[0.4em]
Piece in enemy territory & Side to move has at least one piece positioned on the opponent half of the board (ranks 4-7 for White, ranks 0-3 for Black). & 9.0 \\
\addlinespace[0.4em]
Central advanced pawn & Side to move has a pawn on the 5th rank or further on a central file (C, D, E, or F). & 6.6 \\
\addlinespace[0.4em]
Major piece coordination & Side to move has more major pieces (rooks and queen) with at least one defender than the opponent. & 5.5 \\
\addlinespace[0.4em]
Minor piece development & Side to move has at least 3 of 4 minor pieces developed off starting squares. & 3.8 \\
\addlinespace[0.4em]
Queen centralized & Side to move has queen on or attacking central squares d4, d5, e4, e5. & 3.6 \\
\addlinespace[0.4em]
Opponent knight on rim & Opponent has at least one knight on the edge of the board (a-file, h-file, 1st rank, or 8th rank). & 2.8 \\
\addlinespace[0.4em]
Bishop pair & Side to move has both bishops while opponent has fewer than two bishops. & 2.6 \\
\addlinespace[0.4em]
Centralized knight & Side to move has a knight on one of the four central squares (d4, d5, e4, e5) & 2.3 \\
\addlinespace[0.4em]
Piece fork potential & Side to move has a piece attacking two or more opponent pieces. & 1.6 \\
\addlinespace[0.4em]
\end{longtable}
\endgroup

\end{document}